\documentclass[10pt,twocolumn,letterpaper]{article}

\usepackage{cvpr}              % To produce the CAMERA-READY version
\usepackage{multirow} 
\usepackage[accsupp]{axessibility}
\usepackage[dvipsnames]{xcolor}
\usepackage{tikz}
\usepackage[abs]{overpic}
\usepackage[normalem]{ulem}

\usepackage[symbol]{footmisc}

\usepackage{tikz}
\usetikzlibrary{fadings,patterns}
\newcommand\shadetext[2][]{%
  \setbox0=\hbox{{#2}}%
  \tikz[baseline=0]\path [#1] \pgfextra{\rlap{\copy0}} (0,-\dp0) rectangle (\wd0,\ht0);%
}
\usepackage{blindtext}
\usepackage{ctable}

\definecolor{jiapeng}{rgb}{0.9, 0.4,0.1}

\definecolor{cvprblue}{rgb}{0.21,0.49,0.74}
\usepackage[pagebackref,breaklinks,colorlinks,citecolor=cvprblue]{hyperref}
\usepackage{todonotes}
\usepackage{makecell}

\def\paperID{3676} % *** Enter the Paper ID here
\def\confName{CVPR}
\def\confYear{2024}

\title{Motion2VecSets: 4D Latent Vector Set Diffusion \\ for Non-rigid Shape Reconstruction and Tracking}

\author{
    Wei Cao$^{1}$\footnotemark[1] \,\footnotemark[3] \quad
    Chang Luo$^{1}$\footnotemark[1] \quad
    Biao Zhang$^{2}$\quad
    Matthias Nießner$^{1}$ \quad
    Jiapeng Tang$^{1}$\footnotemark[2] \vspace{0.4em} \\
    {\normalsize $^1$Technical University of Munich} \quad
    {\normalsize $^2$King Abdullah University of Science and Technology} \quad \\
    \texttt{\url{https://vveicao.github.io/projects/Motion2VecSets}}
    \vspace*{-4mm}
}

\begin{document}

\twocolumn[{%
	\renewcommand\twocolumn[1][]{#1}%
	\maketitle
	\thispagestyle{empty}
	\begin{center}
		\includegraphics[width=\textwidth]{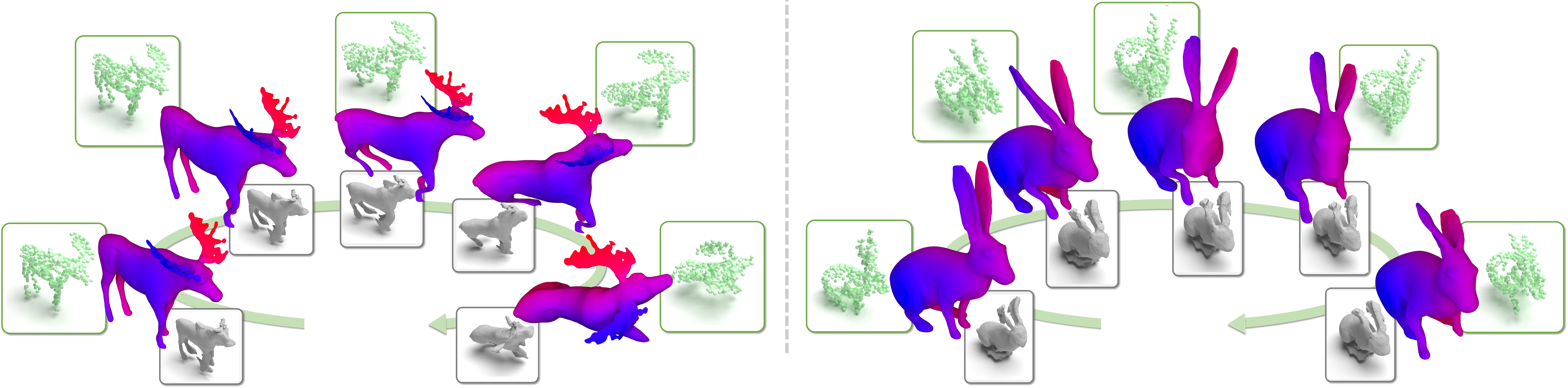}
            \captionof{figure}{
            We present \shadetext[left color=blue, right color=purple, middle color=violet, shading angle=90]{\textbf{Motion2VecSets}}, a 4D diffusion model for dynamic surface reconstruction from sparse, noisy, or partial \textcolor[RGB]{108, 198, 111}{point cloud sequences}. Compared to the existing state-of-the-art method \textcolor[RGB]{128, 128, 128}{CaDeX} \cite{Lei2022CaDeX}, our method can reconstruct more plausible non-rigid object surfaces with complicated structures and achieve more robust motion tracking. }
            \label{fig:teaser}
	\end{center}
}]

\maketitle

% Change footnote symbols
\renewcommand{\thefootnote}{\fnsymbol{footnote}}

\footnotetext[1]{Equal Contribution.}
\footnotetext[2]{Corresponding author.}
\footnotetext[3]{Work done during master's thesis.}

\begin{abstract}
\vspace*{-2mm}
We introduce Motion2VecSets, a 4D diffusion model for dynamic surface reconstruction from point cloud sequences. 
While existing state-of-the-art methods have demonstrated success in reconstructing non-rigid objects using neural field representations, conventional feed-forward networks encounter challenges with ambiguous observations from noisy, partial, or sparse point clouds.
To address these challenges,  we introduce a diffusion model that explicitly learns the shape and motion distribution of non-rigid objects through an iterative denoising process of compressed latent representations. The diffusion-based priors enable more plausible and probabilistic reconstructions when handling ambiguous inputs.
We parameterize 4D dynamics with latent sets instead of using global latent codes. This novel 4D representation allows us to learn local shape and deformation patterns, leading to more accurate non-linear motion capture and significantly improving generalizability to unseen motions and identities.
For more temporally-coherent object tracking, we synchronously denoise deformation latent sets and exchange information across multiple frames.
To avoid computational overhead, we designed an interleaved space and time attention block to alternately aggregate deformation latents along spatial and temporal domains.  
Extensive comparisons against state-of-the-art methods demonstrate the superiority of our Motion2VecSets in 4D reconstruction from various imperfect observations.
%, notably achieving a \red{19\%} improvement in Intersection over Union (IoU) compared to CaDeX \cite{Lei2022CaDeX} for reconstructing unseen individuals from sparse point clouds on the DeformingThings4D-Animals dataset \cite{DeformingThings4D}.
\vspace*{-2mm}
\end{abstract}
    
% \vspace*{-3mm}
\section{Introduction}
\label{sec:intro}
\vspace*{-1mm}
Our world, dynamic in its 4D nature, demands an increasingly sophisticated understanding and simulation of our living environment. This offers significant potential for practical applications, including Virtual Reality (VR), Augmented Reality (AR), and robotic simulations. 
There have been notable advances in 3D object modeling, particularly in representations through parametric models \cite{SMPL, STAR, MANO, FLAME, SMAL}. 
Unfortunately, these template-based models are not effectively suited to capture the 4D dynamics of general non-rigid objects, due to the assumption of a fixed template mesh. Model-free approaches \cite{Occupancy_Networks, LPDC, Lei2022CaDeX} represent a significant advance by using coordinate-MLP representations for deformable object reconstruction with arbitrary topologies and non-unified structures. However, these state-of-the-art methods still encounter challenges when facing ambiguous observations of noisy, sparse, or partial point clouds since it is an ill-posed problem where multiple possible reconstructions can match the input.
In addition, they represent dynamics as a sequence of single latent codes and thus struggle to capture shape and motion priors accurately. These issues become even more severe with unseen identities due to the limited generalizability of global latent representation.

To address the above-mentioned challenges, we propose Motion2VecSets, a diffusion model designed for 4D dynamic surface reconstruction from sparse, noisy, or partial point clouds.
It explicitly learns the joint distribution of non-rigid object surfaces and temporal dynamics through an iterative denoising process of compressed latent representations. This enables more realistic and varied reconstructions, particularly when dealing with ambiguous inputs.
Inspired by the observation that objects with varying topologies often share similar local geometry and deformation patterns, we represent dynamic surfaces as a sequence of latent sets to preserve local shape and deformation details:
one for shape modeling of the initial frame and others for describing the temporal evolution from the initial frame. This latent set representation naturally enables the learning of more accurate shape and motion priors, enhancing the model's generalization capacity to unseen identities and motions.
Specifically, we introduce the Synchronized Deformation Vector Set Diffusion, which simultaneously denoises the deformation latent sets across different time frames to enforce spatio-temporal consistency over dynamic surfaces. To manage the memory consumption associated with multiple deformation latent set diffusions, we design an interleaved space and time attention block as the basic unit for the denoiser. These blocks aggregate deformation latent sets along spatial and temporal domains alternately. As illustrated in Figure~\ref{fig:teaser}, our Motion2VecSets can reconstruct more plausible non-rigid object surfaces with complicated structures and achieve more robust motion tracking than the state-of-the-art method. %\\
Our contributions can be summarized as follows:
\begin{itemize}
    \item We present a 4D latent diffusion model designed for dynamic surface reconstruction, adept at handling sparse, partial, and noisy point clouds.
    \item We introduce a 4D neural representation with latent sets, significantly enhancing the capacity to represent complicated shapes as well as motions and improving generalizability to unseen identities and motions.
    \item We design an Interleaved Spatio-Temporal Attention mechanism for synchronized diffusion of deformation latent sets, achieving robust spatio-temporal consistency and advanced computational efficiency.
\end{itemize}
Extensive comparisons against
state-of-the-art methods demonstrate the superiority of our
Motion2VecSets in dynamic surface reconstruction on the Dynamic FAUST \cite{DFAUST} and the DeformingThings4D-Animals \cite{DeformingThings4D} datasets.
\vspace*{-2mm}
\section{Related works}
\label{sec:related_works}
\vspace*{-1mm}
\paragraph{3D Shapes} Traditional methods in 3D representation have primarily used meshes \cite{Pixel2Mesh, AtlasNet, DeepMarchingCube, Tang_2019_CVPR, pan2019deep,tang2021skeletonnet}, point clouds \cite{fan2017point, PointCloudMethod2, PointTransformer}, and voxels \cite{VoxelMethod2,3DR2N2, VoxelMethod3, Stutz2018CVPR, Riegler2017OctNet} to represent geometry. Complementing these are parametric models, which have effectively modeled specific shape categories, such as human bodies (e.g., SMPL \cite{SMPL}, STAR \cite{STAR}), faces (e.g., FLAME \cite{FLAME}), hands (e.g., MANO \cite{MANO}), and animals (e.g., SMAL \cite{SMAL}). However, these parametric approaches often rely on fixed templates, which can result in difficulties accurately modeling general non-rigid objects without consistent topological structures. Meanwhile, recent advancements in 3D representation are increasingly using implicit methods \cite{chen2018implicit_decoder, icml2020_2086, Occupancy_Networks, DeepSDF, IFNet, chabra2020deep, PatchNets, Shape2Vecset, Gridpull,tang2021sa,zhang20223dilg}, known for their greater flexibility to represent objects with arbitrary topologies. In particular, Occupancy Networks \cite{Occupancy_Networks} and DeepSDF \cite{DeepSDF} employ a continuous implicit framework, enabling the representation of volumetric grids offering potentially infinite resolution.
\vspace*{-3mm}
\paragraph{4D Dynamics} Recent advancements have successfully extended 3D representations to 4D, which more effectively captures object dynamics \cite{NPMS, OccupancyFlow, bozic2021neuraldeformationgraphs, LCRODE, LPDC, Lei2022CaDeX, fan2017point, tang2022neural, lei2023nap, singer2023text, zou20234d}. For example, OFlow \cite{OccupancyFlow} incorporates Neural-ODE \cite{NeuralODE} for simulating deformations. LPDC \cite{LPDC} replaces Neural-ODE with an MLP and learns local spatio-temporal codes, capturing both shape and deformations. CaDeX \cite{Lei2022CaDeX} employs a learnable deformation embedding between each frame and its canonical shape. However, methods relying on either ODE \cite{OccupancyFlow} solvers or a single global latent vector \cite{LPDC,Lei2022CaDeX,NPMS} coupled with an MLP network face challenges in capturing complex real-world 4D dynamics, particularly in non-rigid objects. 
Drawing inspiration from 3DShape2Vecset \cite{Shape2Vecset}, which uses a set of latent codes to represent similar local geometry patterns across objects, our proposed method leveragesa similar approach for characterizing 4D dynamics. Different objects share similar local deformation patterns, our framework uniquely assigns a distinct learnable latent code to each local region, significantly enhancing their ability to precisely model and generalize to unseen identities and motions. 
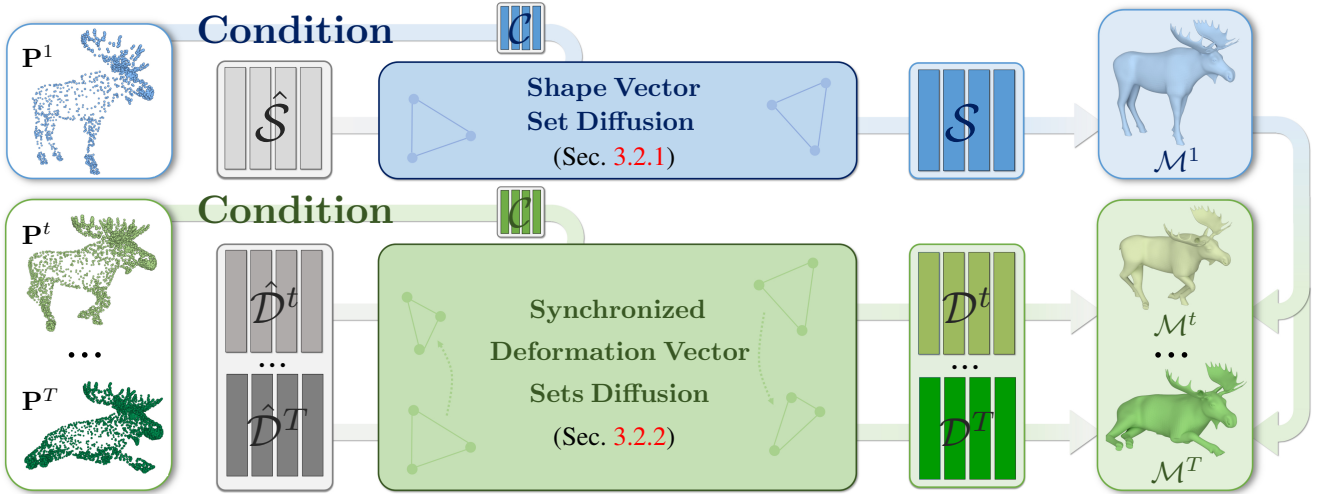
\begin{figure*}[t]
  \centering
  \begin{overpic}[width=\linewidth]{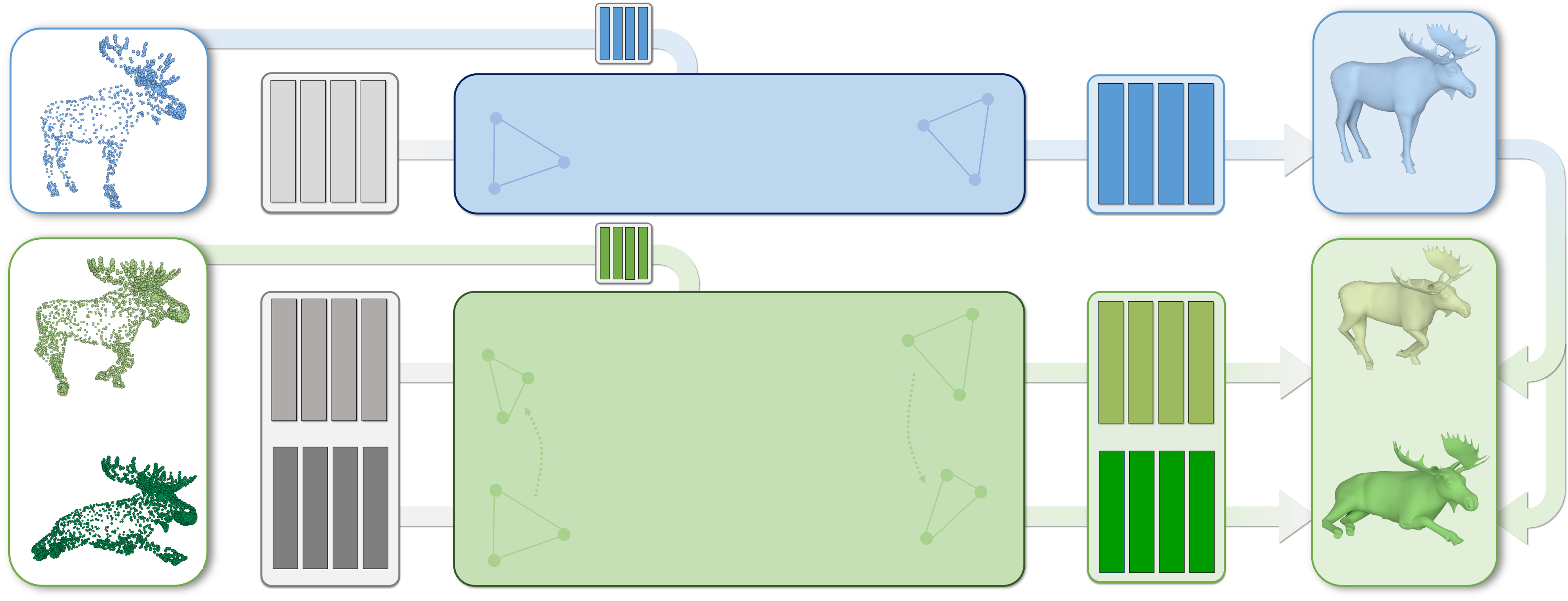}
     \put (9,165) {\textcolor{black}{$\mathbf{P}^1$}}
     \put (9,98) {\textcolor{black}{$\mathbf{P}^t$}}
     \put (9,35) {\textcolor{black}{$\mathbf{P}^{T}$}}
     \put (25,51) {$\scalebox{2}{\ensuremath{\cdot}}$}
     \put (30,51) {$\scalebox{2}{\ensuremath{\cdot}}$}
     \put (35,51) {$\scalebox{2}{\ensuremath{\cdot}}$}
        
     \put (75, 174) {${\textcolor[RGB]{0, 32, 96}{\scalebox{1.5}{\ensuremath{\mathbf{{Condition}}}}}}$}
     \put (193,174) {$\textcolor[RGB]{0, 32, 96}{\scalebox{1.6}{\ensuremath{\mathcal{C}}}}$}
     \put (97,137) {$\textcolor[RGB]{38, 38, 38}{\scalebox{2}{\ensuremath{\hat{\mathcal{S}}}}}$}
     
     \put (200, 154) {$\textcolor[RGB]{0, 32, 96}{{\scalebox{1.0}{\ensuremath{\mathbf{Shape \text{ } Vector}}}}}$}
     \put (200, 141) {$\textcolor[RGB]{0, 32, 96}{\scalebox{1.0}{\ensuremath{\mathbf{Set \text{ } Diffusion}}}}$}
     \put (210, 128) {(Sec.~\ref{me:shape_diff})}
     
     \put (358,137) {$\textcolor[RGB]{0, 32, 96}{\scalebox{2}{\ensuremath{{\mathcal{S}}}}}$}
     \put (437,125.5) {$\textcolor[RGB]{0, 32, 96}{\scalebox{1.1}{\ensuremath{{\mathcal{M}}^1}}}$}

     \put (75, 105.8) {${\textcolor[RGB]{56, 87, 35}{\scalebox{1.5}{\ensuremath{\mathbf{{Condition}}}}}}$}
      \put (193,104.5) {$\textcolor[RGB]{56, 87, 35}{\scalebox{1.6}{\ensuremath{\mathcal{C}}}}$}
      \put (95.2,68) {$\textcolor[RGB]{38, 38, 38}{\scalebox{1.7}{\ensuremath{\hat{\mathcal{D}}^t}}}$}
      \put (94,21.5) {$\textcolor[RGB]{38, 38, 38}{\scalebox{1.7}{\ensuremath{\hat{\mathcal{D}}^{T}}}}$}

       \put (201,70) {$\textcolor[RGB]{56, 87, 35}{{\scalebox{1.0}{\ensuremath{\mathbf{Synchronized}}}}}$}
       \put (186, 54) {$\textcolor[RGB]{56, 87, 35}{{\scalebox{1.0}{\ensuremath{\mathbf{Deformation \text{ } Vector}}}}}$}
       \put (201, 38) {$\textcolor[RGB]{56, 87, 35}{{\scalebox{1.0}{\ensuremath{\mathbf{Sets \text{ } Diffusion}}}}}$}
       \put (210, 22) {(Sec.~\ref{me:deform_diff})}

      \put (357,68) {$\textcolor[RGB]{38, 38, 38}{\scalebox{1.7}{\ensuremath{{\mathcal{D}}^t}}}$}
      \put (356,21.5) {$\textcolor[RGB]{38, 38, 38}{\scalebox{1.53}{\ensuremath{{\mathcal{D}}^{T}}}}$}

     \put (437,63.5) {$\textcolor[RGB]{56, 87, 35}{\scalebox{1.1}{\ensuremath{{\mathcal{M}}^t}}}$}
     \put (436,8) {$\textcolor[RGB]{56, 87, 35}{\scalebox{1.1}{\ensuremath{{\mathcal{M}}^{T}}}}$}

     \put (97.7,48.4) {$\scalebox{1.5}{\ensuremath{\cdot}}$}
     \put (101.7,48.4) {$\scalebox{1.5}{\ensuremath{\cdot}}$}
     \put (105.7,48.4) {$\scalebox{1.5}{\ensuremath{\cdot}}$}

     \put (359.7,47.8) {$\scalebox{1.5}{\ensuremath{\cdot}}$}
     \put (363.7,47.8) {$\scalebox{1.5}{\ensuremath{\cdot}}$}
     \put (367.7,47.8) {$\scalebox{1.5}{\ensuremath{\cdot}}$}
     
   \put (436,51) {$\scalebox{2}{\ensuremath{\cdot}}$}
     \put (441,51) {$\scalebox{2}{\ensuremath{\cdot}}$}
     \put (446,51) {$\scalebox{2}{\ensuremath{\cdot}}$}
\end{overpic}
  
\caption{\textbf{Overview Pipeline of Motion2VecSets.} 
Given a sequence of sparse and noisy point clouds as inputs$\{\mathbf{P}^t\}_{t=1}^{T}$, Motion2VecSets outputs a continuous mesh sequence $\{\mathcal{M}^t\}_{t=1}^{T}$. The initial input frame $\mathbf{P}^1$ (top left) is used as a condition in the \textcolor[RGB]{0, 32, 96}{Shape Vector Set Diffusion}, yielding denoised shape codes $\mathcal{S}$ for reconstructing the geometry of the reference frame $\mathcal{M}^1$ (top right). Concurrently, the subsequent input frames $\{\mathbf{P}^t\}_{t=2}^{T}$ (bottom left) are utilized in the \textcolor[RGB]{56, 87, 35}{Synchronized Deformation Vector Sets Diffusion} to produce denoised deformation codes $\{\mathcal{D}^t\}_{t=2}^{T}$, where each latent set $\mathcal{D}^t$ encodes the deformation from the reference frame $\mathcal{M}^1$ to subsequent frames $\mathcal{M}^t$.
}
  \label{fig:onecol}
\end{figure*}
\vspace*{-3mm}
\paragraph{Diffusion Models} 
Diffusion models \cite{ho2020denoising}, known for their Markov chain-based denoising capability, have made impressive progress in multiple tasks, including image and video processing \cite{chen2020wavegrad, nichol2021glide, dhariwal2021diffusion, DiffusionVideo, mittal2021symbolic}, 3D vision \cite{PointCloudDiffusion,rombach2022high, blattmann2022retrieval,DiffusionSDF,holmquist2023diffpose,tevet2022human, tang2024diffuscene,tang2024dphms, zhang2023functional}. These models are adept at capturing complex data distributions. In the field of 3D vision, their applications are varied: Luo et al. \cite{PointCloudDiffusion} have successfully applied diffusion models to point cloud generation, and Rombach \etal \cite{rombach2022high, blattmann2022retrieval} have adapted them for latent space representations. Additionally, integration with PointNet \cite{PointNet} and triplane features \cite{Triplane}, as seen in DiffusionSDF \cite{DiffusionSDF}, has further enhanced their training capabilities. Concurrent work NAP \cite{lei2023nap} advances 3D object generation by effectively modeling articulated objects with a novel parameterization and diffusion-denoising approach. A key challenge in representing 4D dynamics with existing diffusion models is their tendency to adapt 3D models directly to 4D and process each frame independently, which can result in discontinuities in temporal and spatial relationships. To bridge this gap, our approach implements synchronous denoising processes for sets of codes. This innovation ensures not only a reduction in spatial complexity but also consistent deformations in latent space. Moreover, recent works \cite{holmquist2023diffpose,tevet2022human,dabral2023mofusion,yuan2023physdiff,Shi_2023_ICCV,ren2023diffusion,zhang2024motiondiffuse} in the field of 3D pose estimation and generation also indicate the power of diffusion models. DiffPose \cite{holmquist2023diffpose} utilizes the diffusion model to handle very ambiguous poses and can even predict an infinite number of poses. PhysDiff \cite{yuan2023physdiff} produces physically plausible motions by incorporating a physics-based motion projection within its diffusion process. However, these methods are still in the realm of pose, our method expands the application of diffusion models to a broader range of deformable surfaces of general non-rigid objects. Similar to the concurrent work DPHMs~\cite{tang2024dphms}, our approach utilizes diffusion priors to facilitate robust 4D reconstruction from imperfect observations.
% \vspace*{-2mm}
\section{Approach}
\label{sec:method}
\vspace*{-1mm}
The inputs are $T$ frames of sparse, partial, or noisy point clouds, represented by $\mathcal{P}=\left\{\mathbf{P}^t\right\}_{t=1}^{T}$, where $\mathbf{P}^t=\left\{\mathbf{p}_i \in \mathbb{R}^3\right\}_{i=1}^L$, $L$ represents the number of points. The goal is to reconstruct continuous 3D meshes with high fidelity, denoted as $\{\mathcal{M}^t\}_{t=1}^{T} = \{\mathcal{V}^t, \mathcal{F}^t\}_{t=1}^{T}$, where $\mathcal{V}^t$ and $\mathcal{F}^t$ refer to the set of vertices faces of the reconstructed mesh at time frame $t$.
Conventional feed-forward models face challenges in handling ambiguous inputs within an ill-posed problem setting. Particularly, when observations are sparse, partial, and noisy, generating meaningful reconstructions becomes highly challenging without robust prior knowledge.
To reconstruct high-fidelity dynamic shapes accurately, we proposed 4D latent set diffusion to learn shape and motion priors, explicitly learning the distribution of deformable object surface sequences via compressed latent vector sets. 
While the diffusion model enhances realistic surface reconstruction and deformation tracking, generating multi-modal outputs, the latent set representation and transformer architecture provide the capability to capture more accurate geometry and deformation priors.
\subsection{4D Neural Representation with Latent Sets}\label{sec:4d-rep}
\vspace*{-1mm}
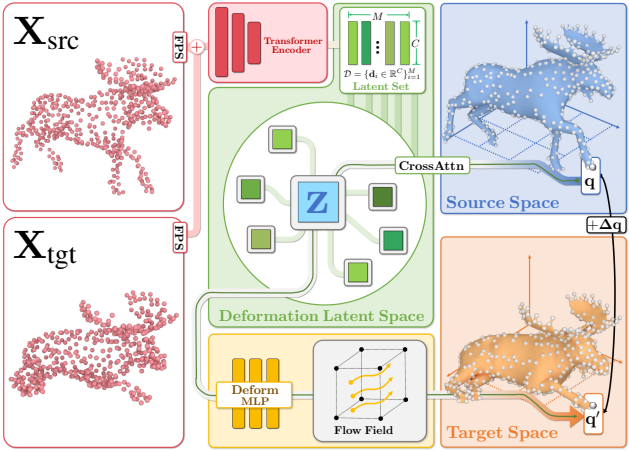
\begin{figure}[t]
  \centering
  \begin{overpic}[width=\linewidth]{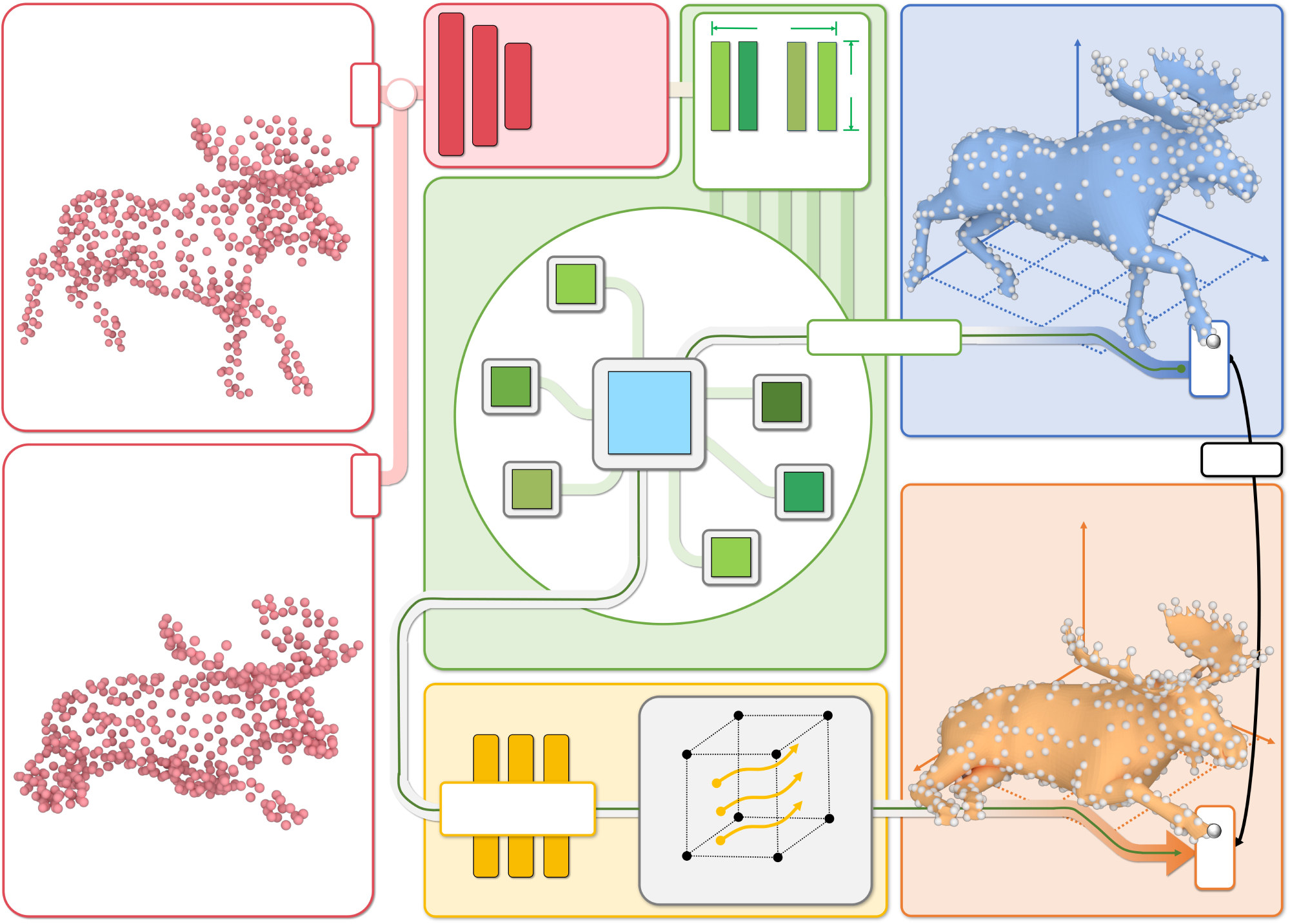}
  \put (4, 73) {$\scalebox{1.5}{\ensuremath{\mathbf{X}_{\text{tgt}}}}$}
  \put (4, 154) {$\scalebox{1.5}{\ensuremath{\mathbf{X}_{\text{src}}}}$}
  \put (71.4, 151) {$\textcolor[RGB]{192,0,0}{\scalebox{0.58}{\ensuremath{\mathrm{+}}}}$}

  \put (140.7, 152) {$\scalebox{1}{\ensuremath{\cdot}}$}
  \put (140.7, 150) {$\scalebox{1}{\ensuremath{\cdot}}$}
  \put (140.7, 148) {$\scalebox{1}{\ensuremath{\cdot}}$}

  \put (65.5, 157.4) {$\scalebox{0.45}{\rotatebox{-90}{\ensuremath{\mathbf{FPS}}}}$}
  \put (65.5, 85.4) {$\scalebox{0.45}{\rotatebox{-90}{\ensuremath{\mathbf{FPS}}}}$}

  \put (99, 154) {$\textcolor[RGB]{192, 0, 0}{\scalebox{0.36}{\ensuremath{\mathbf{Transformer}}}}$}
  \put (102, 150.5) {$\textcolor[RGB]{192, 0, 0}{\scalebox{0.36}{\ensuremath{\mathbf{Encoder}}}}$}

  \put (140, 163) {$\scalebox{0.4}{\ensuremath{M}}$}
  \put (154.8, 152.65) {$\scalebox{0.4}{\ensuremath{C}}$}
  
  \put (129.3, 141.5) {$\textcolor[RGB]{0, 0, 0}{\scalebox{0.36}{\ensuremath{\mathcal{D} = \{\mathbf{d}_i \in \mathbb{R}^C\}_{i=1}^{M}}}}$}
  \put (132.6, 136.5) {$\textcolor[RGB]{112, 173, 71}{\scalebox{0.42}{\ensuremath{\mathbf{Latent \text{ } Set}}}}$}

  \put (150, 106.2) {$\textcolor[RGB]{0, 0, 0}{\scalebox{0.48}{\ensuremath{\mathbf{CrossAttn}}}}$}
  
  \put (114.7, 89.8) {$\textcolor[RGB]{68, 114, 196}{\scalebox{1.8}{\ensuremath{\mathbf{z}}}}$}
  
  \put (82, 48.9) {$\textcolor[RGB]{112, 173, 71}{\scalebox{0.6}{\ensuremath{\mathbf{Deformation \text{ } Latent \text{ } Space}}}}$}
  
  \put (86.7, 21.5) {$\textcolor[RGB]{127, 75, 0}{\scalebox{0.48}{\ensuremath{\mathbf{Deform}}}}$}
  \put (90.4, 17.3) {$\textcolor[RGB]{127, 75, 0}{\scalebox{0.42}{\ensuremath{\mathbf{MLP}}}}$}
  \put (125.6, 6) {${\scalebox{0.43}{\ensuremath{\mathbf{Flow \text{ } Field}}}}$}
  
  \put (168, 91.9) {$\textcolor[RGB]{68,114,196}{\scalebox{0.66}{\ensuremath{\mathbf{Source \text{ } Space}}}}$}
  \put (168, 4.5) {$\textcolor[RGB]{237,125,49}{\scalebox{0.66}{\ensuremath{\mathbf{Target \text{ } Space}}}}$}
  
  \put (220.3, 100) {$\scalebox{0.66}{\ensuremath{\mathbf{q}}}$}
  \put (220.6, 9.8) {$\scalebox{0.66}{\ensuremath{\mathbf{q'}}}$}
  \put (221.3, 83.8) {$\scalebox{0.56}{\ensuremath{\mathbf{+\Delta q}}}$}
  \end{overpic}
\vspace*{-5mm}
  \caption{\textbf{Deformation Autoencoder.} {Given a pair of point clouds $\mathbf{X}_{\text{src}}$ and $\mathbf{X}_{\text{tgt}}$ from two frames of a dynamic mesh sequence, we initially downsample them using farthest point sampling (FPS). Subsequently, the concatenated points are passed into \textcolor[RGB]{192,0,0}{transformer encoder} to generate the \textcolor[RGB]{122, 173, 71}{Deformation Latent Set $\mathcal{D}$}. For a query point $\mathbf{q}$ in the source space, a cross-attention layer is utilized to match the most relevant \textcolor[RGB]{68, 114, 196}{fused feature $\mathbf{z}$}. This selected feature is subsequently fed into the \textcolor[RGB]{127, 75, 0}{deformation MLP decoder} to predict an offset $\mathbf{\Delta\mathbf{q}}$, translating it to $\mathbf{q'}$ in the target space. To reduce the feature diversity of $\mathcal{D}$, KL-regularization is employed.}}
  \label{fig:deformation_ae}
\vspace*{-5mm}
\end{figure}
Previous works often utilize single global codes \cite{OccupancyFlow,LPDC,Lei2022CaDeX} to represent 4D sequences, potentially losing significant surface geometry and temporal evolution information. To retain as much detail as possible, we advocate the use of two distinct sets of latent vectors. 
Specifically, the \textit{shape latent set} is responsible for reconstructing the initial frame, serving as the reference frame, while the deformation correspondences between the reference and subsequent frames are encoded by the \textit{deformation latent set}. 
Compared with previous methods \cite{OccupancyFlow, LPDC, Lei2022CaDeX} relying on a single global code, we assign local latent codes to individual local regions, which significantly improves the network's capability to accurately model non-linear motions and generalize to unseen identities and motions.
Given that different non-rigid objects share similar local geometry and deformation patterns, the latent sets can also increase the generalization ability to handle unseen motions and identities. 
\vspace*{-5mm}
\paragraph{Shape Latent Set} 
Similar to 3DShape2VecSet~\cite{Shape2Vecset}, we utilize a shape autoencoder to compress the surface of the initial frame into a set of latent codes. Concretely, we leverage a transformer encoder that condenses the 3D surface of the initial frame into a set of latent vectors denoted as \( \mathcal{S} = \{\mathbf{s}_i \in \mathbb{R}^C\}_{i=1}^M \). Here, \( M \) represents the overall count of codes and \( C \) denotes their dimensionality. Following that, a cross-attention layer is used to fuse the latent codes for occupancy field prediction through an MLP. Training involves minimizing the binary cross-entropy (BCE) loss, which aligns the predicted occupancy \( \hat{\mathcal{O}}(\mathcal{Q}) \) with the actual occupancy \( \mathcal{O}(\mathcal{Q}) \), $\mathcal{Q}$ refers to the query points:
%\vspace*{-1mm}
\begin{equation}
\mathcal{L}_{\text{recon}}\left(\mathcal{S},\mathcal{Q}\right)=\mathbb{E}_{\mathcal{Q} \in \mathbb{R}^3}[\operatorname{BCE}(\hat{\mathcal{{O}}}, \mathcal{O})]
\label{eq:shape_codes}
\vspace*{-2mm}
\end{equation}
\vspace*{-7mm}
\paragraph{Deformation Latent Set}
As shown in Figure~\ref{fig:deformation_ae}, to represent the deformation between different non-rigid poses, we first sample a pair of point clouds \( \mathbf{X}_{\text{src}} \) and \( \mathbf{X}_{\text{tgt}} \) of size $N$ with same sampling indices from a mesh sequence.
Then, we employ a uniform farthest point sampling (FPS) strategy to eliminate spatial redundancy while preserving point correspondence. This process facilitates a concatenation step, where we combine the original and downsampled pairs of point clouds 
$\{\mathbf{X}_{\text{src}}, \mathbf{X}_{\text{tgt}}\}$,  respectively. The subsequent transformer encoder is applied to extract deformation details in the local regions around subsampled points, resulting in the deformation latent set denoted as $\mathcal{D} = \{ \mathbf{d}_i \in \mathbb{R}^C \}_{i=1}^M$.
Query point $\mathbf{q} \in \mathcal{Q}_{src} $ from the source space is utilized as the query for cross-attention, extracting the most relevant fused feature \( \mathbf{z} \) in the deformation latent space. A linear deformation layer then maps these features to the predicted target points $\mathbf{q'}$ through a flow field. The correspondence loss calculates the \( \ell_2 \)-norm distance between the predicted and true target point clouds:
\vspace*{-1mm}
\begin{equation}
\mathcal{L}_{\text{corr}}\left(\mathcal{D}, \mathcal{Q}_{\text{src}}\right) = \mathbb{E}_{\mathcal{Q}_{\text{src}} \in \mathbb{R}^3}[\operatorname{MSE}(\hat{\mathcal{Q}}_{\text{tgt}}, \mathcal{Q}_{\text{tgt}})]
\end{equation}
\vspace*{-2mm}
\vspace*{-7mm}
\paragraph{KL Regularization} Consistent with the latent diffusion framework \cite{rombach2022high}, our model incorporates KL-regularization in the latent space to modulate feature diversity. This ensures the preservation of high-level features and keeps coherent global geometric and deformation patterns, which promotes the learning of diffusion models.
In summary, we characterize a sequence through the shape latent set $\mathcal{S}^1$ of the initial reference frame, which describes the implicit surface, and deformation latent sets ${\mathcal{D}^2, \mathcal{D}^3,...,\mathcal{D}^{T}}$ that depict the dense correspondences between the initial reference frame and the subsequent frames.
%
%\subsection{4D Latent Vector Set Diffusion}\label{sec:4d-diff}
\subsection{4D Latent Set Diffusion}\label{sec:4d-diff}
\vspace*{-1mm}
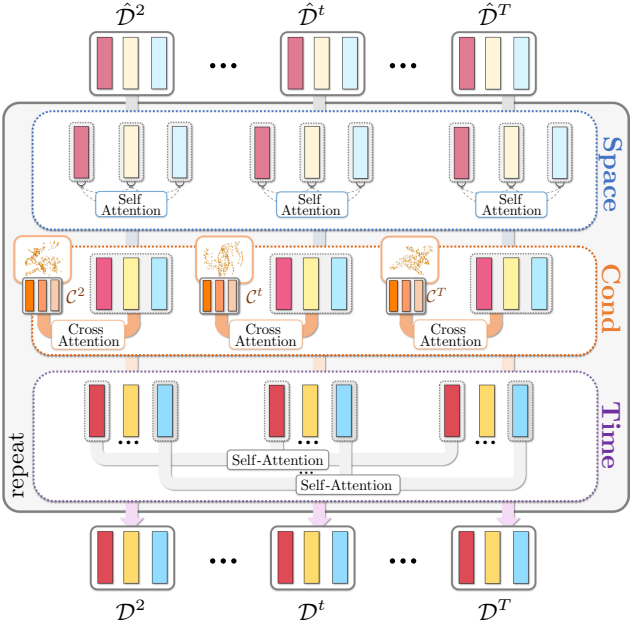
\begin{figure}[t]
  \centering
  \begin{overpic}[width=\linewidth]{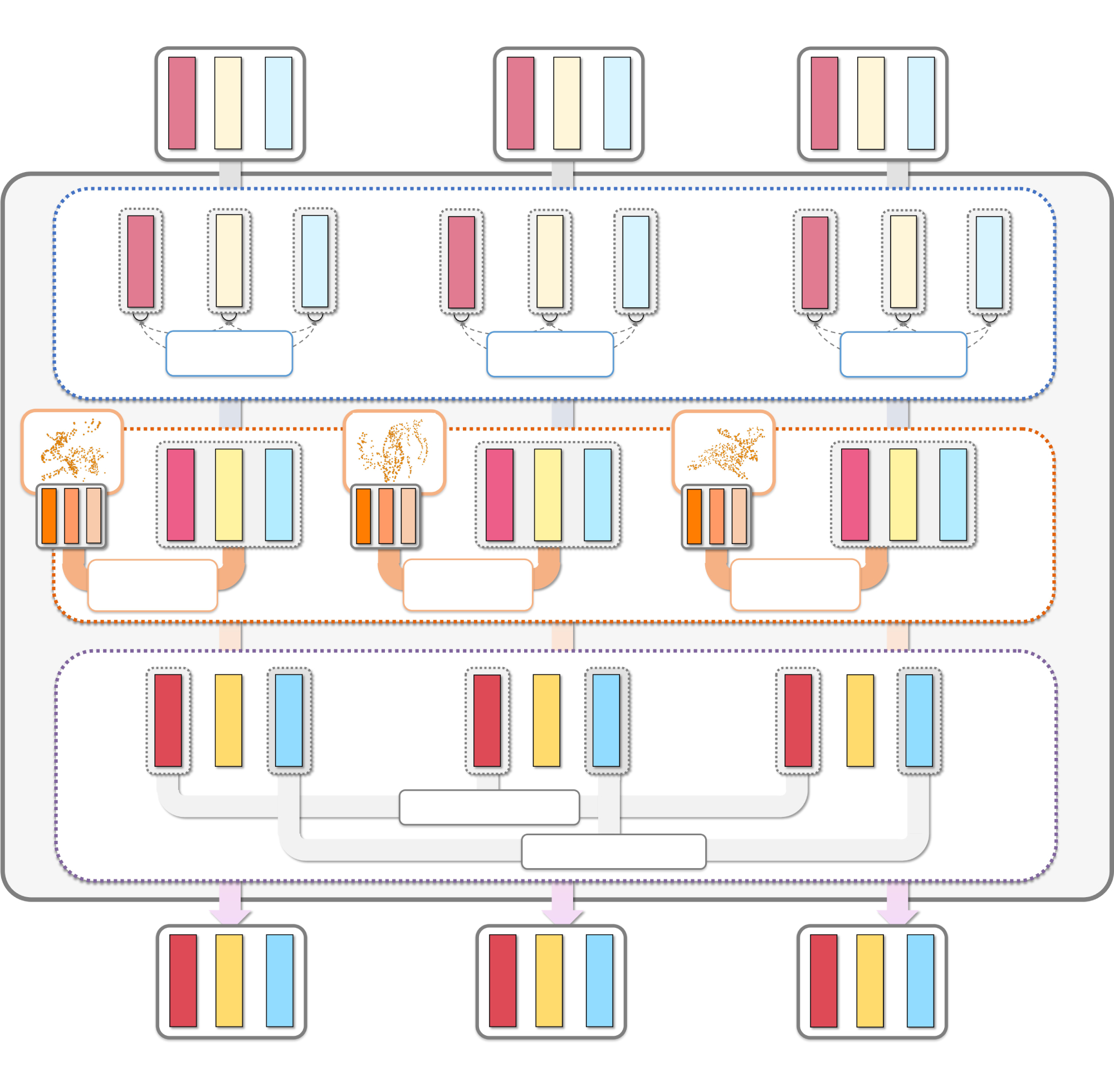}

    \put (43, 225) {\small$\hat{\mathcal{D}}^2$}
    \put (112, 225) {\small$\hat{\mathcal{D}}^t$}
    \put (180, 225) {\small$\hat{\mathcal{D}}^{T}$}

    \put (77, 205) {$\scalebox{1.9}{\ensuremath{\cdot}}$}
    \put (81, 205) {$\scalebox{1.9}{\ensuremath{\cdot}}$}
    \put (85, 205) {$\scalebox{1.9}{\ensuremath{\cdot}}$}

    \put (145, 205) {$\scalebox{1.9}{\ensuremath{\cdot}}$}
    \put (149, 205) {$\scalebox{1.9}{\ensuremath{\cdot}}$}
    \put (153, 205) {$\scalebox{1.9}{\ensuremath{\cdot}}$}

    \put (77, 17.5) {$\scalebox{1.9}{\ensuremath{\cdot}}$}
    \put (81, 17.5) {$\scalebox{1.9}{\ensuremath{\cdot}}$}
    \put (85, 17.5) {$\scalebox{1.9}{\ensuremath{\cdot}}$}

    \put (145, 17.5) {$\scalebox{1.9}{\ensuremath{\cdot}}$}
    \put (149, 17.5) {$\scalebox{1.9}{\ensuremath{\cdot}}$}
    \put (153, 17.5) {$\scalebox{1.9}{\ensuremath{\cdot}}$}

    \put (225.7, 183) {$\rotatebox{-90}{\textcolor[RGB]{68, 114, 196}{\scalebox{1}{\ensuremath{\mathbf{Space}}}}}$}
    \put (227, 135) {$\rotatebox{-90}{\textcolor[RGB]{237, 125, 49}{\scalebox{1}{\ensuremath{\mathbf{Cond}}}}}$}
    \put (227, 83) {$\rotatebox{-90}{\textcolor[RGB]{112, 48, 160}{\scalebox{1}{\ensuremath{\mathbf{Time}}}}}$}
 
    \put (44, 157.3) {$\scalebox{0.54}{\ensuremath{\mathrm{Self}}}$}
    \put (37.5, 153.3) {$\scalebox{0.54}{\ensuremath{\mathrm{Attention}}}$}

    \put (112, 157.3) {$\scalebox{0.54}{\ensuremath{\mathrm{Self}}}$}
    \put (105.5, 153.3) {$\scalebox{0.54}{\ensuremath{\mathrm{Attention}}}$}

    \put (188, 157.3) {$\scalebox{0.54}{\ensuremath{\mathrm{Self}}}$}
    \put (181.5, 153.3) {$\scalebox{0.54}{\ensuremath{\mathrm{Attention}}}$}

    \put (25, 108.35) {$\scalebox{0.56}{\ensuremath{\mathrm{Cross}}}$}
    \put (21, 103.5) {$\scalebox{0.56}{\ensuremath{\mathrm{Attention}}}$}

    \put (92, 108.35) {$\scalebox{0.56}{\ensuremath{\mathrm{Cross}}}$}
    \put (88, 103.5) {$\scalebox{0.56}{\ensuremath{\mathrm{Attention}}}$}

    \put (162, 108.35) {$\scalebox{0.56}{\ensuremath{\mathrm{Cross}}}$}
    \put (158, 103.5) {$\scalebox{0.56}{\ensuremath{\mathrm{Attention}}}$}

    \put (87, 58.5) {$\scalebox{0.56}{\ensuremath{\mathrm{Self\text{-}Attention}}}$}
    \put (113, 49) {$\scalebox{0.56}{\ensuremath{\mathrm{Self\text{-}Attention}}}$}
    
    \put (2.5, 46) {$\rotatebox{90}{\scalebox{0.91}{\ensuremath{\mathrm{repeat}}}}$}

    \put (43, 0) {\small${\mathcal{D}}^2$}
    \put (112, 0) {\small${\mathcal{D}}^t$}
    \put (180, 0) {\small${\mathcal{D}}^{T}$}

     \put (24.4,120) {$\textcolor[RGB]{132, 60, 12}{\scalebox{0.7}{\ensuremath{\mathcal{C}^2}}}$}
     \put (92.4,120) {$\textcolor[RGB]{132, 60, 12}{\scalebox{0.7}{\ensuremath{\mathcal{C}^t}}}$}
     \put (160.5,120.3) {$\textcolor[RGB]{132, 60, 12}{\scalebox{0.65}{\ensuremath{\mathcal{C}^{T}}}}$}
    
    \put (43, 64) {$\scalebox{1.3}{\ensuremath{\cdot}}$}
    \put (46, 64) {$\scalebox{1.3}{\ensuremath{\cdot}}$}
    \put (49, 64) {$\scalebox{1.3}{\ensuremath{\cdot}}$}

    \put (111, 64) {$\scalebox{1.3}{\ensuremath{\cdot}}$}
    \put (114, 64) {$\scalebox{1.3}{\ensuremath{\cdot}}$}
    \put (117, 64) {$\scalebox{1.3}{\ensuremath{\cdot}}$}

    \put (178, 64) {$\scalebox{1.3}{\ensuremath{\cdot}}$}
    \put (181, 64) {$\scalebox{1.3}{\ensuremath{\cdot}}$}
    \put (184, 64) {$\scalebox{1.3}{\ensuremath{\cdot}}$}
    
    \put (112, 53.8) {$\scalebox{0.8}{\ensuremath{\cdot}}$}
    \put (114, 53.8) {$\scalebox{0.8}{\ensuremath{\cdot}}$}
    \put (116, 53.8) {$\scalebox{0.8}{\ensuremath{\cdot}}$}
  \end{overpic}
  \caption{\textbf{Synchronized Deformation Vector Set Diffusion.}
  {Given noised deformation vector sets  $\{\hat{\mathcal{D}}^t\}_{t=2}^{T}$ (top) from a sequence, each set denoted as $ \hat{\mathcal{D}^{t}} = \{\hat{\mathbf{d}}_1^t,...,\hat{\mathbf{d}}_{M}^t \} $ of timestep $t \in [2,T]$, we use repeated Interleaved Spatio-Temporal Attention Blocks (ISTA) as our denoising network.  In each ISTA block, we first pass them to the space self-attention layer (\textcolor[RGB]{68, 114, 196}{Space Attention}) to aggregate latent features $ \hat{\mathcal{D}}^{t}$ across different spatial locations within each frame to explore spatial contexts.  Next, we inject conditional information extracted from sparse or partial point clouds via cross-attention (\textcolor[RGB]{237, 125, 49}{Condition Attention}) between conditional codes $ \mathcal{C}^t$ and noised deformation codes $ \hat{\mathcal{D}}^{t}$ at each frame. Lastly, to enhance temporal coherence, a time self-attention layer (\textcolor[RGB]{112, 48, 160}{Time Attention}) is used to aggregate latent codes from the same position but from different frames,~\ie $\{\hat{\mathbf{d}}_i^t\}_{t=2}^{T}$}. Repeat this ISTA block and we finally get denoised deformation latent sets $\{\mathcal{D}^t\}_{t=2}^{T}$ (bottom).
  Within each layer, different colored latents represent the dynamics of distinct local regions, while the same colored latents represent the dynamics of a local region at different time steps.
  }
\vspace*{-5.5mm}
  \label{fig:lsta}
\end{figure}
\subsubsection{Shape Diffusion}\label{me:shape_diff}
\vspace*{-1.5mm}
Following the diffusion paradigm in EDM by Karras et al. \cite{Karras2022edm}, we aim to minimize the expected \( \ell_2 \)-denoising error. This is achieved by adding the noise $\mathbf{\epsilon}$ sampled from the Gaussian distribution to the shape latent set $\mathcal{S}$, and then feeding the noise-added code $\hat{\mathcal{S}} = \mathcal{S}+\mathbf{\epsilon}$ to the denoiser (to avoid confusing, we also use $\mathcal{S}$ to represent its matrix form $\mathbb{R}^{N \times C}$). The whole process is denoted as:
\vspace*{-1mm}
\begin{equation}
\mathbb{E}_{\mathbf{\epsilon} \sim \mathcal{N}(0, \sigma^2 \mathbf{I})} \left\| \text{Denoiser}\left(\hat{\mathcal{S}}, \sigma, \mathcal{C} \right) - \mathcal{S} \right\|_2^2
\end{equation}
Here, \( \sigma \) represents the noise level. Conditional latent set $\mathcal{C}$ is defined as $ \mathcal{C}(\mathbf{P}^1) = \{c_i \in \mathbb{R}^C\}_{i=1}^M $, which is generated by sending the first input frame \( \mathbf{P}^1 \) to the  conditional encoder.
\vspace*{-7mm}
\subsubsection{Synchronized Deformation Diffusion}
\label{me:deform_diff}
\vspace*{-2mm}
To adapt these 3D models  \cite{STAR,FLAME,MANO,SMAL}  directly to 4D, the most straightforward approach is frame-by-frame processing, which may lead to discontinuities in temporal and spatial correspondence. Another approach is to aggregate all spatial-temporal point clouds, which would significantly increases the time complexity to \(O(T^2 N^2)\) for a sequence of \(T\) frames and \(N\) points each.
However, our 4D latent set representation allows us to bypass the need for brute-force attention across spatial and temporal domains. As discussed in Sec.~\ref{sec:4d-rep}, the deformation latents at identical spatial positions across different frames correspond to the deformation behaviors of the same local surface region. Leveraging this property, we implement an alternating aggregation approach for the latent features, systematically switching between the spatial and temporal domains. This method not only enhances efficiency but also preserves the accuracy of our model, leading to a reduction in time complexity to \(O(T N^2)\) in the spatial domain and \(O(N T^2)\) in the temporal domain. %cumulatively much lower than the \(O(T^2 N^2)\) observed in the previous approach. 
The details of synchronized deformation diffusion are described as follows.
Given sparse input point clouds \(\mathcal{P}=\left\{\mathbf{P}^t\right\}_{t=1}^{T}\), we pair subsequent frames with the first reference frame \( \mathbf{P}^1 \), \textit{i.e.}, \(\{ \mathbf{P}^1,\mathbf{P}^t \}_{t=2}^{T}\). These pairs are encoded  into  a series of conditional latents \(\mathcal{C}^t(\mathbf{P}^1, \mathbf{P}^t) = \{\mathbf{c}_i \in \mathbb{R}^C\}_{i=1}^M \) via a transformer encoder. Then these conditional latents, together with the diffused shape latent set  \(\mathcal{S}^1\) in Sec.~\ref{me:shape_diff}, are injected into the denoising network as the condition providing guidance for the network to handle ambiguous scenarios, like partial observation.
\vspace*{-7mm}
\paragraph{Interleaved Spatio-Temporal Attention}\label{method:ista}
Figure~\ref{fig:lsta} depicts the denoiser network of our proposed synchronized deformation latent set diffusion. The basic unit is the designed Interleaved Spatio-temporal Attention Block (ISTA). 
% within the synchronized deformation diffusion module, we introduce the Interleaved Spatio-Temporal Attention mechanism to leverage correspondence among different timesteps, effectively addressing the computational overhead.
Each ISTA block contains three attention layers: \textit{Space Self-Attention Layer},  \textit{Conditional Cross-Attention Layer} and \textit{Time Self-Attention Layer}.
The \textit{Space Self-Attention Layer} initiates spatial information exchange within each set of noised deformation codes $ \hat{\mathcal{D}^{t}} = \{\hat{\mathbf{d}}_i^t \}_{i=1}^{M} $:
%
% \vspace*{-2mm}
\begin{equation}
\mathbf{SpaceAttn}= \operatorname{SelfAttn}(\{\hat{\mathbf{d}}_i^t \}_{i=1}^{M})
\end{equation}

This is then followed by the \textit{Conditional Cross-Attention Layer}. Conditional codes $\mathcal{C}^t(\mathbf{P}^{1}, \mathbf{P}^{t}) = \{\mathbf{c}_i \in \mathbb{R}^C\}_{i=1}^M$ from a partial or sparse point cloud are subjected to cross-attention with $\mathbf{CondAttn}$:
% \vspace*{-2mm}
\begin{equation}
\mathbf{CondAttn} = \operatorname{CrossAttn}( \{\hat{\mathbf{d}}_i^t \}_{i=1}^{M},\mathcal{C}^t)
\end{equation}
Finally, to improve coherence in the time dimension, a \textit{Time Self-Attention Layer} is implemented among deformation codes from different timesteps but from the same position (same index $i$ but different $t$). Consequently, through this setup, the $\mathbf{TimeAttn}$ is effectively obtained:
%
% \vspace*{-2mm}
\begin{equation}
\mathbf{TimeAttn}= \operatorname{SelfAttn}(\{\hat{\mathbf{d}}_i^t\}_{t=2}^{T})
\end{equation}

In the denoising phase, we regard the entire sequence of deformation codes as a unified entity and apply a uniform noise reduction strategy across all codes, which preserves the consistency of local deformation patterns. Contrary to assigning individual noise to each set of shape codes, we add a consistent uniform noise $\mathbf{\epsilon}$ to the deformation codes of the entire sequence $\{\hat{\mathcal{D}^t}\}_{t=2}^{T} = \{\mathcal{D}^t\}_{t=2}^{T} + \mathbf{\epsilon}$. The denoising objective is thus formulated as:
\begin{equation}
\mathbb{E}_{\mathbf{\epsilon} \sim \mathcal{N}(0, \sigma^2 \mathbf{I})} \left\| \text{Denoiser}\left(\{\hat{\mathcal{D}^t}\}_{t=2}^{T}, \sigma, \mathcal{C} \right) - \{\mathcal{D}^t\}_{t=2}^{T} \right\|_2^2 
\end{equation}
Here, $\mathcal{C}$ represents conditional codes derived from observations, $\{\mathcal{D}^t\}_{t=2}^{T}$ can also be represented in its 3D matrix form as $\mathbb{R}^{M \times (T-1) \times C}$. This approach not only ensures uniformity in the denoising process but also significantly reduces computational overhead.

\section{Experiments}
\label{sec:experiments}
\vspace*{4mm}
\begin{figure*}[t]
  \centering
  \begin{overpic}[width=\linewidth]{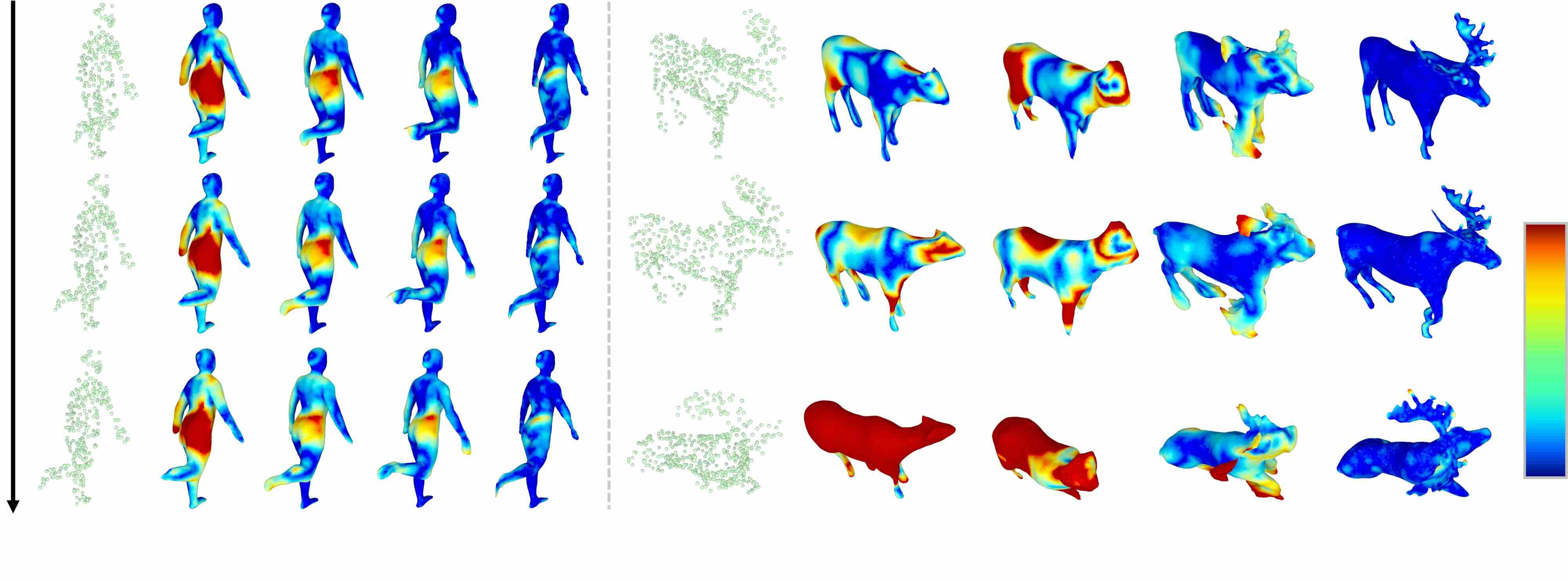}
  \put (2.3,12) {\large t}
  % \put (14,12) {Sparse} 
  \put (14,12) {Input}
  \put (49,12) {OFlow} 
  \put (84,12) {LPDC} 
  \put (118,12) {CaDeX} 
  \put (158,12) {Ours} 
  % \put (208,12) {Sparse} 
  \put (208,12) {Input}
  \put (265,12) {OFlow} 
  \put (320,12) {LPDC} 
  \put (374,12) {CaDeX} 
  \put (436,12) {Ours} 
  \put (487,23) {$0$} 
  \put (483.7,116.8) {$0.4$} 
  \end{overpic}
\vspace*{-8mm}
    \caption{Comparisons of 4D Shape Reconstruction from \textbf{sparse and noisy} point clouds on the D-FAUST \cite{DFAUST} (left) and the DT4D-A \cite{DeformingThings4D} (right) datasets. We visualize the Chamfer Distance between reconstruction and ground-truths as error maps. Our method can reconstruct more accurate surface geometries and motion dynamics. }
  \label{fig:res_quali_sparse}
  
\end{figure*}
\begin{figure*}[t]
  \centering
  \begin{overpic}[width=\linewidth]{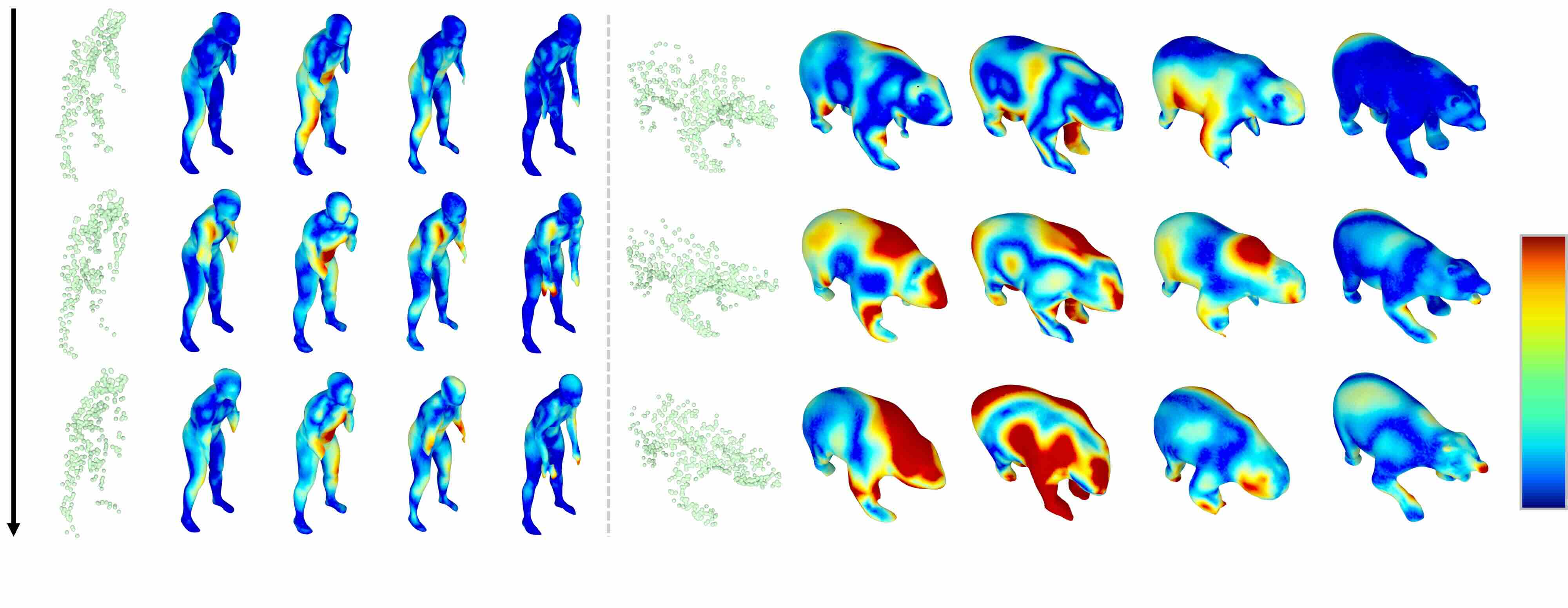}
\put (2.3,12) {\large t}
  % \put (16,12) {Partial} 
  \put (16,12) {Input}
  \put (53,12) {OFlow} 
  \put (88,12) {LPDC} 
  \put (122,12) {CaDeX} 
  \put (163,12) {Ours} 
  % \put (208,12) {Partial} 
  \put (208,12) {Input}
  \put (263,12) {OFlow} 
  \put (317,12) {LPDC} 
  \put (374,12) {CaDeX} 
  \put (435,12) {Ours} 
  \put (486.7,21.6) {$0$} 
  \put (482.7,121.5) {$0.4$} 
  \end{overpic}
\vspace*{-8mm}
  \caption{Comparisons of 4D Shape Completion from \textbf{monocular noisy depth scans} on D-FAUST \cite{DFAUST} (left) and DT4D-A \cite{DeformingThings4D} (right) datasets. Our method exhibits lower reconstruction errors and achieves more plausible tracking.}
  \label{fig:res_quali_partial}
\vspace*{-2.5mm}
\end{figure*}
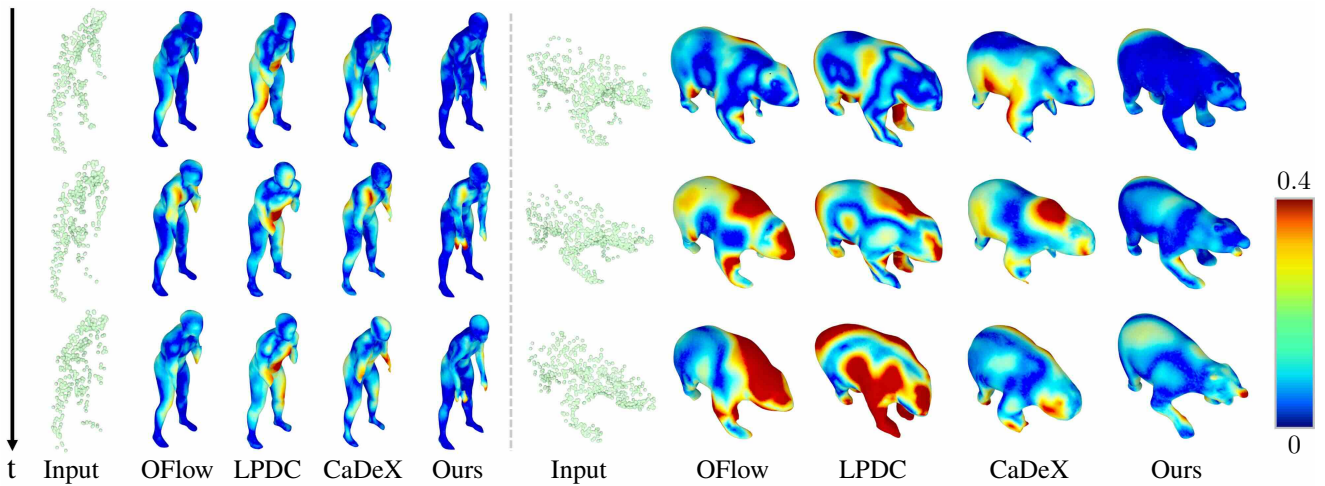

\vspace{-5mm}
\begin{table*}[ht]
    \centering
    \begin{minipage}{.5\linewidth}
        \centering
        \scalebox{0.73}{
        \begin{tabular}{@{}ccccc|ccc@{}}
        \toprule
        \multirow{2}{*}{Input} & \multirow{2}{*}{Method} & \multicolumn{3}{c|}{Unseen Motion} & \multicolumn{3}{c}{Unseen Individual} \\
        \cmidrule(l){3-8}
        & & IoU $\uparrow$ & CD $\downarrow$ & Corr $\downarrow$ & IoU $\uparrow$ & CD $\downarrow$ & Corr $\downarrow$ \\  \midrule
        
        \multirow{4}{*}{\makecell{DT4D-A \\ \cite{DeformingThings4D}}} & \multicolumn{1}{c|}{OFlow~\cite{OccupancyFlow}} & 70.6\% & 0.104 & 0.204 & 57.3\% & 0.175 & 0.285 \\
        & \multicolumn{1}{c|}{LPDC~\cite{LPDC}} & 72.4\% & 0.085 & 0.162 & 59.4\% & 0.149 & 0.262 \\
        & \multicolumn{1}{c|}{CaDex~\cite{Lei2022CaDeX}}  & 80.3\% & 0.061 & 0.133 & 64.7\% & 0.127 & 0.239 \\
        & \multicolumn{1}{c|}{Ours} & \textbf{88.9\%} & \textbf{0.050} & \textbf{0.061} & \textbf{83.7\%} & \textbf{0.058} & \textbf{0.074} \\  
        
        \midrule
        
        \multirow{4}{*}{\makecell{D-FAUST \\ \cite{DFAUST}}} & \multicolumn{1}{c|}{OFlow~\cite{OccupancyFlow}} & 81.5\% & 0.065 & 0.094 & 72.3\% & 0.084 & 0.117 \\
        & \multicolumn{1}{c|}{LPDC~\cite{LPDC}} & 84.9\% & 0.055 & 0.080 & 76.2\% & 0.071 & 0.098 \\
        & \multicolumn{1}{c|}{CaDex~\cite{Lei2022CaDeX}}  & 89.1\% & 0.039 & 0.070 & 80.7\% & 0.055 & 0.087 \\
        & \multicolumn{1}{c|}{Ours} & \textbf{90.7\%} & \textbf{0.033} & \textbf{0.047} & \textbf{83.7\%} & \textbf{0.045} & \textbf{0.064} \\  
        
        \bottomrule
        \end{tabular}}
        \captionsetup{width=.9\linewidth}
        \caption{Quantitative comparisons of 4D Shape Reconstruction from \textbf{sparse and noisy} point cloud sequences on the DT4D-A \cite{DeformingThings4D} and the D-FAUST \cite{DFAUST} datasets.}
        \label{tab:sparse}
        
    \end{minipage}%
    \hfill 
    \begin{minipage}{.5\linewidth}
        \centering
        \scalebox{0.72}{
        \begin{tabular}{@{}ccccc|ccc@{}}
        \toprule
        \multirow{2}{*}{Input} & \multirow{2}{*}{Method} & \multicolumn{3}{c|}{Unseen Motion} & \multicolumn{3}{c}{Unseen Individual} \\
        \cmidrule(l){3-8}
        & & IoU $\uparrow$ & CD $\downarrow$ & Corr $\downarrow$ & IoU $\uparrow$ & CD $\downarrow$ & Corr $\downarrow$ \\  \midrule
        
        \multirow{4}{*}{\makecell{DT4D-A \\ \cite{DeformingThings4D}}} & \multicolumn{1}{c|}{OFlow~\cite{OccupancyFlow}}   & 64.2\% & 0.305 & 0.423 & 55.1\% & 0.408 & 0.538 \\
        & \multicolumn{1}{c|}{LPDC~\cite{LPDC}} & 62.2\% & 0.339 & 0.427 & 51.6\% & 0.467 & 0.488 \\
        & \multicolumn{1}{c|}{CaDex~\cite{Lei2022CaDeX}}   & 70.8\% & 0.254 & 0.499 & 59.2\% & 0.379 & 0.498 \\
        & \multicolumn{1}{c|}{Ours}  & \textbf{73.3\%} & \textbf{0.177} & \textbf{0.404} & \textbf{66.3}\% & \textbf{0.193} & \textbf{0.438} \\ 
        
        \midrule
        
        \multirow{4}{*}{\makecell{D-FAUST \\ \cite{DFAUST}}} & \multicolumn{1}{c|}{OFlow~\cite{OccupancyFlow}}  & 76.9\% & 0.084 & 0.165 & 66.4\% & 0.109 & 0.194\\
        & \multicolumn{1}{c|}{{LPDC}~\cite{LPDC}} & 68.3\% & 0.138 & 0.167 & 59.6\% & 0.156 & 0.204 \\
        & \multicolumn{1}{c|}{CaDex~\cite{Lei2022CaDeX}}  & 80.7\% & 0.074 & 0.123 & 70.4\% & 0.096 & 0.157 \\
        & \multicolumn{1}{c|}{Ours} & \textbf{83.8\%} & \textbf{0.054} & \textbf{0.111} & \textbf{74.4\%} & \textbf{0.075} & \textbf{0.140} \\ 
        \bottomrule
        \end{tabular}}
        \captionsetup{width=.9\linewidth}
        \caption{Quantitative comparisons of 4D Shape Completion from \textbf{monocular noisy depth scans} on the DT4D-A \cite{DeformingThings4D} and the D-FAUST \cite{DFAUST} datasets.}
        \label{tab:partial}
    \end{minipage}
\vspace*{-6mm}
\end{table*}

\textbf{Datasets:} 
%To validate our model, 
We conducted experiments on two 4D datasets. The first, Dynamic FAUST (D-FAUST) \cite{DFAUST}, focuses on human body dynamics, including 10 subjects and 129 sequences. It is split into training (70\%), validation (10\%), and test (20\%) subsets, following \cite{OccupancyFlow}. The second, DeformingThings4D-Animals (DT4D-A)~\cite{DeformingThings4D}, includes 38 identities with a total of 1227 animations, divided into training (75\%), validation (7.5\%), and test (17.5\%) subsets as \cite{Lei2022CaDeX}. The training and validation sets use motion sequences of seen individuals. The test set is divided into two parts: unseen motions and unseen individuals.
\\
\textbf{Baselines:} We compare against state-of-the-art methods in 4D reconstruction, including OFlow~\cite{OccupancyFlow}, LPDC~\cite{LPDC}, CaDex~\cite{Lei2022CaDeX}.
% \textbf{OFlow} employs a Neural-ODE framework~\cite{chen2018neural} to learn deformations;
\textbf{OFlow} assigns each 4D point both an occupancy value and a motion velocity vector, utilizing a Neural-ODE framework~\cite{chen2018neural} for learning deformations. \textbf{LPDC} employs a MLP to parallelly learn correspondences among occupancy fields across different time steps via explicitly learning continuous displacement vector fields from spatio-temporal shape representation. \textbf{CaDeX} introduces a canonical map factorization and utilizes invertible deformation networks to maintain homeomorphisms.
For fair comparisons, we follow their original training paradigms.
%established by OFlow~\cite{OccupancyFlow} for D-FAUST~\cite{DFAUST}, and CaDeX ~\cite{Lei2022CaDeX} for DT4D-A~\cite{DeformingThings4D}.
\\
\textbf{Evaluation Metrics:} The Intersection over Union (IoU) evaluates the overlap between predicted and ground truth meshes; The Chamfer distance calculates the average nearest-neighbor distance between two point sets; $\ell_2$-distance error measures the Euclidean distance between corresponding points on the predicted and ground truth meshes. 
\\
\textbf{Implementations:} The training of our approach consists of two stages. The first stage involves two auto-encoders. The input point clouds ($N=2048$) are randomly sampled from object surfaces and near-surface regions. For the shape auto-encoder, the learning rate is $10^{-4}$, with KL-divergence loss weights $10^{-3}$. For the deformation auto-encoder, the learning rate is $10^{-4}$, with KL-divergence loss weights $10^{-6}$. They are trained for 100 epochs with batch size 24.
% We choose the number of latent codes $M = 512$ with $C = 32$ channels. 
The second stage is the diffusion models, the learning rate for both shape and deformation diffusion models is $10^{-4}$ and they are trained for 50 epochs with a batch size of 8 for shape diffusion and 4 for deformation diffusion.
% we leverage $2048$ surface points and the encoder we obtained in the first stage to get the target latent codes for diffusion mdoels.
\\
% \textbf{Training and inference Time:} Training: 60 hours  Inference: 11s for 17 frames (1*RTX 3080). %on a single RTX 3080.
\textbf{Runtime:} The training takes about 60 hours(2*RTX 4090) . The inference takes about 11s for 17 frames(1*RTX 3080).

\subsection{4D Shape Reconstruction} \label{sec:exp_rec}
\vspace*{-1mm}
We initially assessed our models' ability for 4D reconstruction from sparse and noisy point cloud sequences 
%on the two aforementioned datasets %D-FAUST \cite{DFAUST} and DT4D-A \cite{DeformingThings4D}. 
Consistent with the setup in OFlow \cite{OccupancyFlow}, our network processed sequences of $T=17$ continuous frames. 
Each frame represented a sparse point cloud, with $L = 300$ for D-FAUST \cite{DFAUST} or $L = 512$ for DT4D-A \cite{DeformingThings4D}. 
We also simulate noisy observations with Gaussian noise ($\sigma$ = 0.05).

% Despite their sparsity, these points provided a comprehensive representation of body shape and pose by being evenly distributed across various body regions.
Quantitatively, our model demonstrates superior performance over state-of-the-art models on the D-FAUST \cite{DFAUST} and DT4D-A \cite{DeformingThings4D} datasets, as detailed in Tab.~\ref{tab:sparse}. This superiority is particularly notable in the unseen individual category of the DT4D-A dataset, which features more diverse topologies from various animals. 
% Here, our model achieves an IoU that is 19\% higher than the preceding state-of-the-art model. 
Additionally, both chamfer distance and \(\ell_2\)-correspondence error are reduced to less than half of those recorded by the previous state-of-the-art methods.
Qualitatively, as illustrated in Fig.~\ref{fig:res_quali_sparse}, our model outperforms in reconstructing complete shapes and minimizing chamfer distance errors, particularly in capturing fast-moving structures like feet of humans and heads of animals. 

The superiority of our model is attributed to the proposed 4D latent set diffusion, enabling a more precise capture of local geometry and deformation patterns. Methods like LPDC \cite{LPDC} and OFlow \cite{OccupancyFlow} perform well in human settings thanks to similar human topologies, while CaDex \cite{Lei2022CaDeX} benefits from canonical shape learning. However, the diverse topologies and scales in animal setup, such as dragons, present a significant challenge for models that optimize global codes. Our approach, in contrast, effectively captures these complex 4D dynamics of general non-rigid objects. 

\subsection{4D Shape Completion} 
\vspace*{-1mm}
\subsubsection{Monocular Depth Sequences} \label{ex:depth}
\vspace*{-1.5mm}
To simulate sparse and partial real-world scans, we generated monocular depth sequences by rendering from a fixed camera angle. The size of the point cloud input and the frame length are the same as Sec.~\ref{sec:exp_rec}.
The qualitative and quantitative comparisons are presented in Fig.~\ref{fig:res_quali_partial} and Tab.~\ref{tab:partial}.
As seen, our method consistently outperforms all state-of-the-art methods in all metrics and produces more complete surface geometries with more plausible motion tracking. 
This demonstrates the effectiveness of the motion priors learned by our proposed 4D latent set diffusion in addressing ambiguous data such as partial observations.
\vspace*{-3mm}
\subsubsection{Partial Scan Sequences}
\begin{figure}[t]
  \centering
  \begin{overpic}[width=\linewidth]{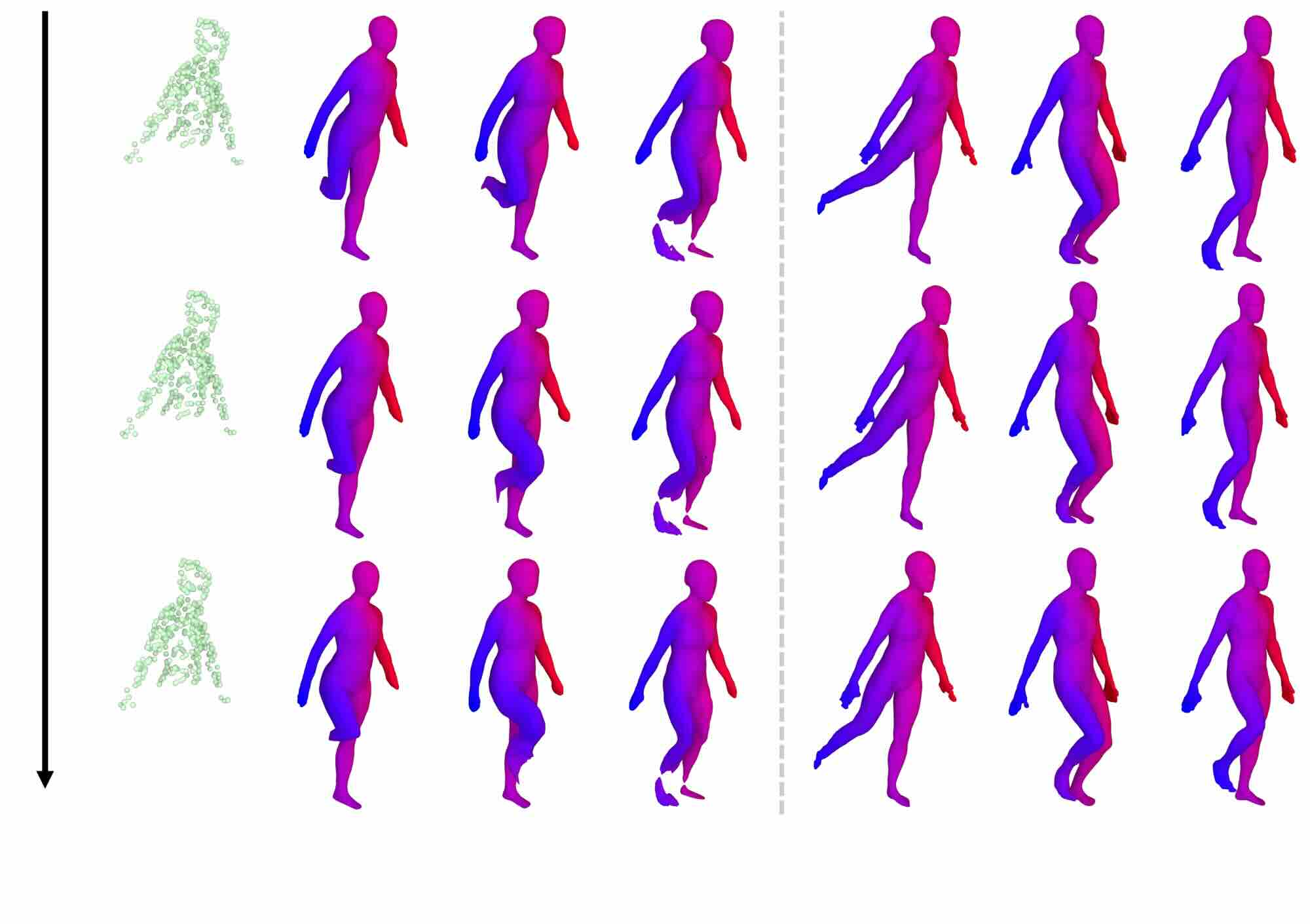}
  \put (5.8,7) {\large t}
  \put (22,7) {Input}
  \put (50,7) {OFlow}
  \put (80.5,7) {LPDC}
  \put (110,7) {CaDeX}
  \put (185,7) {Ours}
  \end{overpic}
\captionsetup{font=normalsize}
\vspace*{-5.8mm}
\caption{
Comparisons of 4D Shape Reconstruction from \textbf{highly partial} point cloud sequences, such as half-body scans obtained from the D-FAUST \cite{DFAUST} dataset. The colors of the meshes encode the correspondence. Our diffusion-based method produces highly complete human shapes with more favorable motions, offering multiple possible outputs that match the input observations.
}
\vspace*{-3mm}
\label{fig:exp_halfbody}
\end{figure}
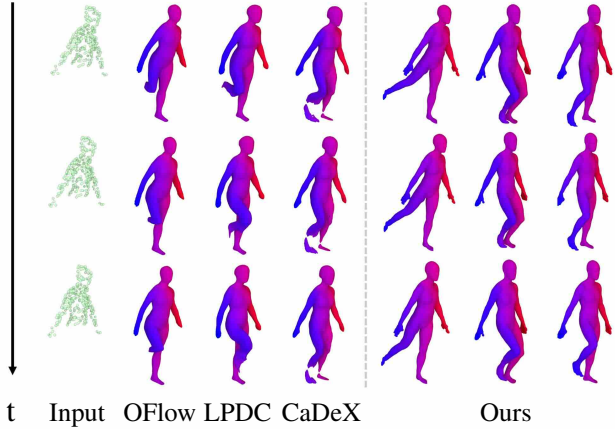

\vspace*{-1.5mm}
To assess the robustness of our method to extremely ambiguous data, we set up a challenging experiment on the D-FAUST \cite{DFAUST} dataset. This involved reconstructing whole body motions based on partial point clouds of the upper bodies. This setup creates a highly ambiguous scenario, as the same upper body motion can correspond to many different lower-body. We adopt the same configuration as Sec.~\ref{sec:exp_rec}, with a frame length ($T=17$) and input point cloud size ($L=300$). As the Fig.~\ref{fig:exp_halfbody} shows,
OFlow~\cite{OccupancyFlow}, LPDC~\cite{LPDC}, and CaDex~\cite{Lei2022CaDeX} face challenges in reconstructing the complete shape, often producing distorted shapes such as broken feet. In contrast, our method excels in reconstructing more complete geometries while achieving temporally coherent tracking.
Additionally, our approach present a diverse range of plausible full-body reconstructions that align with the given upper-body scans. The superior performance is primarily attributed to the 4D latent set diffusion.  
%While previous networks \cite{LPDC,OccupancyFlow,Lei2022CaDeX} struggle with ambiguity, 
Our diffusion-based method is more capable of tackling the `one-to-many' complexities from extremely partial data. 

% \vspace{-2mm}

% \begin{figure}[t]
%   \centering
%   % \begin{overpic}[width=\linewidth,grid,unit=1bp,tics=10]{fig/x_supp/behave_real_full.jpg}
%   \begin{overpic}[width=\linewidth]{fig/4_exp/ablation_wo_diff.jpg}
%   \put (2.5, 242) {\rotatebox{-90}{t}}
%   \put (1, 206) {\rotatebox{-90}{Input}}
%   \put (1, 144) {\rotatebox{-90}{W/o. diffusion}}
%   \put (1,46) {\rotatebox{-90}{Full}}
%   \end{overpic}
% \vspace*{-7mm}
%   \caption{Qualitative comparisons of ablation study on the diffusion model from \textbf{sparse and noisy} point clouds on the D-FAUST \cite{DFAUST}.  Our method with the diffusion model exhibits lower reconstruction errors.}
%   \label{fig:ablation_wo_diff}
% % \vspace*{-4mm}
% \end{figure}

\subsection{Ablation Study}
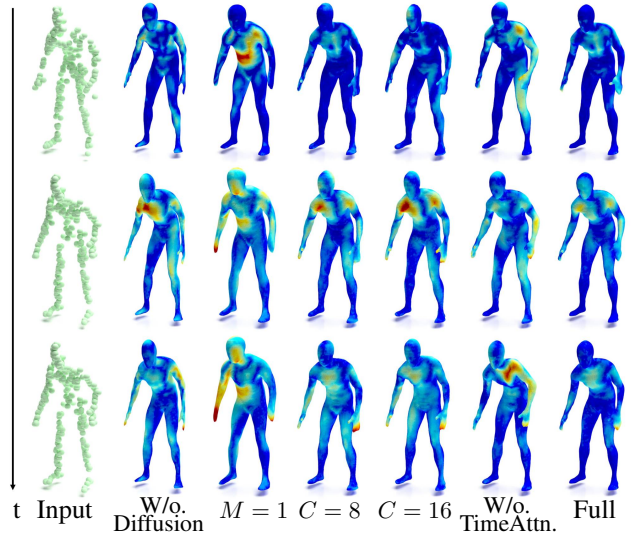
\begin{figure}[t]
  \centering
  % \begin{overpic}[width=\linewidth,grid,unit=1bp,tics=10]{fig/x_supp/behave_real_full.jpg}
  \begin{overpic}[width=\linewidth]{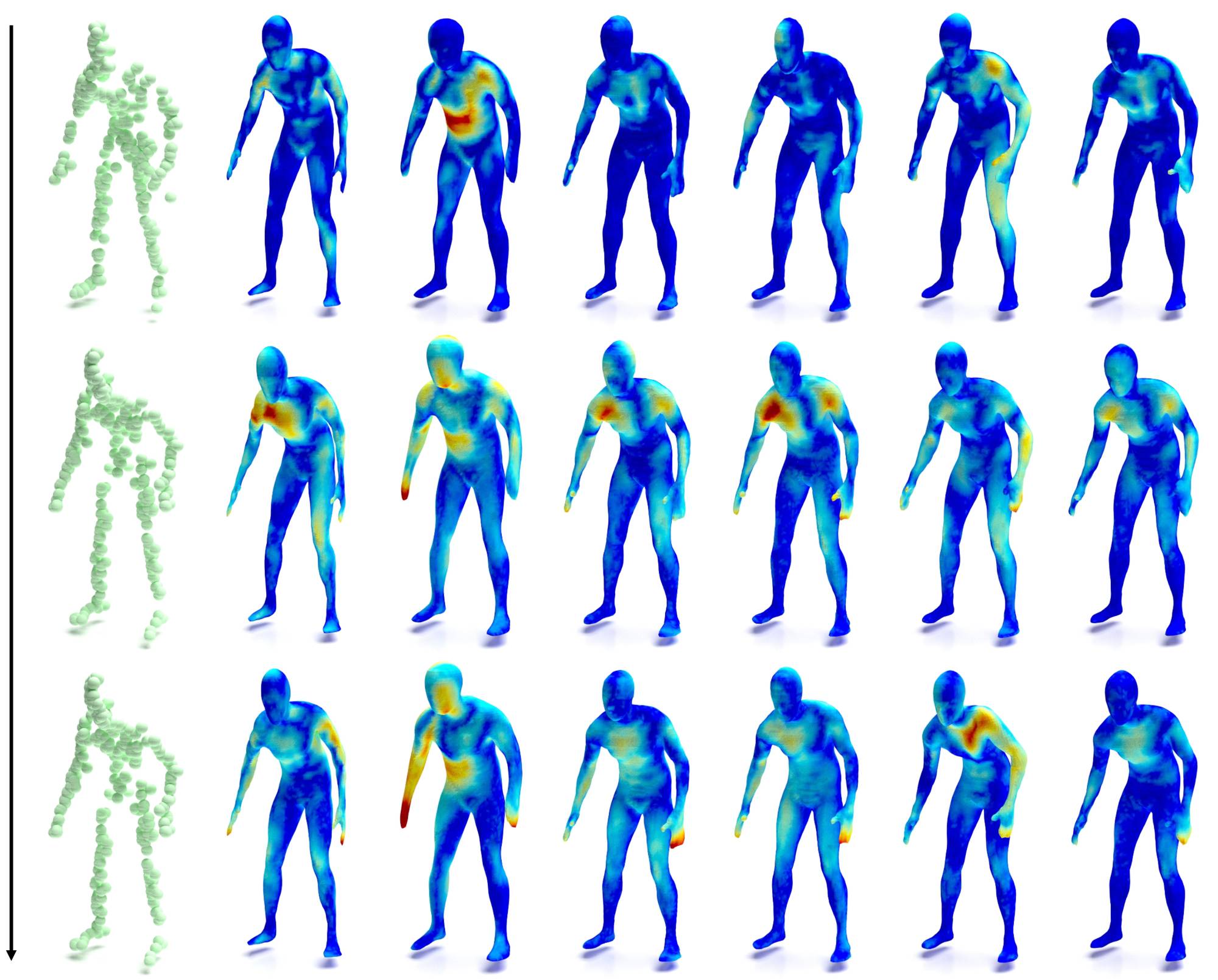}
  \put (2, -7) {t}
  \put (11, -7) {Input}
  \put (50, -5) {\scalebox{0.9}{W/o.}}
  \put (40, -12) {\scalebox{0.9}{Diffusion}}
  \put (80, -7) {$\scalebox{0.9}{\ensuremath{M=1}}$}
  \put (110, -7) {$\scalebox{0.9}{\ensuremath{C=8}}$}
  \put (140, -7) {$\scalebox{0.9}{\ensuremath{C=16}}$}
  \put (180, -5) {\scalebox{0.9}{W/o.}}
  \put (171, -12) {\scalebox{0.9}{TimeAttn.}}
  \put (214, -7) {Full}
  \end{overpic}
\vspace*{-2mm}
  \caption{Qualitative ablation studies on the D-FAUST \cite{DFAUST} dataset.}
  \label{fig:ablation_full}
\vspace*{-6mm}
\end{figure}

\vspace{-1mm}
We conducted ablation studies to validate the effectiveness of each component (See Tab.~\ref{tab:ablation}, Fig.~\ref{fig:ablation_full}) under the setting of 4D Shape Completion from \textbf{monocular noisy depth scans} on the D-FAUST \cite{DFAUST} dataset.
\vspace*{-5.3mm}
\paragraph{What is the effect of diffusion model?}
4D surface reconstruction from ambiguous observations of noisy, sparse, or partial point clouds is an ill-posed problem.
Deterministic models often yield sub-optimal results.
We provide comparisons against the variant of one-step regression without diffusion models. As shown in Fig.~\ref{fig:ablation_full} and Tab.~\ref{tab:ablation}, diffusion model uses a probabilistic way to 
deal with highly ambiguous inputs and generate plausible predictions.
Also, diffusion models can handle ``one-to-many'' problems and generate diverse and creative outputs as shown in Fig.~\ref{fig:exp_halfbody}.
% solve ``one-to-many'' problems with highly ambiguous inputs and generate diverse and creative outputs. More visualizations are shown in the Appendix.
\vspace*{-5.3mm}
\paragraph{What is the effect of latent vector set representation?} Instead of using single global latent code, our approach employs 4D latent vector sets. As indicated in Tab.~\ref{tab:ablation}, our method significantly outperforms the global latent codes (with $M=1$) and captures more accurate 4D motions. It becomes more apparent in unseen identities, demonstrating an enhanced generalization ability.
\vspace*{-5.3mm}
\paragraph{What is the effect of time attention layers?}  For the synchronized deformation latent set diffusion, we have integrated the time self-attention layer in our interleaved spatio-temporal attention mechanism. 
%our experimental setup was aligned with the shape code settings used in Shape2VecSet \cite{Shape2Vecset}
%
We attempted to remove the layer. However, the results showed a decrease in all metrics, highlighting the effectiveness of the time self-attention layer in maintaining temporal coherence.
\vspace*{-5.3mm}
\paragraph{What is the effect of the number of channels of latent set?}
% the settings for the shape codes are kept constant
%
To find out the optimal configuration for learning shape and deformation priors within time-varying deformable surfaces, we tried the channel numbers $C$ of the shape and deformation latent sets. The experimental results indicated that using $C=32$ channels for 4D latent set diffusion is more suitable, yielding more favorable results.
\vspace*{-1mm}
\begin{table}[t!]
\centering
\scalebox{0.80}{
\begin{tabular}{@{}lccc|ccc@{}}
\toprule
\multicolumn{1}{c}{\multirow{2}{*}{Method}} & \multicolumn{3}{c|}{Unseen Motion}              & \multicolumn{3}{c}{Unseen Individual}             \\ \cmidrule(l){2-7} 
\multicolumn{1}{c}{}            & IoU↑   & CD↓   & Corr↓ & IoU↑   & CD↓   & Corr↓ \\ \midrule
\multicolumn{1}{c|}{W/o. Diffusion}               & 71.1\% & 0.097 & 0.173 & 64.2\% & 0.107 & 0.194 \\
\multicolumn{1}{c|}{$M = 1$}               & 68.5\% & 0.120 & 0.301 & 57.7\% & 0.149 & 0.327 \\
\multicolumn{1}{c|}{$C = 8$}               & 78.9\% & 0.078 & 0.180 & 68.0\% & 0.105 & 0.225 \\
\multicolumn{1}{c|}{$C = 16$}               & 78.0\% & 0.080 & 0.189 & 66.8\% & 0.109 & 0.254 \\
\multicolumn{1}{c|}{W/o. TimeAttn.}               & 81.2\% & 0.061 & 0.127 & 70.8\% & 0.086 & 0.158 \\
\multicolumn{1}{c|}{Full}               & \textbf{83.8\%} & \textbf{0.054} & \textbf{0.111} & \textbf{74.4\%} & \textbf{0.075} & \textbf{0.140} \\ \bottomrule
\end{tabular}
}
\vspace*{-1.5mm}
\caption{Quantitative ablation studies on the D-FAUST~\cite{DFAUST} dataset. $M$ denotes the number of latent codes and $C$ represents the number of latent code channels.}
\vspace*{-4.5mm}
\label{tab:ablation}
\end{table}

\section{Conclusion}
\vspace*{-1mm}
We present Motion2VecSets, a 4D diffusion model for dynamic surface reconstruction from point cloud sequences. Our method explicitly models the shape and motion distributions of non-rigid objects through an iterative denoising process, using compressed latent sets to generate plausible and diverse outputs. The learned shape and motion diffusion priors can effectively deal with ambiguous observations, including sparse, noisy, and partial data. Compared to encoding shape and deformation with a global latent, our novel 4D latent set representation enables more accurate non-linear motion capture and improves the generalizability to unseen identities and motions. The designed interleaved space and time attention block for synchronized deformation vector sets diffusion enforces temporal-coherent object tracking while reducing computational overhead. Extensive experiments demonstrate our approach's superiority in reconstructing sparse, partial, and even half-body point clouds on the D-FAUST \cite{DFAUST} and DT4D-A \cite{DeformingThings4D} datasets, underlining its robustness to various types of imperfect observations. We believe that Motion2VecSets has the potential for future extension into multi-modal domains, such as text-driven 4D generation and RGB video-based 4D reconstruction.

\vspace*{-3mm}
\paragraph{Acknowledgement.}
This work was supported by the ERC Starting Grant Scan2CAD (804724) as well as the German Research Foundation (DFG) Research Unit ``Learning and Simulation in Visual Computing''. We further thank Angela Dai for the video voice-over.
% \newpage
{
    \small
    \bibliographystyle{ieeenat_fullname}
    \bibliography{main}
}

\clearpage
\appendix
\section*{Appendix}
\setcounter{section}{0}
In this supplementary material, we complement our main paper with additional details. It begins with an introduction to the notations used in our paper in Sec.~\ref{sec:notations}. The following section, Sec.~\ref{sec:network_arch}, discusses our network architectures, highlighting key design choices and their impact. This is complemented by Sec.~\ref{sec:implementation_details}, which provides essential implementation details. Then, we provide more comparisons between our approach and state-of-the-art methods in Sec.~\ref{sec:more_res}, offering insights into our model's strengths and improvements. 
We also provide an additional ablation study to verify the usefulness of latent set diffusion on the setting of 4D shape reconstruction from sparse and noisy point clouds.
Finally, in Sec.~\ref{sec:other_app}, we demonstrate the real-world applicability of our model, underscoring its practical effectiveness.

\section{Notations}
\label{sec:notations}
%\vspace*{-2mm}
In this paper, we use the following notations, as summarized in Tab.~\ref{tab:symbol_table}. The symbol \( T \) represents the sequence length, with the superscript \( t \) indicating the time step of a frame in the sequence. Our network's input is a sparse or partial point cloud, denoted as \( \mathcal{P} = \left\{\mathbf{P}^t\right\}_{t=0}^{T-1} \), where each frame \( \mathbf{P}^t \) comprises \( L \) points, serving as conditions in reference time. The sets \( \mathcal{S}, \mathcal{D}, \mathcal{C} \) represent shape codes, deformation codes, and conditional codes, respectively, with each \( i \)-th code denoted by the corresponding lowercase letter. The mesh is symbolized by \( \mathcal{M} = \{\mathcal{V},\mathcal{F}\} \), where \( \mathcal{V} \) and \( \mathcal{F} \) correspond to the vertices and faces set of the mesh. Furthermore, \( \mathbf{X} \) refers to surface points from ground truth meshes for learning shape, and \( \mathbf{X}^d \) refers to corresponding sampled points by FPS. We dedicate $\mathbf{X}_\text{src},\mathbf{X}_\text{tgt}$ for representing the sampled surface points for learning deformation, and $\mathbf{X}^{d}_\text{src},\mathbf{X}^{d}_\text{tgt}$ refer to sampled points by FPS. Moreover,\( \mathcal{Q} \) represents the set of query points, where $\mathbf{q}$ refers to a point inside the point set, and $\mathbf{q'}$ is the output by feeding the query point into the deformation network.

% \vspace*{-2mm}
\begin{table}[t]
  \centering
  \begin{tabular}{@{}lll@{}}
    \toprule
     & Symbol                        & Meaning \\ 
    \midrule
    & $T$                           & \# frames \\ \specialrule{.0em}{.2em}{.2em} 
 & $\mathcal{P} = \left\{\mathbf{P}^t\right\}_{t=0}^{T-1}$                  & \makecell[l]{Sparse point clouds as input for \\diffusion models}\\ \specialrule{.0em}{.2em}{.2em} 
 &$ \mathbf{P}^t = \left\{\mathbf{p}_i\in\mathbb{R}^3\right\}_{i=0}^{L}$ & \makecell[l]{A frame of sparse point clouds}  \\\specialrule{.0em}{.2em}{.2em} 
 & $L$                           & \# points in sparse point clouds \\\specialrule{.05em}{.3em}{.3em} 
 & $\mathcal{S} = \{\mathbf{s}_i \in \mathbb{R}^C \}_{i=0}^{M}$ & \makecell[l]{One set of shape codes with $M$ \\latent codes} \\\specialrule{.0em}{.2em}{.2em}
  & $s_i$                         & $i$-th shape code \\\specialrule{.0em}{.2em}{.2em} 
 % \\
 & $\mathcal{D} = \{ \mathbf{d}_i \in \mathbb{R}^C \}_{i=0}^{M}$ & \makecell[l]{One set of deformation codes with\\ $M$ latent codes} \\\specialrule{.0em}{.2em}{.2em} 
 % \\
 & $d_i$                         & $i$-th deformation code \\\specialrule{.0em}{.2em}{.2em} 
 % \\
 & $\mathcal{C} = \{ \mathbf{c}_i \in \mathbb{R}^C \}_{i=0}^{M}$ & \makecell[l]{One set of conditional codes with\\ $M$ latent codes}\\\specialrule{.0em}{.2em}{.2em} 
 % \\
 & $c_i$                         & $i$-th deformation code \\\specialrule{.0em}{.2em}{.2em} 
  & $C$                           & \# latent code channels \\\specialrule{.05em}{.3em}{.5em} 
 % \\
 & $\mathcal{V},\mathcal{F}$     & Vertices and faces of mesh \\\specialrule{.0em}{.2em}{.2em} 
& $\mathbf{X}$                  & \makecell[l]{Points from meshes as\\ input for shape autoencoder} \\\specialrule{.0em}{.2em}{.2em} 
 % \\
  & $\mathbf{X}^d$                  & \makecell[l]{Points sampled by FPS from \\meshes as input for shape\\autoencoder} \\\specialrule{.0em}{.2em}{.2em} 
 & $\mathbf{X}_{\text{src}}, \mathbf{X}_{\text{tgt}}$  &  \makecell[l]{Points from source and\\ target meshes as input for\\ deformation autoencoder} \\\specialrule{.0em}{.2em}{.2em} 
 & $\mathbf{X}^{d}_{\text{src}}, \mathbf{X}^{d}_{\text{tgt}}$  &  \makecell[l]{Points sampled by FPS from \\source and target meshes \\as input for deformation \\ autoencoder} \\\specialrule{.0em}{.2em}{.2em} 
 & $\mathcal{Q}$ & Query points set \\\specialrule{.0em}{.2em}{.2em} 
 & $\mathbf{q}$ & Query point \\\specialrule{.0em}{.2em}{.2em} 
 & $\mathbf{q'}$ & Deformed query point  \\
    \bottomrule
  \end{tabular}
  \caption{\textbf{Notation table} includes the mathematical symbols we mentioned in the paper.}
  \label{tab:symbol_table}
\end{table}

\section{Network Architectures}
\label{sec:network_arch}
% \vspace*{-2mm}
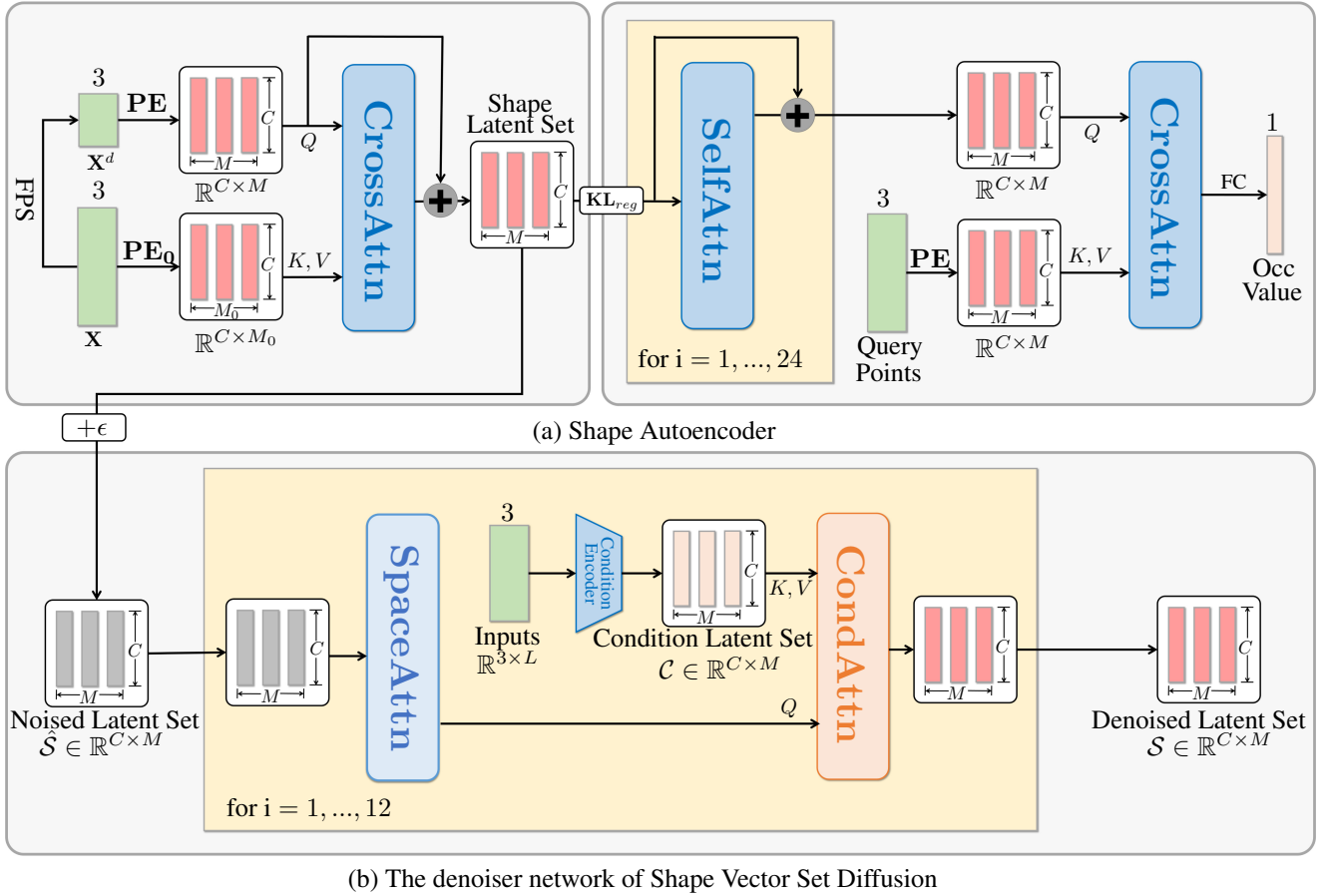
\begin{figure*}[t]
  \centering
  % \begin{overpic}[width=\linewidth,grid,unit=1bp,tics=10]{fig/x_supp/ShapeAE.jpg}
  \begin{overpic}[width=\linewidth]{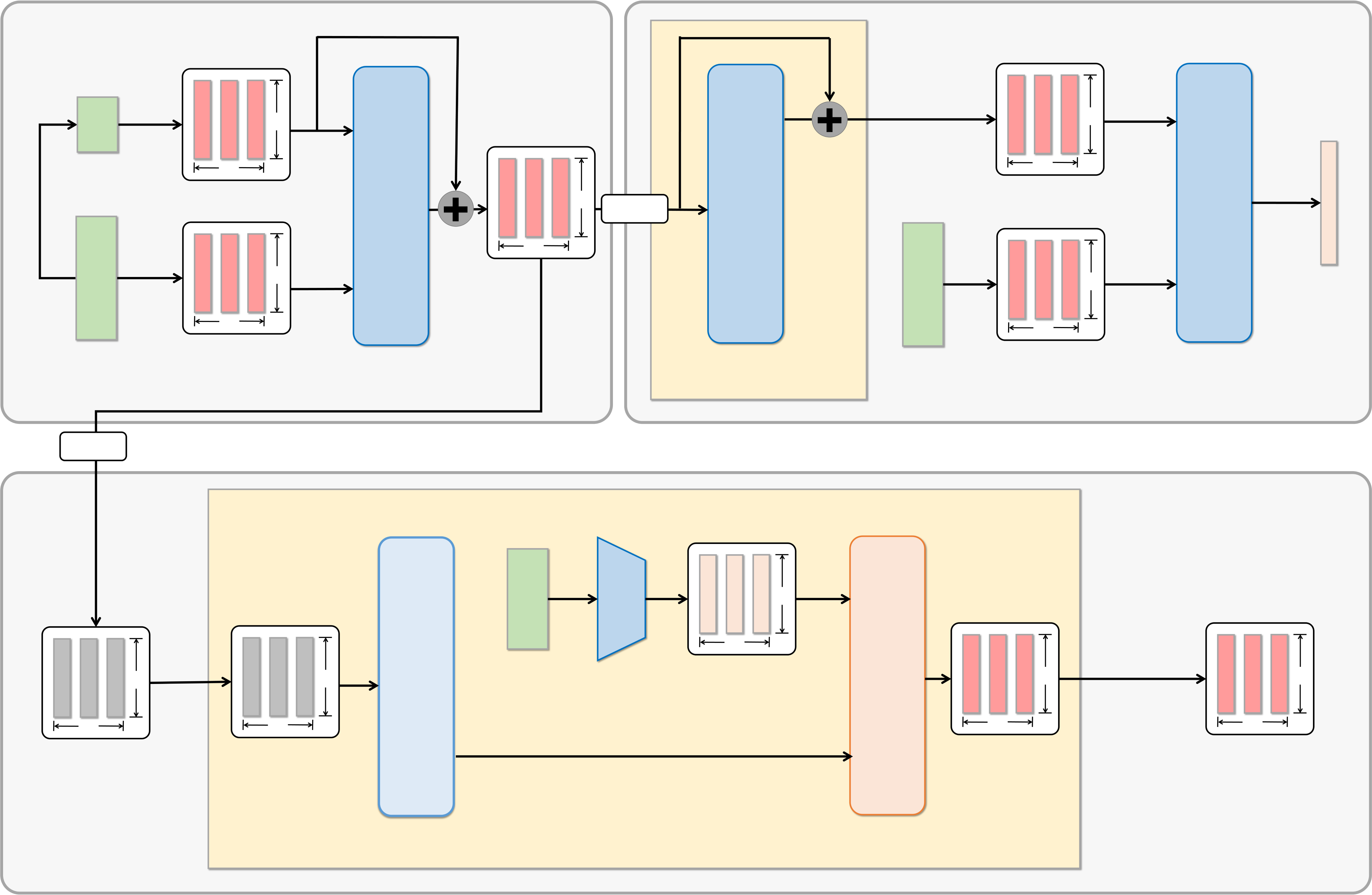}
    % \put (10, 178) {$\scalebox{1.2}{\ensuremath{\mathbf{Encoder}}}$}
    % \put (440, 178) {$\scalebox{1.2}{\ensuremath{\mathbf{Decoder}}}$}
    \put (33, 292) {$3$}
    \put (33, 249) {$3$}
    \put (4, 257) {$\rotatebox{-90}{\text{FPS}}$}
    \put (46, 282) {$\mathbf{PE}$}
    \put (44.5, 226) {$\mathbf{PE_0}$}
    \put (97,277.5) {$\scalebox{0.7}{\ensuremath{C}}$}
    \put (97,222) {$\scalebox{0.7}{\ensuremath{C}}$}
    \put (79,260) {$\scalebox{0.7}{\ensuremath{M}}$}
    \put (79,204.5) {$\scalebox{0.7}{\ensuremath{M_0}}$}
    \put (107,223) {$\scalebox{0.8}{\ensuremath{K,V}}$}
    \put (112,268.5) {$\scalebox{0.8}{\ensuremath{Q}}$}

    \put (30,193) {$\scalebox{0.8}{\ensuremath{\mathbf{X}}}$}
    % \put (29.5,258) {$\scalebox{0.5}{\ensuremath{\mathbb{R}^{3\times M}}}$}
    \put (31,260) {$\scalebox{0.8}{\ensuremath{\mathbf{X}^d}}$}
    % \put (26.5,189.5) {$\scalebox{0.5}{\ensuremath{\mathbf{X} \in \mathbb{R}^{3\times N}}}$}

    \put (72,248) {$\scalebox{1}{\ensuremath{\mathbb{R}^{C\times M}}}$}
    \put (72,191) {$\scalebox{1}{\ensuremath{\mathbb{R}^{C\times M_0}}}$}

    \put (136.5, 287) {$\textcolor[RGB]{0,112,192}{\rotatebox{-90}{\scalebox{1.5}{\ensuremath{\mathbf{CrossAttn}}}}}$}
    \put (208,249.5) {$\scalebox{0.7}{\ensuremath{C}}$}
    \put (189.5,232) {$\scalebox{0.7}{\ensuremath{M}}$}
    \put (183,282) {$\scalebox{1}{\text{Shape}}$}
    \put (175,274) {$\scalebox{1}{\text{Latent Set}}$}
    % \put (188,42) {$\scalebox{0.5}{\ensuremath{\mathbb{R}^{C\times M}}}$}
    
    \put (220.3,246.4) {$\scalebox{0.7}{\ensuremath{\mathbf{KL}_{reg}}}$}
    
    \put (264.5, 280) {$\textcolor[RGB]{0,112,192}{\rotatebox{-90}{\scalebox{1.5}{\ensuremath{\mathbf{SelfAttn}}}}}$}
    
    \put (240, 185) {$\text{for i} = 1,...,24$}
    \put (330.5, 246) {$3$}
    \put (373.5,202) {$\scalebox{0.7}{\ensuremath{M}}$}
    \put (392,219.5) {$\scalebox{0.7}{\ensuremath{C}}$}
    \put (373.5,261.5) {$\scalebox{0.7}{\ensuremath{M}}$}
    \put (392,279.5) {$\scalebox{0.7}{\ensuremath{C}}$}
    \put (343, 223) {$\mathbf{PE}$}
    \put (323, 190) {$\scalebox{1}{\text{Query}}$}
    \put (322.5, 180) {$\scalebox{1}{\text{Points}}$}
    % \put (326,182) {$\scalebox{0.5}{\ensuremath{\mathbb{R}^{3\times 2N}}}$}

    \put (368,188.8) {$\scalebox{1}{\ensuremath{\mathbb{R}^{C\times M}}}$}
    \put (368,249) {$\scalebox{1}{\ensuremath{\mathbb{R}^{C\times M}}}$}

    \put (473, 219.5) {$\scalebox{1}{\text{Occ}}$}
    \put (469, 210.5) {$\scalebox{1}{\text{Value}}$}
    % \put (475,210) {$\scalebox{0.5}{\ensuremath{\mathbb{R}^{1\times 2N}}}$}
    
    \put (434.5, 288) {$\textcolor[RGB]{0,112,192}{\rotatebox{-90}{\scalebox{1.5}{\ensuremath{\mathbf{CrossAttn}}}}}$}
    \put (403,225) {$\scalebox{0.8}{\ensuremath{K,V}}$}
    \put (409,272) {$\scalebox{0.8}{\ensuremath{Q}}$}
    \put (478, 275) {$1$}
    \put (461, 253) {$\scalebox{0.8}{\text{FC}}$}
    % \put (457, 242) {$\scalebox{0.8}{(C,1)}$}
    % Part 2
    \put (27, 159.5) {$+\epsilon$}
    
    \put (46.5,75.5) {$\scalebox{0.7}{\ensuremath{C}}$}
    \put (115.5,76) {$\scalebox{0.7}{\ensuremath{C}}$}
    \put (28,57.5) {$\scalebox{0.7}{\ensuremath{M}}$}
    \put (97,58) {$\scalebox{0.7}{\ensuremath{M}}$}
    \put (2, 48) {$\scalebox{1}{\text{Noised Latent Set}}$}
    \put (13, 38) {$\scalebox{1}{\ensuremath{\hat{\mathcal{S}} \in \mathbb{R}^{C \times M}}}$}
    \put (143.5, 118) {$\textcolor[RGB]{68,114,196}{\rotatebox{-90}{\scalebox{1.5}{\ensuremath{\mathbf{SpaceAttn}}}}}$}
    \put (188, 127) {$3$}
    % \put (174, 83) {$\scalebox{0.5}{\text{Sparse or Partial}}$}
    \put (178, 80) {$\scalebox{1}{\text{Inputs}}$}
    \put (179, 69) {$\scalebox{1}{\ensuremath{\mathbb{R}^{3\times L}}}$}

    \put (223, 79) {$\scalebox{1}{\text{Condition Latent Set}}$}
    \put (248, 67) {$\scalebox{1}{\ensuremath{\mathcal{C} \in \mathbb{R}^{C\times M}}}$}
    
    \put (226, 121) {$\textcolor[RGB]{0,112,192}{\rotatebox{-90}{\scalebox{0.7}{\ensuremath{\text{Condition}}}}}$}
    \put (220, 118.5) {$\textcolor[RGB]{0,112,192}{\rotatebox{-90}{\scalebox{0.7}{\ensuremath{\text{Encoder}}}}}$}
    \put (280.2,105.7) {$\scalebox{0.7}{\ensuremath{C}}$}
    \put (262,88) {$\scalebox{0.7}{\ensuremath{M}}$}
    \put (315.5, 118) {$\textcolor[RGB]{237,125,49}{\rotatebox{-90}{\scalebox{1.5}{\ensuremath{\mathbf{CondAttn}}}}}$}
    \put (289.5,99) {$\scalebox{0.8}{\ensuremath{K,V}}$}
    \put (294,54) {$\scalebox{0.8}{\ensuremath{Q}}$}
    \put (375.5,77) {$\scalebox{0.7}{\ensuremath{C}}$}
    \put (357,59) {$\scalebox{0.7}{\ensuremath{M}}$}
    \put (467.5,76.7) {$\scalebox{0.7}{\ensuremath{C}}$}
    \put (450,59) {$\scalebox{0.7}{\ensuremath{M}}$}
    \put (412, 48) {$\scalebox{1}{\text{Denoised Latent Set}}$}
    \put (433, 38) {$\scalebox{1}{\ensuremath{{\mathcal{S}} \in \mathbb{R}^{C \times M}}}$}
    % \put (410, 20) {$\scalebox{1.2}{\ensuremath{\mathbf{Shape\textbf{ }Vector}}}$}
    % \put (411.5, 5) {$\scalebox{1.2}{\ensuremath{\mathbf{Set\textbf{ }Diffusion}}}$}
    \put (200, 158) {$\scalebox{1}{\text{(a) Shape Autoencoder}}$}
    \put (130, -12) {$\scalebox{1}{\text{(b) The denoiser network of Shape Vector Set Diffusion}}$}
    \put (84, 15) {$\text{for i} = 1,...,12$}
    
\end{overpic}
\vspace{-.2em}
\caption{Network architecture  of \textbf{ Shape  Diffusion}}
\label{fig:shape}
\end{figure*}

The inputs of our model are a sequence of $T$ frames of sparse or partial noisy point clouds, represented by $\mathcal{P}=\left\{\mathbf{P}^t\right\}_{t=1}^{T}$, where $\mathbf{P}^t=\left\{\mathbf{p}_i \in \mathbb{R}^3\right\}_{i=1}^L$, $L$ represents the number of points. The goal is to reconstruct continuous 3D meshes with high fidelity, denoted as $\{\mathcal{M}^t\}_{t=1}^{T} = \{\mathcal{V}^t, \mathcal{F}^t\}_{t=1}^{T}$, where $\mathcal{V}^t$ and $\mathcal{F}^t$ refer to the set of vertices faces of the reconstructed mesh at time frame $t$.

\begin{figure*}[t]
  \centering
  % \begin{overpic}[width=\linewidth,grid,unit=1bp,tics=10]{fig/x_supp/ShapeAE.jpg}
  \begin{overpic}[width=\linewidth]{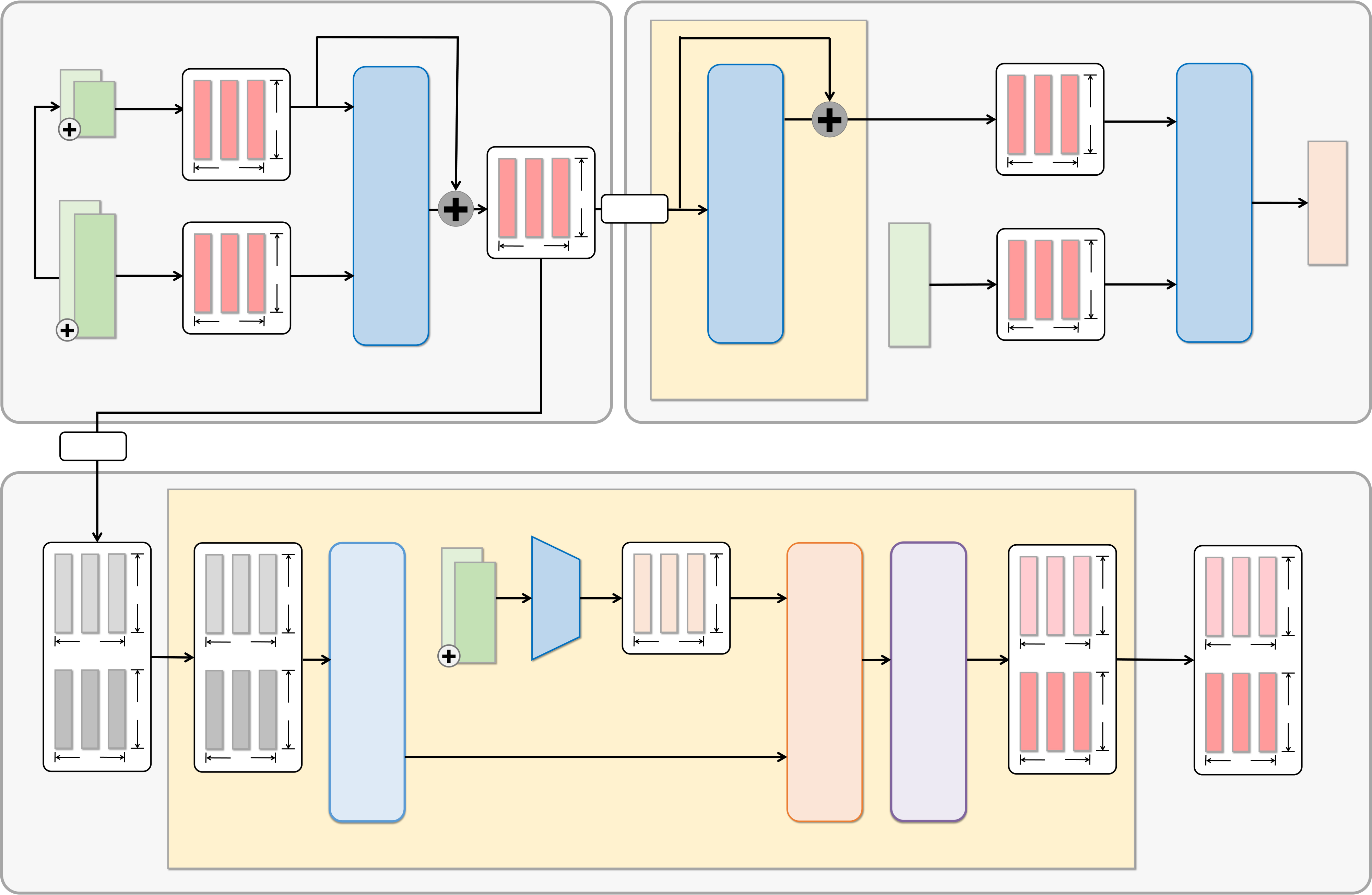}
    % \put (10, 178) {$\scalebox{1.2}{\ensuremath{\mathbf{Encoder}}}$}
    % \put (440, 178) {$\scalebox{1.2}{\ensuremath{\mathbf{Decoder}}}$}
    \put (29, 302) {$3$}
    \put (29, 253) {$3$}
    \put (3, 263) {$\rotatebox{-90}{\text{FPS}}$}
    \put (46, 286) {$\mathbf{PE}$}
    \put (46, 226) {$\mathbf{PE}_0$}
    \put (97,277.5) {$\scalebox{0.7}{\ensuremath{C}}$}
    \put (97,221.7) {$\scalebox{0.7}{\ensuremath{C}}$}
    \put (79,260) {$\scalebox{0.7}{\ensuremath{M}}$}
    \put (78.3,204.5) {$\scalebox{0.7}{\ensuremath{M_0}}$}
    \put (107,228) {$\scalebox{0.8}{\ensuremath{K,V}}$}
    \put (112,276.5) {$\scalebox{0.8}{\ensuremath{Q}}$}

    % \put (16.5,195) {$\scalebox{0.5}{\text{Src \& Tgt}}$}
    % \put (25.5,263) {$\scalebox{0.5}{\ensuremath{\mathbb{R}^{3\times 2M}}}$}
    \put (15,263) {$\scalebox{1}{\ensuremath{\{\mathbf{X}^d_{\text{src}},\mathbf{X}^d_{\text{tgt}}\} }}$}
    \put (6,190) {$\scalebox{1}{\ensuremath{\{\mathbf{X}_{\text{src}},\mathbf{X}_{\text{tgt}}\}}}$}

    \put (73,248) {$\scalebox{1}{\ensuremath{\mathbb{R}^{C\times M}}}$}
    \put (72,192) {$\scalebox{1}{\ensuremath{\mathbb{R}^{C\times M_0}}}$}

    \put (136.5, 287) {$\textcolor[RGB]{0,112,192}{\rotatebox{-90}{\scalebox{1.5}{\ensuremath{\mathbf{CrossAttn}}}}}$}
    \put (207.5,249.3) {$\scalebox{0.7}{\ensuremath{C}}$}
    \put (189.5,232) {$\scalebox{0.7}{\ensuremath{M}}$}
    \put (180,283) {$\scalebox{1}{\text{Deform}}$}
    \put (175,274) {$\scalebox{1}{\text{Latent Set}}$}
    % \put (188,42) {$\scalebox{0.5}{\ensuremath{\mathbb{R}^{C\times M}}}$}
    
    \put (220.3,246.4) {$\scalebox{0.7}{\ensuremath{\mathbf{KL}_{reg}}}$}
    
    \put (264.5, 280) {$\textcolor[RGB]{0,112,192}{\rotatebox{-90}{\scalebox{1.5}{\ensuremath{\mathbf{SelfAttn}}}}}$}
    
    \put (240, 185) {$\text{for i} = 1,...,24$}
    \put (327, 246) {$3$}
    \put (373.8,202) {$\scalebox{0.7}{\ensuremath{M}}$}
    \put (391.7,219.4) {$\scalebox{0.7}{\ensuremath{C}}$}
    \put (373.2,261.5) {$\scalebox{0.7}{\ensuremath{M}}$}
    \put (391.5,279.5) {$\scalebox{0.7}{\ensuremath{C}}$}
    \put (340, 223) {$\mathbf{PE}$}
    \put (318, 190) {$\scalebox{1}{\text{Query}}$}
    \put (318, 179) {$\scalebox{1}{\text{Points}}$}
    % \put (323,181) {$\scalebox{0.5}{\ensuremath{\mathbb{R}^{3\times N}}}$}

    \put (369,189.8) {$\scalebox{1}{\ensuremath{\mathbb{R}^{C\times M}}}$}
    \put (369,250) {$\scalebox{1}{\ensuremath{\mathbb{R}^{C\times M}}}$}

    \put (455, 221.5) {$\scalebox{1}{\text{Deformed}}$}
    \put (462.5, 213) {$\scalebox{1}{\text{Points}}$}
    % \put (474.5,210) {$\scalebox{0.5}{\ensuremath{\mathbb{R}^{3\times N}}}$}
    
    \put (434.5, 288) {$\textcolor[RGB]{0,112,192}{\rotatebox{-90}{\scalebox{1.5}{\ensuremath{\mathbf{CrossAttn}}}}}$}
    \put (403,225) {$\scalebox{0.8}{\ensuremath{K,V}}$}
    \put (409,272) {$\scalebox{0.8}{\ensuremath{Q}}$}
    \put (478, 275) {$3$}
    \put (459, 253) {$\scalebox{0.8}{\text{FC}}$}
    % \put (455, 242) {$\scalebox{0.8}{(C,3)}$}
    % Part 2
    \put (27, 159.5) {$+\epsilon$}
    
    \put (46.5,64) {$\scalebox{0.7}{\ensuremath{C}}$}
    \put (46.5,106) {$\scalebox{0.7}{\ensuremath{C}}$}
    \put (101,64) {$\scalebox{0.7}{\ensuremath{C}}$}
    \put (101,106) {$\scalebox{0.7}{\ensuremath{C}}$}
    \put (28,46) {$\scalebox{0.7}{\ensuremath{M}}$}
    \put (28,88) {$\scalebox{0.7}{\ensuremath{M}}$}
    \put (83,46) {$\scalebox{0.7}{\ensuremath{M}}$}
    \put (83,88) {$\scalebox{0.7}{\ensuremath{M}}$}
    \put (24, 35) {$\scalebox{1}{\text{Noised Latent Set}}$}
    \put (8, 24) {$\scalebox{1}{\ensuremath{\{\hat{\mathcal{D}}_t\}_{t=2}^T \in \mathbb{R}^{C \times (T-1) \times M}}}$}
    \put (126.5, 118) {$\textcolor[RGB]{68,114,196}{\rotatebox{-90}{\scalebox{1.5}{\ensuremath{\mathbf{SpaceAttn}}}}}$}
    \put (167, 128) {$3$}
    % \put (155, 78) {$\scalebox{0.5}{\text{Sparse or Partial}}$}
    \put (151, 75) {$\scalebox{1}{\text{Input Pairs}}$}
    \put (152, 65) {$\scalebox{1}{\ensuremath{\{\mathbf{P}_1, \mathbf{P}_t\}_{t=2}^T}}$}
    \put (155, 53) {$\scalebox{1}{\ensuremath{\in\mathbb{R}^{3\times L \times2}}}$}

    % \put (222, 80) {$\scalebox{0.5}{\text{Conditional embedding}}$}
    % \put (236, 74) {$\scalebox{0.5}{\ensuremath{\mathcal{C} \in \mathbb{R}^{C\times M}}}$}

    \put (202, 75) {$\scalebox{1}{\text{Condition Latent Set}}$}
    \put (224, 63) {$\scalebox{1}{\ensuremath{\mathcal{C} \in \mathbb{R}^{C\times M}}}$}
    \put (202, 121) {$\textcolor[RGB]{0,112,192}{\rotatebox{-90}{\scalebox{0.7}{\ensuremath{\text{Condition}}}}}$}
    \put (195, 118.5) {$\textcolor[RGB]{0,112,192}{\rotatebox{-90}{\scalebox{0.7}{\ensuremath{\text{Encoder}}}}}$}
    \put (256.5,105.8) {$\scalebox{0.7}{\ensuremath{C}}$}
    \put (238.5,88) {$\scalebox{0.7}{\ensuremath{M}}$}
    \put (294.5, 116) {$\textcolor[RGB]{237,125,49}{\rotatebox{-90}{\scalebox{1.5}{\ensuremath{\mathbf{CondAttn}}}}}$}
    \put (331.5, 114) {$\textcolor[RGB]{127,100,158}{\rotatebox{-90}{\scalebox{1.5}{\ensuremath{\mathbf{TimeAttn}}}}}$}
    \put (265.5,99) {$\scalebox{0.8}{\ensuremath{K,V}}$}
    \put (270,54) {$\scalebox{0.8}{\ensuremath{Q}}$}
    \put (396.3,63) {$\scalebox{0.7}{\ensuremath{C}}$}
    \put (396,105) {$\scalebox{0.7}{\ensuremath{C}}$}
    \put (377.8,45.4) {$\scalebox{0.7}{\ensuremath{M}}$}
    \put (377.8,87.5) {$\scalebox{0.7}{\ensuremath{M}}$}
    \put (464,62.9) {$\scalebox{0.7}{\ensuremath{C}}$}
    \put (464,104.9) {$\scalebox{0.7}{\ensuremath{C}}$}
    \put (445.3,45.5) {$\scalebox{0.7}{\ensuremath{M}}$}
    \put (445,87) {$\scalebox{0.7}{\ensuremath{M}}$}
    \put (380, 35) {$\scalebox{1}{\text{Denoised Latent Set}}$}
    \put (375, 23) {$\scalebox{1}{\ensuremath{\{\mathcal{D}_t\}_{t=2}^T \in \mathbb{R}^{C \times (T-1) \times M}}}$}
    % \put (439, 21) {$\scalebox{0.8}{\ensuremath{\mathbf{Synchronized}}}$}
    % \put (414.5, 13) {$\scalebox{0.8}{\ensuremath{\mathbf{Deformation\textbf{ }Vector}}}$}
    % \put (439, 5) {$\scalebox{0.8}{\ensuremath{\mathbf{Sets\textbf{ }Diffusion}}}$}
    \put (195, 158) {$\scalebox{1}{\text{(a) Deform Autoencoder}}$}
    \put (90, -12) {$\scalebox{1}{\text{(b) The denoiser network of Synchronized Deformation Vector Sets Diffusion}}$}
    \put (70, 15) {$\text{for i} = 1,...,12$}

    \put (26,81.5) {$\scalebox{1.2}{\ensuremath{\cdot}}$}
    \put (31,81.5) {$\scalebox{1.2}{\ensuremath{\cdot}}$}
    \put (36,81.5) {$\scalebox{1.2}{\ensuremath{\cdot}}$}

    \put (80,81.5) {$\scalebox{1.2}{\ensuremath{\cdot}}$}
    \put (85,81.5) {$\scalebox{1.2}{\ensuremath{\cdot}}$}
    \put (90,81.5) {$\scalebox{1.2}{\ensuremath{\cdot}}$}
    
    \put (376,81) {$\scalebox{1.2}{\ensuremath{\cdot}}$}
    \put (381,81) {$\scalebox{1.2}{\ensuremath{\cdot}}$}
    \put (386,81) {$\scalebox{1.2}{\ensuremath{\cdot}}$}
    
    \put (444,81) {$\scalebox{1.2}{\ensuremath{\cdot}}$}
    \put (449,81) {$\scalebox{1.2}{\ensuremath{\cdot}}$}
    \put (454,81) {$\scalebox{1.2}{\ensuremath{\cdot}}$}
\end{overpic}
\vspace{-.2em}
\caption{Network architecture of \textbf{Synchronized Deformation Diffusion.}
}
\label{fig:deform}
\end{figure*}

\subsection{Shape Diffusion}
In the shape diffusion part, we leverage the first frame of the sequence, $\mathbf{P}^1$, to reconstruct the object shape. As illustrated in Fig.~\ref{fig:shape}, this process is divided into two distinct networks: (a) \textit{Shape Autoencoder Network} and (b) \textit{Shape Vector Set Diffusion Network}. 
% The yellow area refers to the Shape Vector Set Diffusion Network.
% \vspace*{-2mm}
\paragraph{Shape Autoencoder} To optimize computational efficiency, we adopt the furthest point sampling (FPS) technique. This method pinpoints crucial points within a point cloud, thereby thinning its density.:
\begin{align}
\mathbf{X}^d &= \operatorname{FPS}(\mathbf{X} )
\end{align}
Subsequently, a cross-attention block, designed to compute attention weights across various points, is employed. This block fuses the features of the subsampled points and generates a shape latent set, denoted as  \( \mathcal{S} = \{\mathbf{s}_i \in \mathbb{R}^C\}_{i=1}^M \). Here, \( M \) represents the overall count of codes and \( C \) denotes their dimensionality. In this process, the positional embeddings derived from the points after FPS sampling are utilized as the query, whereas those obtained before FPS sampling serve as the key and value in the attention mechanism. Also, consistent with the latent diffusion framework proposed by \cite{rombach2022high}, our model incorporates KL-regularization within the latent space. This regularization strategy plays a crucial role in modulating feature diversity, ensuring the preservation of high-level features. The query points are encoded and passed to the cross-attention block with the generated shape code. The resulting fused code is then mapped to a dimension of 1 via a fully connected (FC) layer, providing the predicted occupancy value for the query points. 
% \vspace*{-5mm}
\paragraph{Shape Vector Set Diffusion} Nosied shape codes $\hat{\mathcal{S}}$ are sent to the denoising neural network. The denoiser consists of two blocks: The space-attention block facilitates positional information exchange among $M$ codes in different positions, while the condition-attention block injects information from sparse or partial points (conditional input). After repeating this process, we get the denoised latent set $\mathcal{S}$.

\subsection{Synchronized Deformation Diffusion}
% \vspace*{-2mm}
As shown in Fig.~\ref{fig:deform}, the deformation diffusion also contains two parts: \textit{Deformation Autoencoder Network} and \textit{Synchronized Deformation Vector Set Diffusion Network}. 
% The yellow area in Fig.~\ref{fig:deform} refers to the denoising neural network.

% \vspace*{-2mm}
\paragraph{Deformation Autoencoder} In the deformation autoencoder, both surface and near-surface points of different frames are sampled according to the same face indexes, which ensures the correspondence within a sequence. These point cloud frames are pairwise paired together, each with a source point cloud $\mathbf{X}_{\text{src}}$ and a target point cloud $\mathbf{X}_{\text{tgt}}$. 

Similarly to shape diffusion, to ensure the correspondence between the source point cloud and the target point cloud, we use the same FPS for downsampling. We have:
\begin{align}
\mathbf{X}_{\text{src}}^d &= \operatorname{FPS}(\mathbf{X}_{\text{src}} )\\
\mathbf{X}_{\text{tgt}}^d &= \operatorname{FPS}(\mathbf{X}_{\text{tgt}} )
\end{align}
As shown in Fig.~\ref{fig:deform} $(a)$, after obtaining the key points of the source and target point clouds. Positional embeddings of the FPS downsampled and original point clouds are concatenated along the last dimension to preserve spatial consistency, where PosEmb: $: \mathbb{R}^3 \rightarrow \mathbb{R}^M$ refers to positional embedding functions:
\begin{align}
\mathbf{PE} & = \text{Concate}\left( \operatorname{PosEmb}(\mathbf{X}_{\text{src}}), \operatorname{PosEmb}(\mathbf{X}_{\text{tgt}}) \right) \\
\mathbf{PE}^d & = \text{Concate}\left( \operatorname{PosEmb}(\mathbf{X}^d_{\text{src}}), \operatorname{PosEmb}(\mathbf{X}^d_{\text{tgt}}) \right)
\end{align}

We employ the cross attention \(\operatorname{CrossAttn}(Q, KV)\) throughout our method, where \(Q\) denotes the query, and \(KV\) denotes the key-value pair. Similarly, we use KL-divergence to retain high-level features to facilitate the learning of the diffusion model. Here we get the deformation latent set $\mathcal{D} = \{ \mathbf{d}_i \in \mathbb{R}^C \}_{i=1}^M$:
\begin{equation}
\mathcal{D}(\mathbf{X}_{\text{src}},\mathbf{X}_{\text{tgt}}) = \operatorname{CrossAttn}(\mathbf{PE}, \mathbf{PE}^d)
\end{equation}
At inference time, we take the surface points of the source mesh as query points. The interaction between the predicted deformation codes sampled from the \textit{Shape Vector Set Diffusion Network}, and the positional embedding of query points through cross-attention yields the approximated latent features. Subsequently, a linear layer derives the predicted positions of the target point cloud.

\paragraph{Synchronized Deformation Vector Set Diffusion} In this stage, the sparse noisy input point clouds $\{\mathbf{P}_{\text{src}}, \mathbf{P}_{\text{tgt}}\}$ pairs green blocks in Fig.~\ref{fig:deform} $(b)$ are processed with a conditional encoder. The conditional encoder has the same structure with the deformation encoder, through which we get the conditional latent set $\mathcal{C}^t(\mathbf{P}^{1}, \mathbf{P}^{t}) = \{\mathbf{c}_i \in \mathbb{R}^C\}_{i=1}^M$. Simultaneously, the shape latent set from the deformation encoder is added with Gaussian noise and sent to the denoiser of the diffusion model.
% namely the Synchronized Deformation Diffusion. 
As illustrated in the figure, after the space self-attention block, the conditional latent set is injected into the cross-attention block. Following the cross-attention along the temporal domain, and after 18 repetitions, we get the denoised deformation codes.

\section{Implementation Details}
\label{sec:implementation_details}

\subsection{Dataset}
% \vspace*{-1.5mm}
\paragraph{Train/Val/Test Split} 
The datasets used in our method, D-FAUST \cite{DFAUST} and DT4D-A \cite{DeformingThings4D}, encompass a diverse range of human and animal motions, respectively. D-FAUST includes human motions like ‘‘chicken wings’’, ‘‘shake shoulders’’, and ‘‘shake hips.’’ In contrast, DT4D-A features animal animations such as ‘‘bear3EP death’’, ‘‘bunnyQ walk’’, and ‘‘deer2MB rotate.’’ Following the approach in NSDP \cite{tang2022neural} and CaDeX \cite{Lei2022CaDeX}, we organize our train and validation sets with data from seen identities performing seen motions. The test set is divided into two categories: unseen motions of seen identities and seen motions of unseen identities. Specifically, for DT4D-A \cite{DeformingThings4D}, our training set comprises 835 sequences, the validation set includes 59 sequences, and the test set is divided into 89 sequences for unseen motions and 108 sequences for unseen identities. Similarly, in D-FAUST \cite{DFAUST}, the training set contains 104 sequences, the validation set has 5 sequences, and the test set includes 9 sequences for unseen motions and 11 sequences for unseen identities.

\paragraph{Data Processing}
Our data processing strategy is designed to facilitate the learning of shape encoding and deformation. We utilize two distinct datasets for this purpose. 

For shape encoding, our process begins with the application of the Butterfly Subdivision method \cite{zorin1996interpolating}, an interpolating subdivision technique widely used in computer graphics for generating smooth surfaces from polygonal meshes. Following the methodology outlined in \cite{Occupancy_Networks}, we center and rescale all meshes to ensure that the bounding box of each mesh is centered at the origin $(0,0,0)$ and the longest edge is normalized to a length of 1. We then uniformly sample 200k points within this normalized cube and compute their occupancy values to determine whether they lie inside or outside the mesh, as detailed in \cite{Occupancy_Networks}. To enhance our model's understanding of surface properties, we introduce Gaussian noise at two different levels to the mesh surface points. This process generates 200k near-surface points, whose occupancy values are also computed. These points, combined with an additional 200k points sampled directly from the mesh surface, constitute the input to our network, providing a comprehensive set of data points for both the uniform space and near-surface regions.

For deformation encoding, we depart from using the Butterfly Subdivision method \cite{zorin1996interpolating}, which was applied in the context of shape encoding. This choice is primarily driven by the need to maintain spatial correspondence between points. To achieve this, we directly sample 200k surface points from the mesh. Subsequently, we sample an equal number of near-surface points. These near-surface points are generated along the normal direction of each surface point, with a predefined distance that ensures closeness to the surface while preserving the detail of the mesh structure. Crucially, both sets of points are selected based on the same face indices of the mesh. This methodical selection guarantees that the spatial correspondence is not disrupted, allowing for a more accurate representation of deformation.

\begin{figure}[t]
  \centering
  % \begin{overpic}[width=\linewidth,grid,unit=1bp,tics=10]{fig/x_supp/ShapeAE.jpg}
  \begin{overpic}[width=\linewidth]{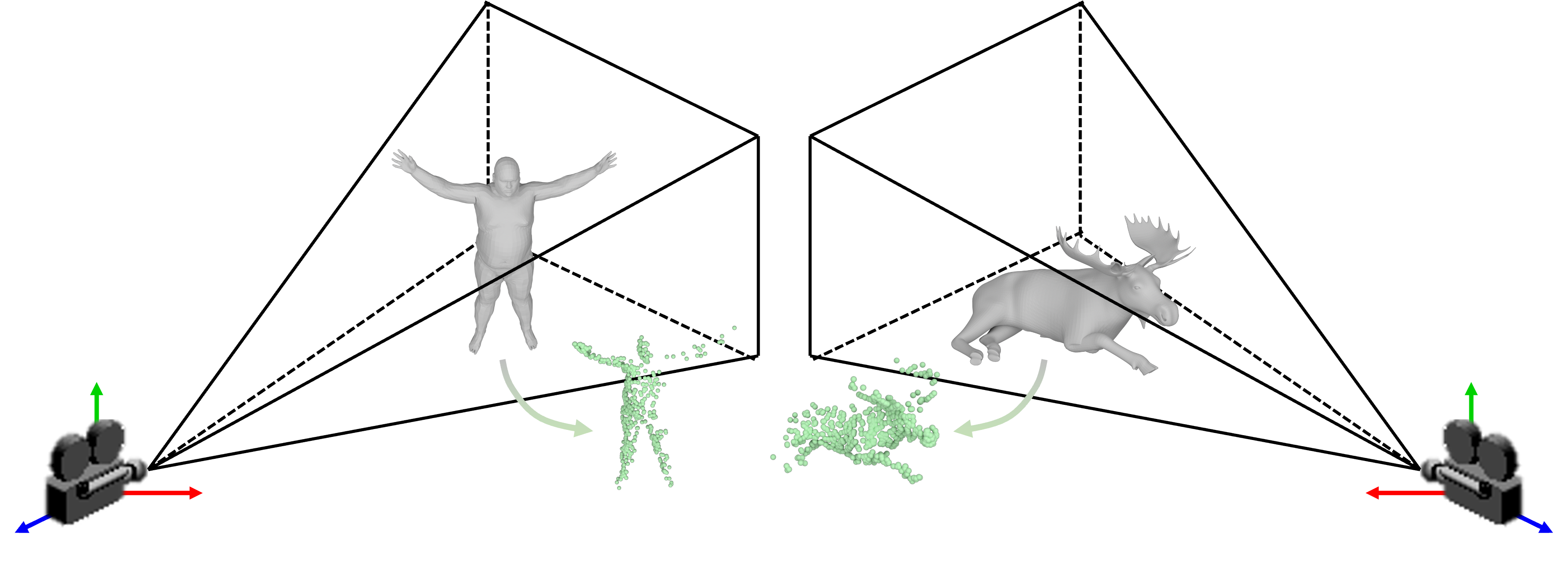}
  \put (0,0) {$\textcolor[RGB]{0, 0, 240}{\text{z}}$}
  \put (28,6) {$\textcolor[RGB]{250, 0, 0}{\text{x}}$}
  \put (13,35) {$\textcolor[RGB]{0, 210, 0}{\text{y}}$}

  \put (85, 4) {\footnotesize\text{Partial observation}}

\put (231,0) {$\textcolor[RGB]{0, 0, 240}{\text{z}}$}
  \put (205,6) {$\textcolor[RGB]{250, 0, 0}{\text{x}}$}
  \put (220,35) {$\textcolor[RGB]{0, 210, 0}{\text{y}}$}

  \end{overpic}
  \caption{The camera setup for the generation of partial observation in D-FAUST \cite{DFAUST} and DT4D-A \cite{DeformingThings4D} dataset.}
  \label{fig:cam}
  % \vspace{-.3cm}
\end{figure}
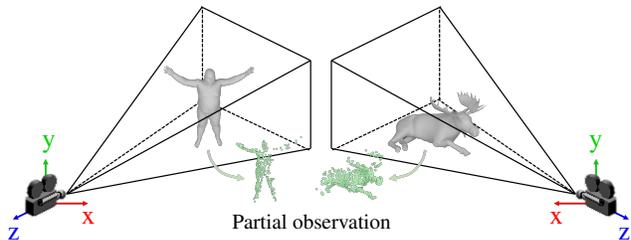

Within the context of the partial challenge configuration, a camera was positioned with a fixed viewing angle of 45 degrees directed towards the human or animal subject within the scene. This positioning was undertaken to capture a partial depth observation of the subject, as illustrated in Figure \ref{fig:cam}. Additionally, in the more restricted partial setting, only half of the human body was observed. This was accomplished by intentionally selecting vertex indices corresponding to the upper body of the SMPL model \cite{SMPL}.

\section{Additional Results}
\label{sec:more_res}
% \vspace*{-1mm}

\subsection{Effectiveness of diffusion models}

\begin{figure}[t]
  \centering
  % \begin{overpic}[width=\linewidth,grid,unit=1bp,tics=10]{fig/x_supp/behave_real_full.jpg}
  \begin{overpic}[width=\linewidth]{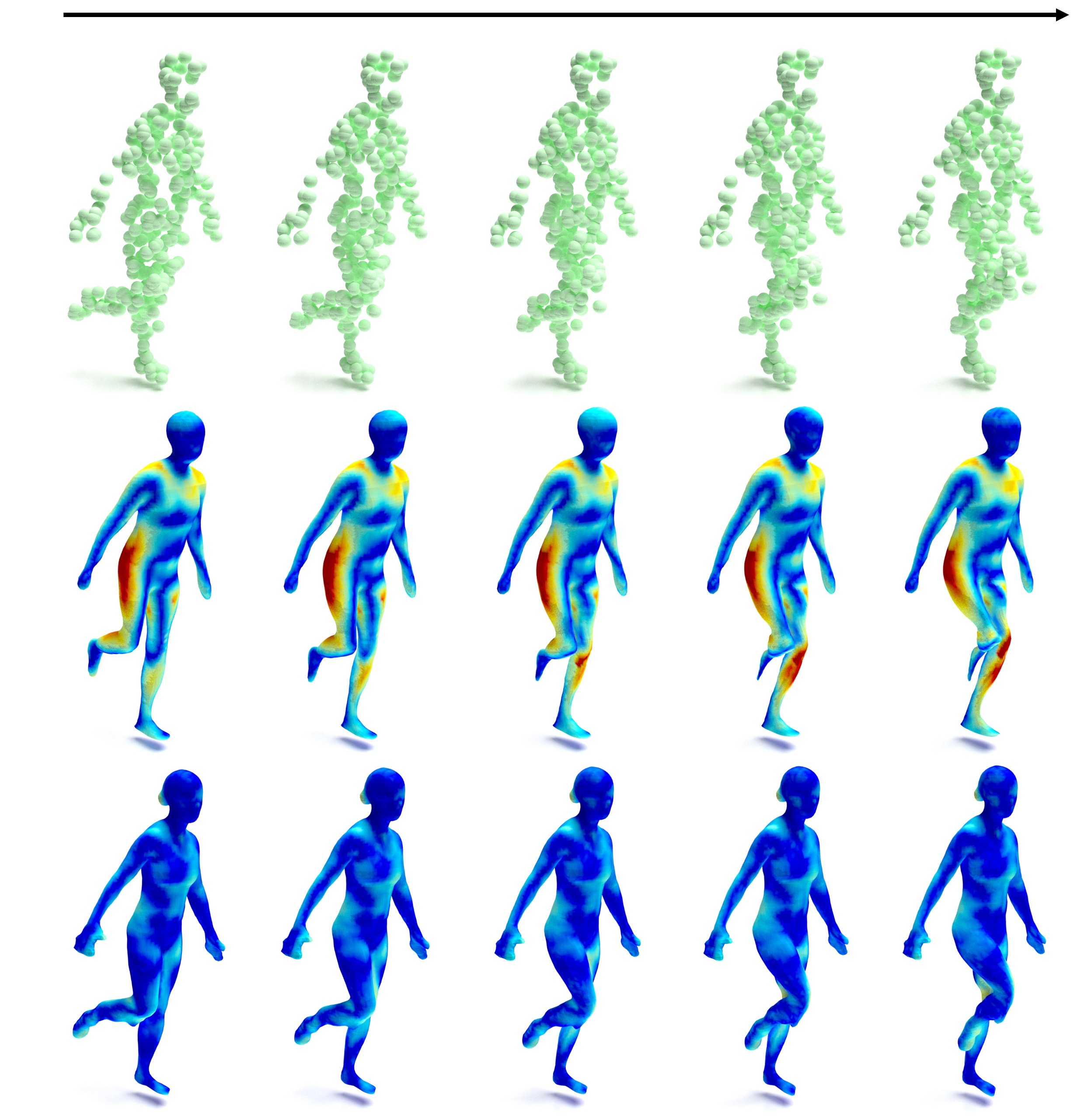}
  \put (2.5, 242) {\rotatebox{-90}{t}}
  \put (1, 206) {\rotatebox{-90}{Input}}
  \put (1, 144) {\rotatebox{-90}{W/o. diffusion}}
  \put (1,46) {\rotatebox{-90}{Full}}
  \end{overpic}
% \vspace*{-7mm}
  \caption{Qualitative comparisons of ablation study on the diffusion model from \textbf{sparse and noisy} point clouds on D-FAUST \cite{DFAUST}.  Our method with diffusion model exhibits lower errors.}
  \label{fig:ablation_wo_diff}
\vspace*{-4mm}
\end{figure}
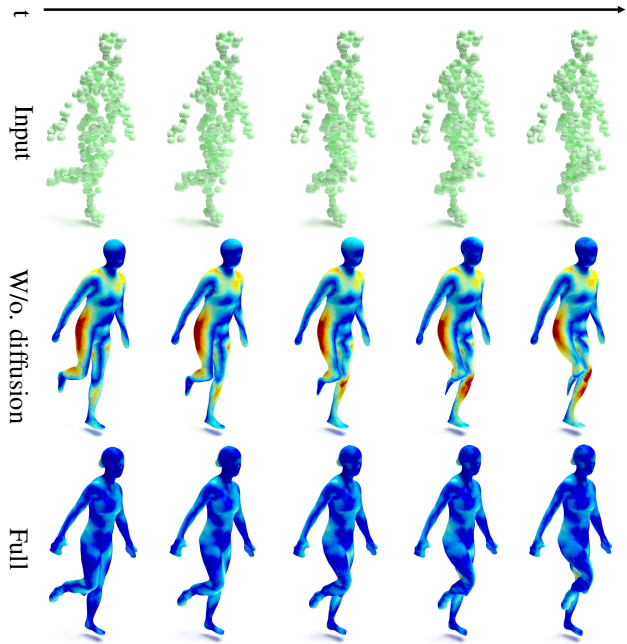

To further demonstrate the efficacy of the diffusion model, we expanded our ablation study to include 4D Shape Reconstruction from sparse and noisy point cloud sequences, utilizing the D-FAUST dataset \cite{DFAUST}. This study provides a more comparative analysis against variants of one-step regression integrated with diffusion models, where the input comprises sequences of point clouds of size $L = 300$. Fig.~\ref{fig:ablation_wo_diff} and Tab.~\ref{tab:wo_diff} present both quantitative and qualitative comparisons. The results clearly indicate that incorporating the diffusion model significantly reduces reconstruction error and yields more precise motion outputs.

\begin{table}[t]
% \vspace{-3.5mm}
\centering
\scalebox{0.8}{
\renewcommand{\arraystretch}{0.75} 
\begin{tabular}{@{}lccc|ccc@{}}
\toprule
\multicolumn{1}{c}{\multirow{2}{*}{D-FAUST}} & \multicolumn{3}{c|}{Unseen Motion}              & \multicolumn{3}{c}{Unseen Individual}           \\ \cmidrule(l){2-7} 
\multicolumn{1}{c}{}            & IoU↑   & CD↓   & Corr↓ & IoU↑   & CD↓   & Corr↓ \\ \midrule
\multicolumn{1}{c|}{W/o. diffusion}               & 79.1\% & 0.070 & 0.076 & 69.3\% & 0.089 & 0.098 \\
\multicolumn{1}{c|}{Full}             
 & \textbf{90.7\%} & \textbf{0.033} & \textbf{0.047} & \textbf{83.7\%} & \textbf{0.045} & \textbf{0.064} \\   
\bottomrule
\end{tabular}
}
%\vspace{-3.5mm}
\caption{Quantitative comparisons of ablation study on the diffusion model from \textbf{sparse and noisy} point clouds on D-FAUST \cite{DFAUST}.}
% \vspace{-4.2mm}
\label{tab:wo_diff}
\end{table}

\subsection{4D Shape Reconstruction}
\begin{figure*}[t]
  \centering
  % \begin{overpic}[width=\linewidth,grid,unit=1bp,tics=10]{fig/x_supp/ShapeAE.jpg}
  \begin{overpic}[width=\linewidth]{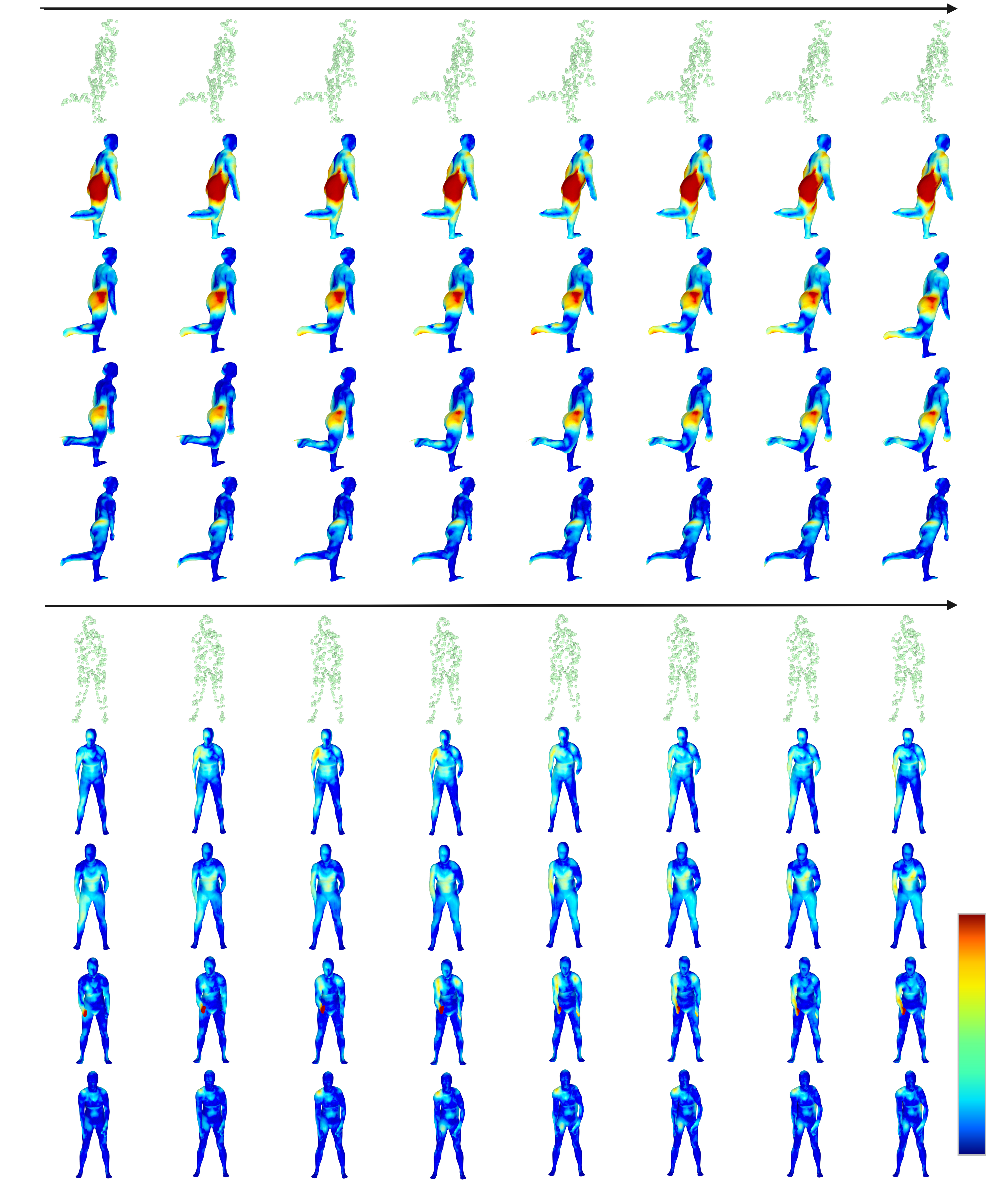}
      \put (13,50) {$\large\rotatebox{-90}{Ours}$}
    \put (10,354) {$\large\rotatebox{-90}{Ours}$}
    \put (13,115) {$\large\rotatebox{-90}{CaDeX}$}
    \put (10,416) {$\large\rotatebox{-90}{CaDeX}$}
    \put (13,169) {$\large\rotatebox{-90}{LPDC}$}
    \put (10,472) {$\large\rotatebox{-90}{LPDC}$}
    \put (13,228) {$\large\rotatebox{-90}{OFlow}$}
    \put (10,533) {$\large\rotatebox{-90}{OFlow}$}
    \put (13,283) {$\large\rotatebox{-90}{Input}$}
    \put (10,584) {$\large\rotatebox{-90}{Input}$}
    % \put (15,286) {$\large\rotatebox{-90}{Sparse}$}
    % \put (10,587) {$\large\rotatebox{-90}{Sparse}$}
    \put (10,303) {$\large\rotatebox{-90}{t}$}
    \put (10,604.5) {$\large\rotatebox{-90}{t}$}
    \put (486, 12) {\Large$0$}
    \put (480, 150) {\Large$0.4$}
  \end{overpic}
\caption{\textbf{4D Shape Reconstruction from sparse and noisy point clouds on the D-FAUST\cite{DFAUST} dataset}. One for unseen motion (upper) and another for unseen individuals (lower).}
%We provide two examples of unseen motion (upper part) and unseen individual (lower part) on 4D Shape Reconstruction from \textbf{sparse point cloud} task on D-FAUST \cite{DFAUST} dataset.
\label{fig:human_sparse}
\end{figure*}

\begin{figure*}[t]
  \centering
  % \begin{overpic}[width=\linewidth,grid,unit=1bp,tics=10]{fig/x_supp/ShapeAE.jpg}
  \begin{overpic}[width=\linewidth]{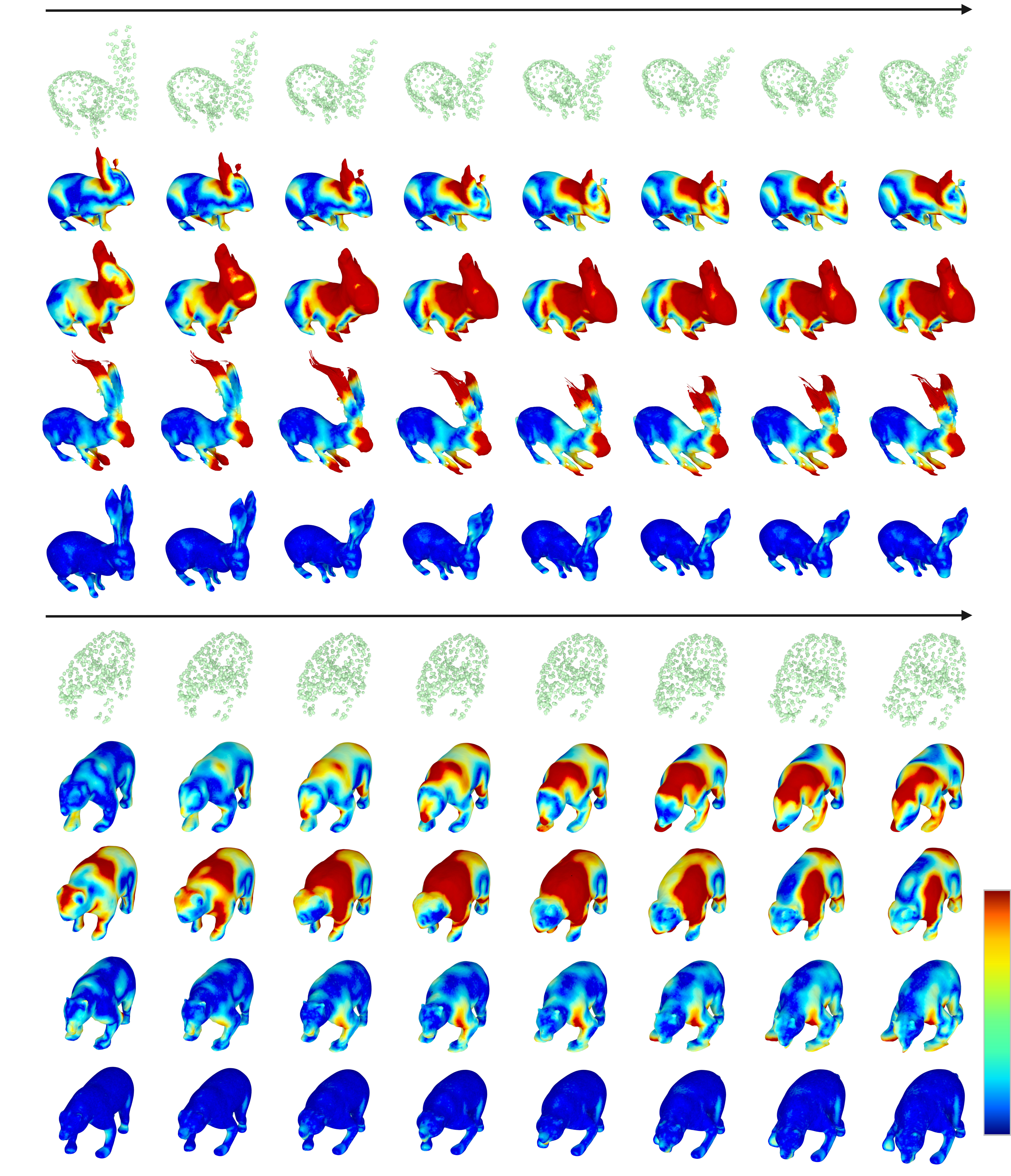}
      \put (10,40) {$\large\rotatebox{-90}{Ours}$}
    \put (5,320) {$\large\rotatebox{-90}{Ours}$}
    \put (10,105) {$\large\rotatebox{-90}{CaDeX}$}
    \put (5,387) {$\large\rotatebox{-90}{CaDeX}$}
    \put (10,155) {$\large\rotatebox{-90}{LPDC}$}
    \put (5,445) {$\large\rotatebox{-90}{LPDC}$}
    \put (10,205) {$\large\rotatebox{-90}{OFlow}$}
    \put (5,500) {$\large\rotatebox{-90}{OFlow}$}
    \put (10,255) {$\large\rotatebox{-90}{Input}$}
    \put (5,552) {$\large\rotatebox{-90}{Input}$}
    % \put (15,258) {$\large\rotatebox{-90}{Sparse}$}
    % \put (5,555) {$\large\rotatebox{-90}{Sparse}$}
    \put (10,276) {$\large\rotatebox{-90}{t}$}
    \put (5,575) {$\large\rotatebox{-90}{t}$}
    \put (486.8, 7) {\Large$0$}
    \put (480, 145) {\Large$0.4$}
  \end{overpic}
\caption{
\textbf{4D Shape Reconstruction from sparse and noisy point clouds on the DT4D-A \cite{DeformingThings4D} dataset.} One for unseen motion (upper) and another for unseen individuals (lower).
%\textbf{4D Shape Reconstruction (DT4D-A).} The figure shows two examples of unseen motion (upper part) and unseen individual (lower part) on 4D Shape Reconstruction from \textbf{sparse point cloud} task on DT4D-A \cite{DeformingThings4D} dataset.
}
\label{fig:animal_sparse}
\end{figure*}

% \vspace*{-1.5mm}
In evaluating our model's performance in 4D shape reconstruction, we adopted a comprehensive set of metrics: Intersection over Union (IoU), chamfer distance, and correspondence distance. These metrics were chosen for their relevance in accurately quantifying shape reconstruction quality. IoU measures the overlap between predicted and ground truth shapes, Chamfer distance quantifies the average closest point distance, and correspondence distance evaluates the accuracy of point-wise correspondences. These metrics align with the standards set in recent studies, such as those by LPDC \cite{LPDC} and OFlow \cite{OccupancyFlow}.
We present the average metrics across all 17 frames for both D-FAUST \cite{DFAUST} and DT4D-A \cite{DeformingThings4D} datasets. The quantitative results are shown in \cref{Tab:human_sparse_us}, \cref{Tab:human_sparse_ui}, \cref{Tab:animal_sparse_us}, and \cref{Tab:animal_sparse_ui}. Our analysis reveals that our model demonstrates superior generalization capabilities in scenarios involving both unseen motions and unseen individuals. Specifically, it outperforms existing approaches, including OFlow \cite{OccupancyFlow}, LPDC \cite{LPDC}, and CaDeX \cite{Lei2022CaDeX}, across all evaluated metrics for all time frames. This advancement underscores the efficacy of our approach in handling the complexities of 4D shape reconstruction.
Furthermore, qualitative results in \cref{fig:human_sparse} and \cref{fig:animal_sparse} show a selection of 8 frames, chosen to represent a diverse range of motions and shapes, from the total 17 to illustrate our model's performance. In each figure, the upper part displays the results for unseen motion, while the lower part corresponds to unseen individuals. We utilize a chamfer distance error map for visualization, where blue indicates lower error and red signifies higher error. The color-coded error map, computed based on the distance between predicted and ground truth points, provides an intuitive understanding of the model's accuracy in different scenarios. Our model not only has an overall smaller error on both datasets, but also captures motions more accurately.
\begin{figure}[t]
  \centering
  % \begin{overpic}[width=\linewidth]{fig/x_supp/real_dataset.png}
  %   \put (-2, 45) {\rotatebox{-90}{Ours}}
  %   \put (-2, 140) {\rotatebox{-90}{Partial input}}
  %   \put (-2, 220) {\rotatebox{-90}{Reference}}
  % \end{overpic}
  % \begin{overpic}[width=\linewidth,grid,unit=1bp,tics=10]{fig/x_supp/behave_real_full.jpg}
  \begin{overpic}[width=\linewidth]{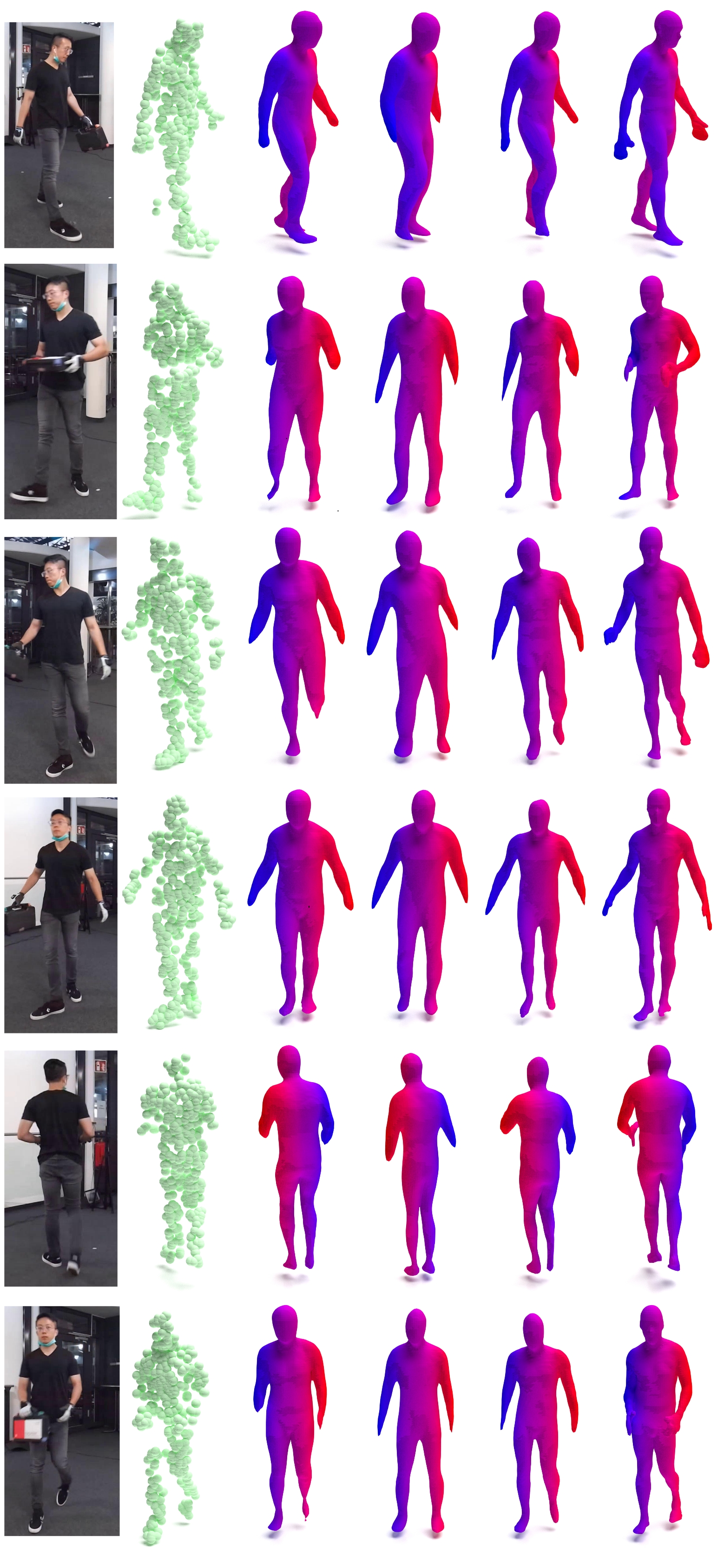}
  \put (8, 517) {RGB}
  \put (43, 524) {Partial}
  \put (45, 515) {Input}
  \put (83, 517) {OFlow}
  \put (123, 517) {LPDC}
  \put (161, 517) {CaDeX}
  \put (206, 517) {Ours}
  \end{overpic}
    \vspace{-24pt}
  \caption{4D Shape Completion on the BEHAVE dataset \cite{bhatnagar22behave}.} 
  % The input, consisting of $300$ points, is sub-sampled from a point cloud back-projected from the depth frame. Our approach successfully recovers accurate 4D human motion captured in real-world scenarios, showcasing its generalization ability to real data.}
    \vspace{-18pt}
  \label{fig:real}
\end{figure}
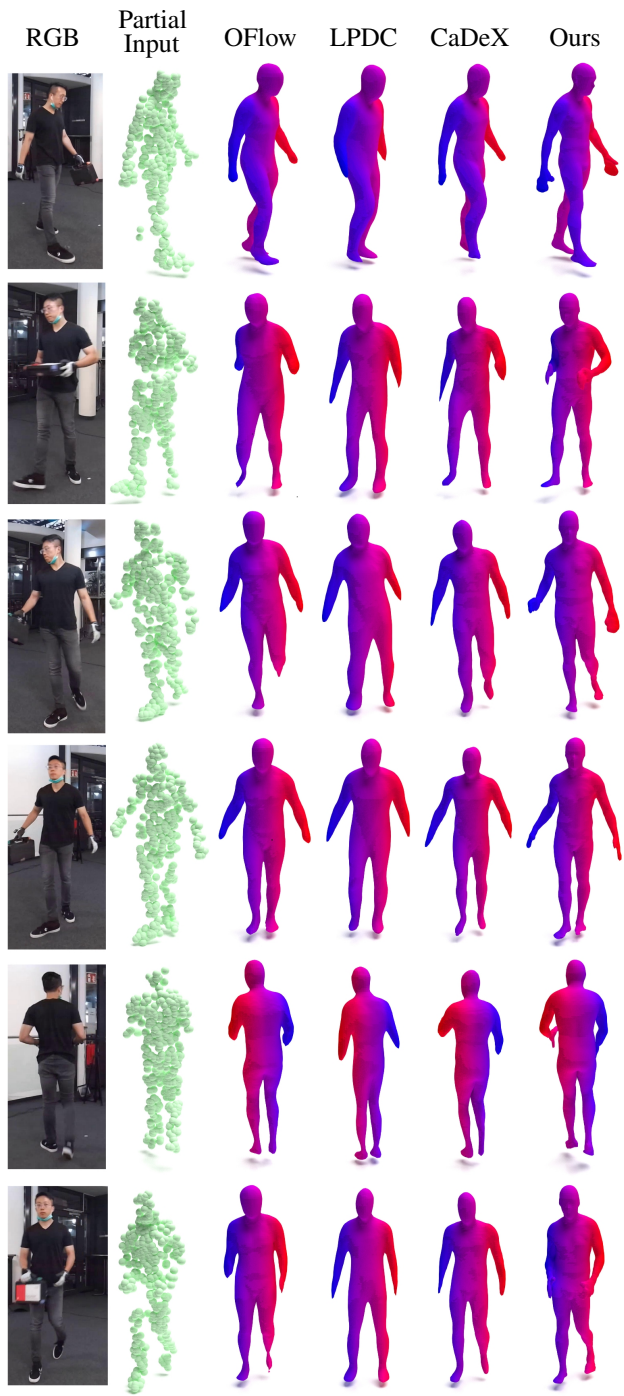

\begin{figure}[t]
  \centering
  % \begin{overpic}[width=\linewidth,grid,unit=1bp,tics=10]{fig/x_supp/behave_real_full.jpg}
    % \vspace{-10pt}
  
  \begin{overpic}[width=\linewidth]{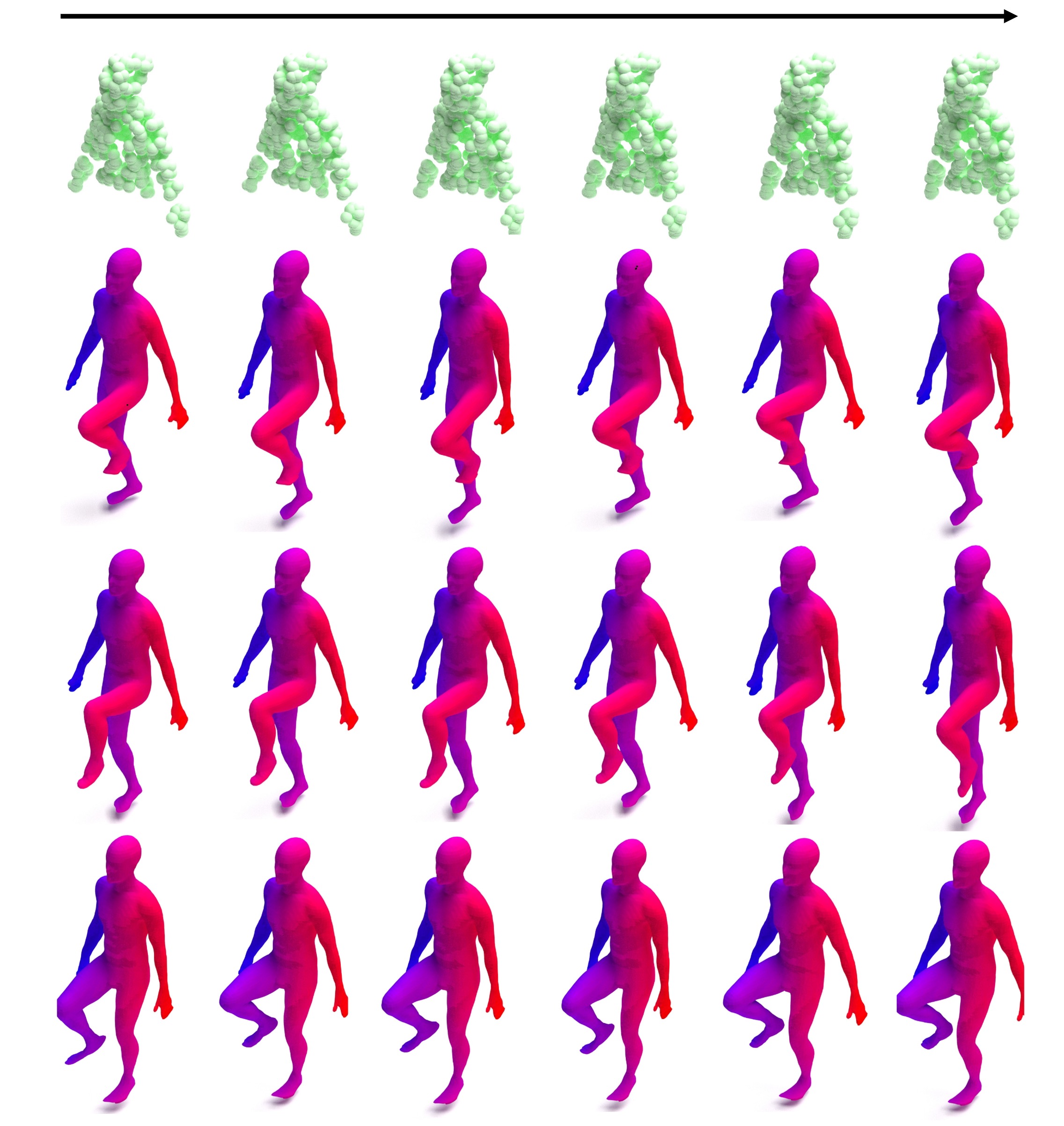}
  \put (2, 248) {\rotatebox{-90}{t}}
  \put (0, 230) {\rotatebox{-90}{Input}}
  \put (0, 180) {\rotatebox{-90}{Ours 1}}
  \put (0, 115) {\rotatebox{-90}{Ours 2}}
  \put (0, 50) {\rotatebox{-90}{Ours 3}}
  \end{overpic}
  \caption{More results of 4D Shape Completion from highly partial point clouds (half human) on the D-FAUST \cite{DFAUST} dataset.}
  \label{fig:more_half}
\end{figure}
\begin{table}[t]
\centering

\scalebox{1}{
\renewcommand{\arraystretch}{1} 
    \begin{tabular}{@{}lcccccc@{}}
    \toprule
    \multicolumn{1}{c}{BEHAVE}     & OFlow   & LPDC   & CaDeX & Ours   \\ \midrule
    \multicolumn{1}{c}{Chamfer↓}   & 0.137 & 0.201 & 0.126 & \textbf{0.062}  \\  
    \bottomrule
    \end{tabular}
}

\caption{Quantitative evaluation on real dataset BEHAVE \cite{bhatnagar22behave}. The chamfer distance is computed from the reconstructed mesh and partial point cloud input.}
\label{tab:behave}
\end{table}

\subsection{4D Shape Completion}
\begin{figure*}[t]
  \centering
  % \begin{overpic}[width=\linewidth,grid,unit=1bp,tics=10]{fig/x_supp/ShapeAE.jpg}
  \begin{overpic}[width=\linewidth]{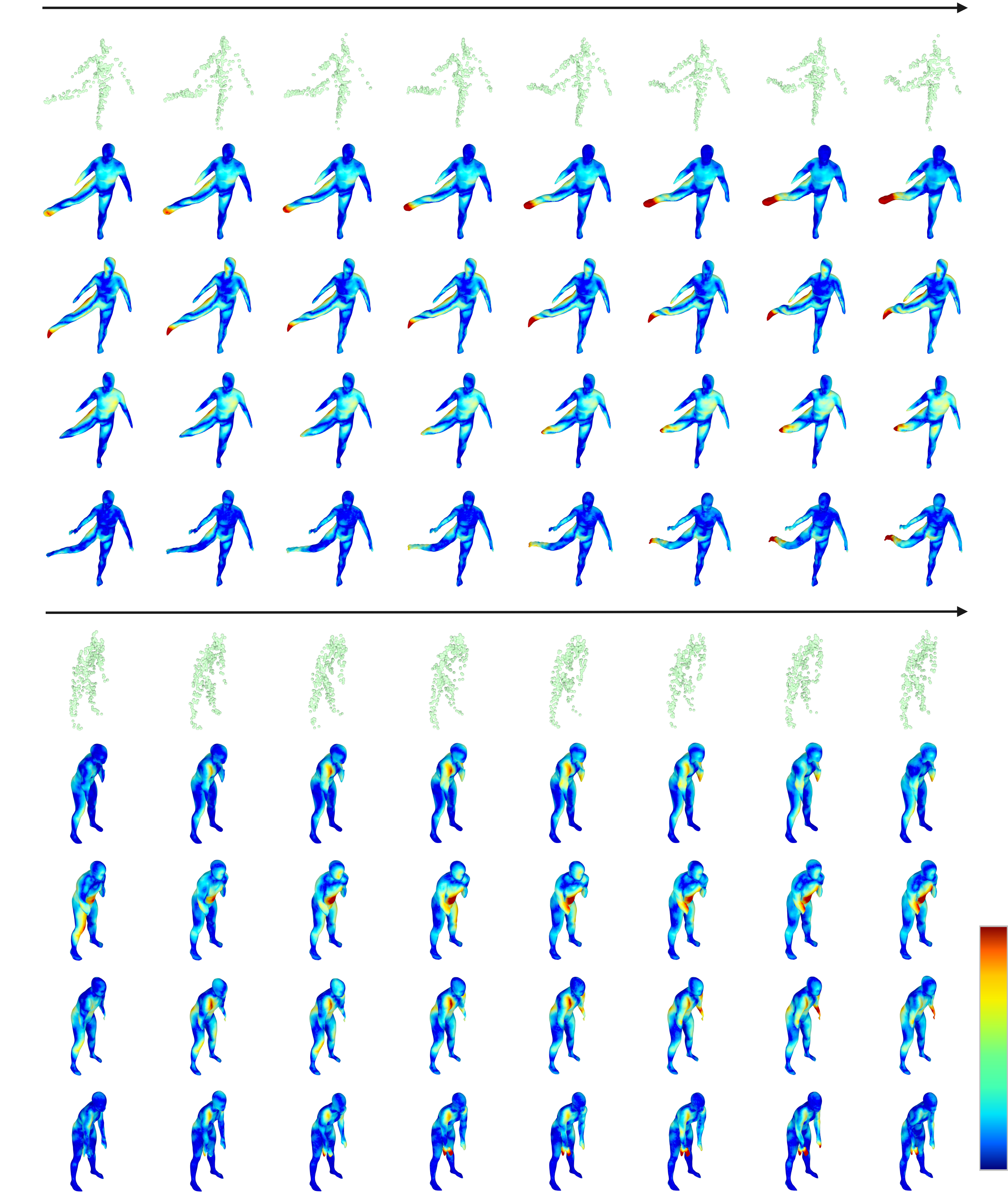}
    \put (13,40) {$\large\rotatebox{-90}{Ours}$}
    \put (10,340) {$\large\rotatebox{-90}{Ours}$}
    \put (13,105) {$\large\rotatebox{-90}{CaDeX}$}
    \put (10,405) {$\large\rotatebox{-90}{CaDeX}$}
    \put (13,160) {$\large\rotatebox{-90}{LPDC}$}
    \put (10,455) {$\large\rotatebox{-90}{LPDC}$}
    \put (13,218) {$\large\rotatebox{-90}{OFlow}$}
    \put (10,515) {$\large\rotatebox{-90}{OFlow}$}
    \put (13,270) {$\large\rotatebox{-90}{Input}$}
    \put (8,567) {$\large\rotatebox{-90}{Input}$}
    % \put (15,273) {$\large\rotatebox{-90}{Partial}$}
    % \put (8,570) {$\large\rotatebox{-90}{Partial}$}
    \put (10,292) {$\large\rotatebox{-90}{t}$}
    \put (10,590.5) {$\large\rotatebox{-90}{t}$}
    \put (486.8, 2.5) {\Large$0$}
    \put (480, 140.5) {\Large$0.4$}
  \end{overpic}
\caption{
\textbf{4D Shape Completion from partial point clouds on the D-FAUST\cite{DFAUST} dataset}. One for unseen motion (upper) and another for unseen individuals (lower).
%\textbf{4D Shape Completion (D-FAUST).} The figure shows two examples of unseen motion (upper part) and unseen individual (lower part) on 4D Shape Completion from \textbf{partial point cloud} task on D-FAUST \cite{DFAUST} dataset.
}
\label{fig:human_partial}

\end{figure*}

\begin{figure*}[t]
  \centering
  \begin{overpic}[width=\linewidth]{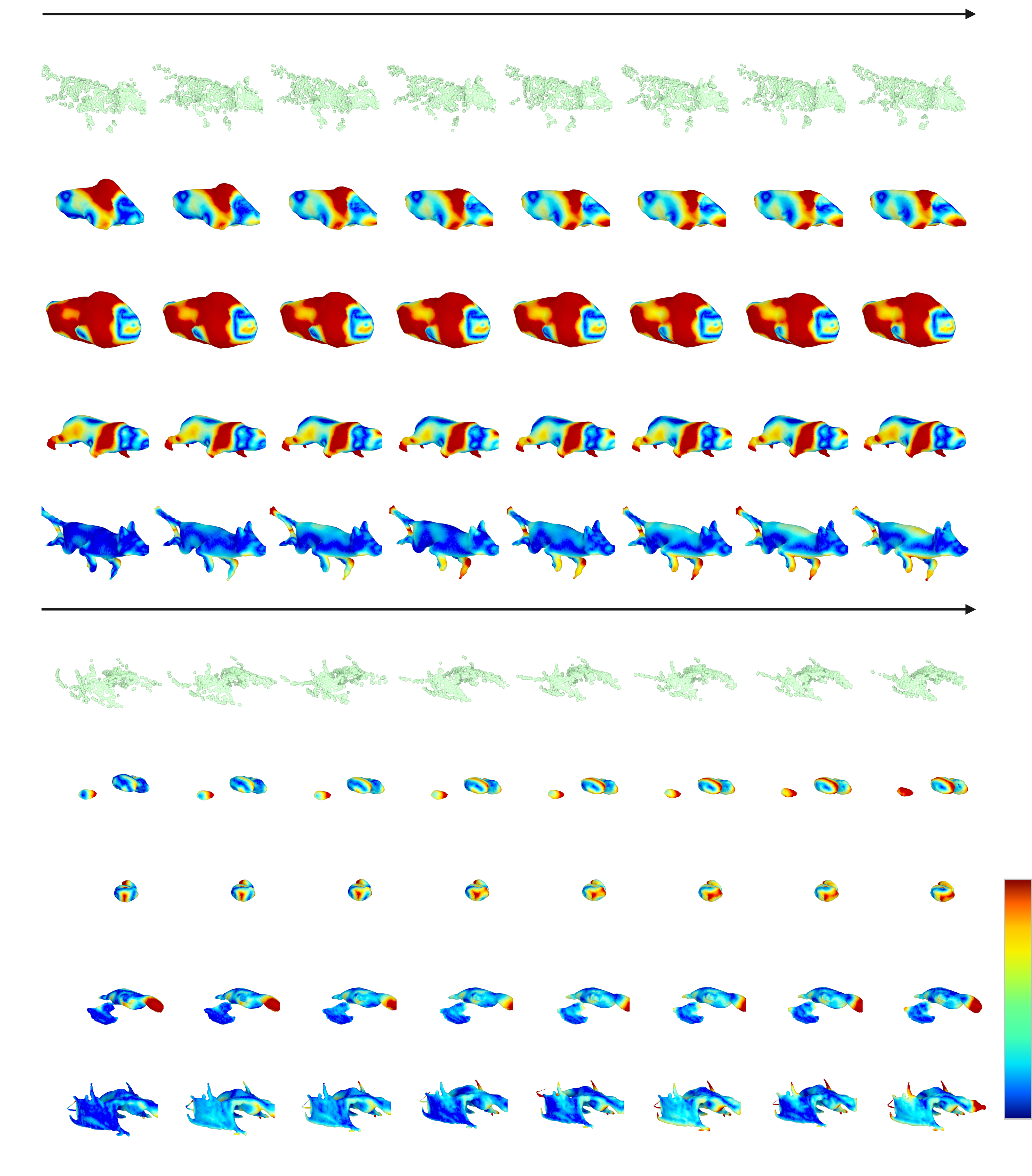}
    \put (13,45) {$\large\rotatebox{-90}{Ours}$}
    \put (5,318) {$\large\rotatebox{-90}{Ours}$}
    \put (13,105) {$\large\rotatebox{-90}{CaDeX}$}
    \put (4,378) {$\large\rotatebox{-90}{CaDeX}$}
    \put (13,155) {$\large\rotatebox{-90}{LPDC}$}
    \put (5,430) {$\large\rotatebox{-90}{LPDC}$}
    \put (13,205) {$\large\rotatebox{-90}{OFlow}$}
    \put (4,485) {$\large\rotatebox{-90}{OFlow}$}
    \put (13,253) {$\large\rotatebox{-90}{Input}$}
    \put (5,538) {$\large\rotatebox{-90}{Input}$}
    % \put (15,253) {$\large\rotatebox{-90}{Partial}$}
    % \put (5,538) {$\large\rotatebox{-90}{Partial}$}
    \put (10,275) {$\large\rotatebox{-90}{t}$}
    \put (10,561.5) {$\large\rotatebox{-90}{t}$}
    \put (486.8, 15) {\Large$0$}
    \put (480, 147) {\Large$0.4$}
  \end{overpic}
\caption{
\textbf{4D Shape Completion from partial point clouds on the DT4D-A~\cite{DeformingThings4D} dataset.} One for unseen motion (upper) and another for unseen individuals (lower).
}
%\textbf{4D Shape Completion (DT4D-A).} The figure shows two examples of unseen motion (upper part) and unseen individual (lower part) on 4D Shape Completion from \textbf{partial point cloud} task on DT4D-A \cite{DeformingThings4D} dataset.}
\label{fig:animal_partial}
\end{figure*}

\vspace*{-1.5mm}
In the more challenging task of 4D shape completion from partial point clouds, our model shows noticable improvements over existing state-of-the-art methods. The quantitative results, as illustrated in \cref{fig:human_partial} and \cref{fig:animal_partial}, demonstrate substantial performance enhancements, with a lower error rate in scenarios involving unseen motions and individuals. This underscores the robustness of our approach. Also, we present more results about the challenging half human setting. The result is demonstrated in the \cref{fig:more_half}. 

% Our model's capabilities are especially evident in the D-FAUST dataset \cite{DFAUST}, achieving more accurate reconstructions of complex human body parts, like hands and feet. These parts, typically difficult to reconstruct, are rendered with higher fidelity.

\section{Real-world Data Test}
\label{sec:other_app}

\vspace*{-1.5mm}
In this section, we validate our model using data from the real-world BEHAVE dataset \cite{bhatnagar22behave}, employing four Kinect RGB-D cameras to capture RGB color and depth frames. In our case, we utilize a single view from a fixed camera to align with our previous partial scan setting. Similarly, the depth map is back-projected into 3D point cloud, serving as a partial input for our model.
In Fig.~\ref{fig:real} and Tab.~\ref{tab:behave}, we present a qualitative and quantitative evaluation of our reconstruction process with the corresponding input and reference RGB frame. 
The results demonstrate the robustness of our model in scenarios characterized by incomplete scans, such as instances where limbs, like the leg or arm, are obscured by structures, such as a grasped object in the hand. This resilience stems from the inherent capabilities of the diffusion model, empowering our model to infer potential structures even in the presence of significant occlusions. 
% This effectiveness is attributed to the learned latent distribution, highlighting the model's capacity to generalize and extrapolate beyond the limitations imposed by incomplete or obstructed input scans. 
% Notably, our method exhibits remarkable reliability even without the need for model fine-tuning on the real dataset. This stability is particularly evident when addressing inherent noise present in sensor captures, further affirming the efficacy of our approach in handling challenging real-world scenarios.

% {
%     \small
%     \bibliographystyle{ieeenat_fullname}
%     \bibliography{main}
    
% }

\begin{table*}[t]
    % \vspace{5pt}
	\renewcommand\arraystretch{1.1}
	\begin{center}
            \scalebox{0.92}{
		\begin{tabular}{*{14}{c}}
			\toprule
			\multirow{2}*{Time step} & \multicolumn{4}{c}{IoU}   & \multicolumn{4}{c}{Chamfer} & \multicolumn{4}{c}{Correspond.}\\
			\cmidrule(lr){2-5} \cmidrule(lr){6-9} \cmidrule(lr){10-13}

			& OFlow & LPDC & CaDex & Ours & OFlow & LPDC & CaDex & Ours & OFlow & LPDC & CaDex & Ours \\
			\midrule
                \midrule
			0  & 83.1\% & 85.6\% & 89.1\% & \textbf{91.2\%} & 0.059 & 0.052 & 0.044 & \textbf{0.031} & 0.057 & 0.047 & 0.043 & \textbf{0.031} \\
			1  & 83.1\% & 85.5\% & 89.2\% & \textbf{91.2\%} & 0.059 & 0.053 & 0.044 & \textbf{0.031} & 0.062 & 0.053 & 0.047 & \textbf{0.034} \\
			2  & 82.8\% & 85.4\% & 89.3\% & \textbf{91.1\%} & 0.061 & 0.053 & 0.043 & \textbf{0.031} & 0.069 & 0.059 & 0.054 & \textbf{0.036} \\
			3  & 82.5\% & 85.3\% & 89.4\% & \textbf{91.0\%} & 0.061 & 0.053 & 0.043 & \textbf{0.032} & 0.077 & 0.064 & 0.061 & \textbf{0.039} \\
			4  & 82.2\% & 85.0\% & 89.4\% & \textbf{91.0\%} & 0.062 & 0.054 & 0.043 & \textbf{0.032} & 0.083 & 0.070 & 0.068 & \textbf{0.040} \\
			5  & 82.0\% & 85.1\% & 89.5\% & \textbf{90.9\%} & 0.063 & 0.054 & 0.043 & \textbf{0.032} & 0.088 & 0.074 & 0.074 & \textbf{0.043} \\
			6  & 81.8\% & 85.1\% & 89.5\% & \textbf{90.7\%} & 0.064 & 0.054 & 0.043 & \textbf{0.033} & 0.092 & 0.078 & 0.080 & \textbf{0.045} \\
			7  & 81.6\% & 85.1\% & 89.5\% & \textbf{90.6\%} & 0.064 & 0.054 & 0.043 & \textbf{0.032} & 0.095 & 0.082 & 0.085 & \textbf{0.047} \\
			8  & 81.4\% & 85.0\% & 89.5\% & \textbf{90.6\%} & 0.065 & 0.055 & 0.043 & \textbf{0.033} & 0.098 & 0.085 & 0.090 & \textbf{0.048} \\
			9  & 81.3\% & 85.0\% & 89.5\% & \textbf{90.5\%} & 0.066 & 0.055 & 0.043 & \textbf{0.034} & 0.101 & 0.088 & 0.094 & \textbf{0.051} \\
			10 & 81.1\% & 84.6\% & 89.5\% & \textbf{90.3\%} & 0.066 & 0.055 & 0.043 & \textbf{0.034} & 0.103 & 0.090 & 0.097 & \textbf{0.052} \\
			11 & 81.0\% & 84.6\% & 89.5\% & \textbf{90.3\%} & 0.067 & 0.055 & 0.043 & \textbf{0.034} & 0.106 & 0.092 & 0.100 & \textbf{0.054} \\
			12 & 80.8\% & 84.6\% & 89.5\% & \textbf{90.2\%} & 0.068 & 0.056 & 0.043 & \textbf{0.034} & 0.108 & 0.094 & 0.103 & \textbf{0.055} \\
			13 & 80.6\% & 84.4\% & 89.4\% & \textbf{90.2\%} & 0.069 & 0.056 & 0.043 & \textbf{0.034} & 0.111 & 0.095 & 0.105 & \textbf{0.056} \\
			14 & 80.3\% & 84.3\% & 89.3\% & \textbf{90.0\%} & 0.070 & 0.056 & 0.044 & \textbf{0.035} & 0.114 & 0.096 & 0.107 & \textbf{0.057} \\
			15 & 80.0\% & 84.3\% & 89.2\% & \textbf{90.0\%} & 0.071 & 0.056 & 0.044 & \textbf{0.035} & 0.119 & 0.097 & 0.109 & \textbf{0.059} \\
			16 & 79.5\% & 84.3\% & 89.1\% & \textbf{90.0\%} & 0.073 & 0.056 & 0.044 & \textbf{0.035} & 0.125 & 0.098 & 0.110 & \textbf{0.059} \\
			\midrule
            Mean  & 81.5\% & 84.9\% & 89.4\% & \textbf{90.7\%} & 0.065 & 0.055 & 0.043 & \textbf{0.033} & 0.095 & 0.080 & 0.084 & \textbf{0.047} \\
			\bottomrule
		\end{tabular}}
		% \caption{human sparse us }
            \caption{\textbf{4D Shape Reconstruction for Unseen Motions (DFAUST).} We evaluate IoU, Chamfer distance, and correspondence distance for 17 timeframes for the 4D shape reconstruction from sparse point clouds on seen individuals but unseen motions of DFAUST \cite{DFAUST} dataset.}
		\label{Tab:human_sparse_us}
	\end{center}
	% \vspace{5pt}
\end{table*}
\begin{table*}[t]
    % \vspace{5pt}
	\renewcommand\arraystretch{1.1}
	\begin{center}
            \scalebox{0.92}{
		\begin{tabular}{*{14}{c}}
			\toprule
			\multirow{2}*{Time step} & \multicolumn{4}{c}{IoU}   & \multicolumn{4}{c}{Chamfer} & \multicolumn{4}{c}{Correspond.}\\
			\cmidrule(lr){2-5} \cmidrule(lr){6-9} \cmidrule(lr){10-13}

			& OFlow & LPDC & CaDex & Ours & OFlow & LPDC & CaDex & Ours & OFlow & LPDC & CaDex & Ours \\
			\midrule
                \midrule
			0  & 74.2\% & 76.8\% & 80.4\% & \textbf{84.6\%} & 0.077 & 0.068 & 0.055 & \textbf{0.042} & 0.077 & 0.065 & 0.057 & \textbf{0.044} \\
			1  & 74.1\% & 76.8\% & 80.5\% & \textbf{84.6\%} & 0.077 & 0.069 & 0.055 & \textbf{0.042} & 0.082 & 0.071 & 0.060 & \textbf{0.046} \\
			2  & 73.8\% & 76.7\% & 80.6\% & \textbf{84.5\%} & 0.078 & 0.069 & 0.054 & \textbf{0.043} & 0.089 & 0.075 & 0.064 & \textbf{0.048} \\
			3  & 73.4\% & 76.5\% & 80.7\% & \textbf{84.4\%} & 0.079 & 0.069 & 0.054 & \textbf{0.043} & 0.096 & 0.080 & 0.069 & \textbf{0.051} \\
			4  & 73.0\% & 76.5\% & 80.8\% & \textbf{84.4\%} & 0.081 & 0.070 & 0.054 & \textbf{0.043} & 0.102 & 0.085 & 0.074 & \textbf{0.053} \\
			5  & 72.7\% & 76.6\% & 80.8\% & \textbf{84.2\%} & 0.082 & 0.070 & 0.054 & \textbf{0.044} & 0.108 & 0.089 & 0.078 & \textbf{0.056} \\
			6  & 72.4\% & 76.3\% & 80.8\% & \textbf{84.0\%} & 0.083 & 0.070 & 0.054 & \textbf{0.043} & 0.113 & 0.094 & 0.083 & \textbf{0.059} \\
			7  & 72.2\% & 76.2\% & 80.8\% & \textbf{83.8\%} & 0.084 & 0.071 & 0.054 & \textbf{0.045} & 0.117 & 0.098 & 0.086 & \textbf{0.061} \\
			8  & 72.0\% & 76.0\% & 80.8\% & \textbf{83.6\%} & 0.085 & 0.071 & 0.054 & \textbf{0.046} & 0.121 & 0.101 & 0.090 & \textbf{0.065} \\
			9  & 71.9\% & 76.1\% & 80.8\% & \textbf{83.6\%} & 0.085 & 0.071 & 0.054 & \textbf{0.046} & 0.124 & 0.104 & 0.093 & \textbf{0.067} \\
			10 & 71.8\% & 76.1\% & 80.8\% & \textbf{83.5\%} & 0.086 & 0.072 & 0.054 & \textbf{0.046} & 0.127 & 0.107 & 0.096 & \textbf{0.069} \\
			11 & 71.7\% & 75.9\% & 80.7\% & \textbf{83.3\%} & 0.086 & 0.072 & 0.054 & \textbf{0.047} & 0.130 & 0.109 & 0.098 & \textbf{0.072} \\
			12 & 71.5\% & 75.9\% & 80.7\% & \textbf{83.1\%} & 0.087 & 0.072 & 0.054 & \textbf{0.048} & 0.133 & 0.112 & 0.101 & \textbf{0.076} \\
			13 & 71.4\% & 75.8\% & 80.6\% & \textbf{83.1\%} & 0.087 & 0.072 & 0.054 & \textbf{0.048} & 0.136 & 0.114 & 0.103 & \textbf{0.077} \\
			14 & 71.3\% & 75.7\% & 80.5\% & \textbf{83.0\%} & 0.088 & 0.073 & 0.055 & \textbf{0.049} & 0.140 & 0.116 & 0.105 & \textbf{0.079} \\
			15 & 71.0\% & 75.6\% & 80.4\% & \textbf{82.8\%} & 0.089 & 0.073 & 0.055 & \textbf{0.049} & 0.145 & 0.119 & 0.106 & \textbf{0.081} \\
			16 & 70.7\% & 75.7\% & 80.2\% & \textbf{82.8\%} & 0.090 & 0.074 & 0.056 & \textbf{0.049} & 0.150 & 0.121 & 0.107 & \textbf{0.082} \\
   			\midrule
            Mean  & 72.3\% & 76.2\% & 80.6\% & \textbf{83.7\%} & 0.084 & 0.071 & 0.054 & \textbf{0.045} & 0.117 & 0.098 & 0.086 & \textbf{0.064} \\
			\bottomrule
		\end{tabular}}
		% \caption{human sparse ui}
              \caption{\textbf{4D Shape Reconstruction for Unseen Individuals (DFAUST).} We evaluate I0U, Chamfer distance, and correspondence distance for 17 timeframes for the 4D shape reconstruction from sparse point cloud task on unseen individuals of DFAUST \cite{DFAUST} dataset.}
		\label{Tab:human_sparse_ui}
	\end{center}
	% \vspace{5pt}
\end{table*}
\begin{table*}[t]
    % \vspace{5pt}
    \renewcommand\arraystretch{1.1}
    \begin{center}
        \scalebox{0.92}{
        \begin{tabular}{*{14}{c}}
            \toprule
            \multirow{2}*{Time step} & \multicolumn{4}{c}{IoU}   & \multicolumn{4}{c}{Chamfer} & \multicolumn{4}{c}{Correspond.}\\
            \cmidrule(lr){2-5} \cmidrule(lr){6-9} \cmidrule(lr){10-13}
            & OFlow & LPDC & CaDex & \textbf{Ours} & OFlow & LPDC & CaDex & \textbf{Ours} & OFlow & LPDC & CaDex & \textbf{Ours} \\
            \midrule
            \midrule
            0  & 74.8\% & 63.3\% & 79.6\% & \textbf{89.6\%} & 0.191 & 0.302 & 0.063 & \textbf{0.046} & 0.163 & 0.252 & 0.078 & \textbf{0.045} \\
            1  & 74.4\% & 62.6\% & 79.8\% & \textbf{89.6\%} & 0.193 & 0.308 & 0.062 & \textbf{0.047} & 0.182 & 0.285 & 0.082 & \textbf{0.048} \\
            2  & 73.7\% & 62.3\% & 80.0\% & \textbf{89.4\%} & 0.197 & 0.310 & 0.061 & \textbf{0.047} & 0.208 & 0.311 & 0.091 & \textbf{0.051} \\
            3  & 73.1\% & 62.0\% & 80.2\% & \textbf{89.3\%} & 0.202 & 0.313 & 0.060 & \textbf{0.048} & 0.233 & 0.341 & 0.100 & \textbf{0.053} \\
            4  & 72.7\% & 61.6\% & 80.3\% & \textbf{89.2\%} & 0.206 & 0.316 & 0.060 & \textbf{0.048} & 0.254 & 0.372 & 0.110 & \textbf{0.056} \\
            5  & 72.3\% & 61.3\% & 80.2\% & \textbf{89.1\%} & 0.209 & 0.320 & 0.059 & \textbf{0.049} & 0.271 & 0.402 & 0.118 & \textbf{0.058} \\
            6  & 72.1\% & 61.0\% & 80.5\% & \textbf{88.9\%} & 0.211 & 0.323 & 0.059 & \textbf{0.050} & 0.285 & 0.431 & 0.126 & \textbf{0.060} \\
            7  & 71.9\% & 60.6\% & 80.6\% & \textbf{88.9\%} & 0.213 & 0.327 & 0.059 & \textbf{0.050} & 0.298 & 0.459 & 0.133 & \textbf{0.062} \\
            8  & 71.7\% & 60.3\% & 80.6\% & \textbf{88.7\%} & 0.215 & 0.331 & 0.058 & \textbf{0.050} & 0.308 & 0.485 & 0.139 & \textbf{0.063} \\
            9  & 71.5\% & 59.9\% & 80.7\% & \textbf{88.6\%} & 0.216 & 0.335 & 0.058 & \textbf{0.051} & 0.318 & 0.510 & 0.145 & \textbf{0.065} \\
            10 & 71.3\% & 59.6\% & 80.7\% & \textbf{88.6\%} & 0.218 & 0.339 & 0.059 & \textbf{0.051} & 0.327 & 0.533 & 0.150 & \textbf{0.066} \\
            11 & 71.1\% & 59.2\% & 80.6\% & \textbf{88.6\%} & 0.219 & 0.343 & 0.059 & \textbf{0.052} & 0.335 & 0.555 & 0.154 & \textbf{0.067} \\
            12 & 70.9\% & 58.9\% & 80.5\% & \textbf{88.5\%} & 0.221 & 0.347 & 0.059 & \textbf{0.052} & 0.343 & 0.576 & 0.158 & \textbf{0.068} \\
            13 & 70.7\% & 58.6\% & 80.4\% & \textbf{88.5\%} & 0.223 & 0.351 & 0.060 & \textbf{0.052} & 0.352 & 0.598 & 0.162 & \textbf{0.069} \\
            14 & 70.3\% & 58.2\% & 80.2\% & \textbf{88.4\%} & 0.226 & 0.355 & 0.060 & \textbf{0.052} & 0.364 & 0.619 & 0.166 & \textbf{0.070} \\
            15 & 69.7\% & 57.9\% & 80.0\% & \textbf{88.4\%} & 0.231 & 0.360 & 0.061 & \textbf{0.053} & 0.380 & 0.641 & 0.169 & \textbf{0.071} \\
            16 & 68.9\% & 57.6\% & 79.6\% & \textbf{88.3\%} & 0.239 & 0.365 & 0.063 & \textbf{0.053} & 0.402 & 0.662 & 0.173 & \textbf{0.072} \\
            \midrule
            Mean  & 71.8\% & 60.3\% & 80.3\% & \textbf{88.9\%} & 0.214 & 0.332 & 0.060 & \textbf{0.050} & 0.295 & 0.472 & 0.133 & \textbf{0.061} \\
            \bottomrule
        \end{tabular}}
        % \caption{animal sparse us}
          \caption{\textbf{4D Shape Reconstruction for Unseen Motions (DT4D-A)} We evaluate IoU, Chamfer distance, and correspondence distance for 17 timeframes for the 4D shape reconstruction from sparse point cloud task on seen individuals but unseen motions of DT4D-A \cite{DeformingThings4D} dataset.}
        \label{Tab:animal_sparse_us}
    \end{center}
    % \vspace{5pt}
\end{table*}
\begin{table*}[t]
    % \vspace{5pt}
    \renewcommand\arraystretch{1.1}
    \begin{center}
        \scalebox{0.92}{
        \begin{tabular}{*{14}{c}}
            \toprule
            \multirow{2}*{Time step} & \multicolumn{4}{c}{IoU}   & \multicolumn{4}{c}{Chamfer} & \multicolumn{4}{c}{Correspond.}\\
            \cmidrule(lr){2-5} \cmidrule(lr){6-9} \cmidrule(lr){10-13}
            & OFlow & LPDC & CaDex & \textbf{Ours} & OFlow & LPDC & CaDex & \textbf{Ours} & OFlow & LPDC & CaDex & \textbf{Ours} \\
            \midrule
            \midrule
            0  & 62.4\% & 53.5\% & 64.2\% & \textbf{84.8\%} & 0.294 & 0.404 & 0.129 & \textbf{0.056} & 0.216 & 0.296 & 0.150 & \textbf{0.057} \\
            1  & 62.1\% & 52.8\% & 64.4\% & \textbf{84.7\%} & 0.296 & 0.412 & 0.128 & \textbf{0.056} & 0.235 & 0.330 & 0.157 & \textbf{0.060} \\
            2  & 61.7\% & 52.6\% & 64.5\% & \textbf{84.6\%} & 0.300 & 0.414 & 0.127 & \textbf{0.056} & 0.260 & 0.358 & 0.169 & \textbf{0.062} \\
            3  & 61.4\% & 52.4\% & 64.6\% & \textbf{84.4\%} & 0.304 & 0.417 & 0.127 & \textbf{0.057} & 0.283 & 0.390 & 0.183 & \textbf{0.065} \\
            4  & 61.1\% & 52.2\% & 64.6\% & \textbf{84.3\%} & 0.307 & 0.420 & 0.126 & \textbf{0.057} & 0.303 & 0.422 & 0.197 & \textbf{0.068} \\
            5  & 60.9\% & 51.9\% & 64.7\% & \textbf{84.1\%} & 0.309 & 0.423 & 0.126 & \textbf{0.057} & 0.321 & 0.453 & 0.211 & \textbf{0.070} \\
            6  & 60.7\% & 51.7\% & 64.8\% & \textbf{83.9\%} & 0.311 & 0.427 & 0.126 & \textbf{0.058} & 0.336 & 0.483 & 0.224 & \textbf{0.072} \\
            7  & 60.5\% & 51.4\% & 64.8\% & \textbf{83.8\%} & 0.313 & 0.430 & 0.125 & \textbf{0.058} & 0.350 & 0.511 & 0.235 & \textbf{0.074} \\
            8  & 60.4\% & 51.2\% & 64.8\% & \textbf{83.6\%} & 0.315 & 0.433 & 0.125 & \textbf{0.059} & 0.362 & 0.538 & 0.246 & \textbf{0.076} \\
            9  & 60.2\% & 51.0\% & 64.8\% & \textbf{83.4\%} & 0.317 & 0.437 & 0.125 & \textbf{0.059} & 0.374 & 0.563 & 0.256 & \textbf{0.077} \\
            10 & 60.1\% & 50.7\% & 64.8\% & \textbf{83.3\%} & 0.319 & 0.440 & 0.125 & \textbf{0.059} & 0.385 & 0.588 & 0.266 & \textbf{0.078} \\
            11 & 60.0\% & 50.5\% & 64.8\% & \textbf{83.3\%} & 0.321 & 0.443 & 0.125 & \textbf{0.059} & 0.395 & 0.611 & 0.274 & \textbf{0.080} \\
            12 & 59.8\% & 50.3\% & 64.8\% & \textbf{83.2\%} & 0.323 & 0.446 & 0.125 & \textbf{0.060} & 0.405 & 0.633 & 0.282 & \textbf{0.081} \\
            13 & 59.7\% & 50.1\% & 64.7\% & \textbf{83.1\%} & 0.325 & 0.449 & 0.126 & \textbf{0.060} & 0.416 & 0.654 & 0.289 & \textbf{0.082} \\
            14 & 59.4\% & 49.9\% & 64.6\% & \textbf{83.0\%} & 0.329 & 0.452 & 0.126 & \textbf{0.060} & 0.428 & 0.675 & 0.296 & \textbf{0.083} \\
            15 & 59.1\% & 49.7\% & 64.5\% & \textbf{82.9\%} & 0.334 & 0.455 & 0.127 & \textbf{0.060} & 0.445 & 0.696 & 0.303 & \textbf{0.084} \\
            16 & 58.6\% & 49.5\% & 64.3\% & \textbf{82.8\%} & 0.340 & 0.459 & 0.128 & \textbf{0.061} & 0.466 & 0.716 & 0.309 & \textbf{0.085} \\
            \midrule
            Mean  & 60.5\% & 51.3\% & 64.6\% & \textbf{83.7\%} & 0.315 & 0.433 & 0.126 & \textbf{0.058} & 0.352 & 0.525 & 0.238 & \textbf{0.074} \\
            \bottomrule
        \end{tabular}}
        % \caption{animal sparse ui}
          \caption{\textbf{4D Shape Reconstruction for Unseen Individuals (DT4D-A)} We evaluate IoU, Chamfer distance, and correspondence distance for 17 timeframes for the 4D shape reconstruction from sparse point cloud task on unseen individuals of DT4D-A \cite{DeformingThings4D} dataset.}
        \label{Tab:animal_sparse_ui}
    \end{center}
    % \vspace{5pt}
\end{table*}

\end{document}